\RequirePackage{fix-cm}
\documentclass[twocolumn]{svjour3}          
\smartqed  
\usepackage{graphicx}
\usepackage{algorithm}
\usepackage{subfigure}
\usepackage{algpseudocode}
\usepackage{multirow}
\usepackage{hhline}
\usepackage{balance}
\usepackage{color}
\usepackage{tikz}
\usetikzlibrary{arrows,shapes,positioning,shadows,trees}

\def\etal{\emph{et al.}}
\def\eg{\emph{e.g.}}
\def\etc{\emph{etc.}}
\graphicspath{{figures/}}

\usepackage{xcolor}

\newcount\Comments  
\usepackage{color}
\definecolor{darkgreen}{rgb}{0,0.5,0}
\definecolor{purple}{rgb}{1,0,1}
\newcommand{\kibitz}[2]{\ifnum\Comments=1\textcolor{#1}{#2}\fi}

\begin{document}

\title{Human Action Recognition and Prediction: A Survey
}


\author{Yu~Kong         \and
        Yun~Fu 
}


\institute{Yu Kong \at
              B. Thomas Golisano College of Computing and Information Sciences, Rochester Institute of Technology, Rochester, NY, USA \\
              \email{yu.kong@rit.edu}           
           \and
           Yun~Fu \at
              Department of ECE and College of CIS, Northeastern University, Boston, MA, USA\\
              \email{yunfu@ece.neu.edu}
}

\date{Received: date / Accepted: date}

\maketitle

\begin{abstract}
Derived from rapid advances in computer vision and machine learning, video analysis tasks have been moving from inferring the present state to predicting the future state. Vision-based action recognition and prediction from videos are such tasks, where action recognition is to infer human actions (present state) based upon complete action executions, and action prediction to predict human actions (future state) based upon incomplete action executions. These two tasks have become particularly prevalent topics recently because of their explosively emerging real-world applications, such as visual surveillance, autonomous driving vehicle, entertainment, and video retrieval, etc. Many attempts have been devoted in the last a few decades in order to build a robust and effective framework for action recognition and prediction. In this paper, we survey the complete state-of-the-art techniques in action recognition and prediction. Existing models, popular algorithms, technical difficulties, popular action databases, evaluation protocols, and promising future directions are also provided with systematic discussions.
\end{abstract}

\section{Introduction}
Every human action, no matter how trivial, is done for some purpose. For example, in order to complete a physical exercise, a patient is interacting with and responding to the environment using his/her hands, arms, legs, torsos, bodies, etc. An action like this denotes everything that can be observed, either with bare eyes or measured by visual sensors. Through the human vision system, we can understand the action and the purpose of the actor. We can easily know that a person is exercising, and we could guess with a certain confidence that the person's action complies with the instruction or not. However, it is way too expensive to use human labors to monitor human actions in a variety of real-world scenarios, such as smart rehabilitation and visual surveillance. Can a machine perform the same as a human? 


One of the ultimate goals of artificial intelligence research is to build a machine that can accurately understand humans' actions and intentions, so that it can better serve us. Imagine that a patient is undergoing a rehabilitation exercise at home, and his/her robot assistant is capable of recognizing the patient's actions, analyzing the correctness of the exercise, and preventing the patient from further injuries. Such an intelligent machine would be greatly beneficial as it saves the trips to visit the therapist, reduces the medical cost, and makes remote exercise into reality. Other important applications including visual surveillance, entertainment, and video retrieval also need to analyze human actions in videos. In the center of these applications is the computational algorithms that can understand human actions. Similar to the human vision system, the algorithms ought to produce a label after observing the entire or part of a human action execution \cite{BobickTPAMI2001,RyooICCV2011}. Building such algorithms is typically addressed in computer vision research, which studies how to make computers gain high-level understanding from digital images and videos.



\begin{figure*}[!t]
\caption{Framework of the survey. The picture presents the topics discussed in the survey organized in a hierarchical tree, a list of representative works are also included for each topic.}
\label{fig:IJCVframework}
\begin{center}
\includegraphics[width=16cm]{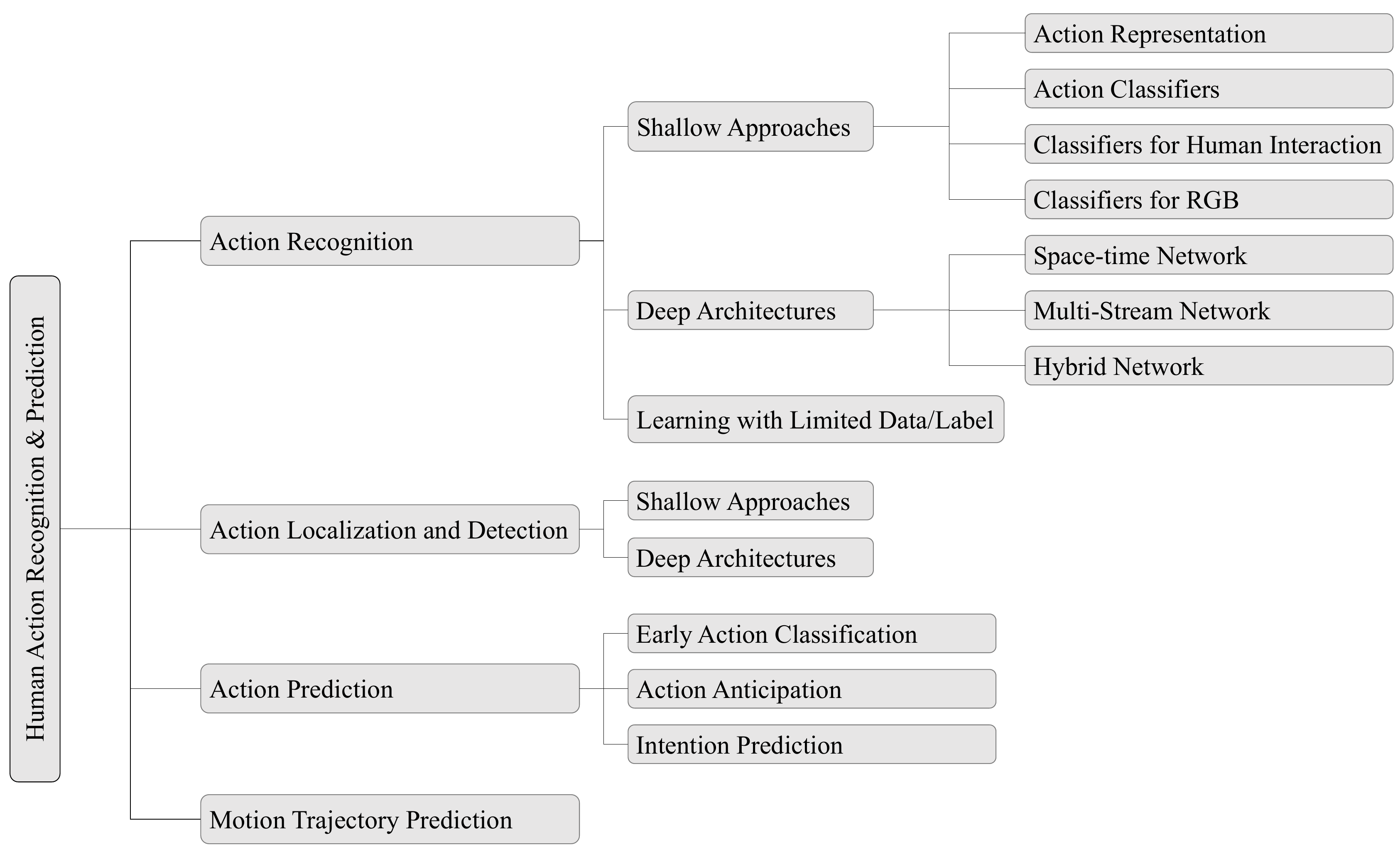}
\end{center}

\end{figure*}

\tikzset{
  basic/.style  = {draw, text width=2cm, drop shadow, font=\sffamily, rectangle},
  root/.style   = {basic, rounded corners=2pt, thin, align=center,
                   fill=gray!20},
  level 2/.style = {basic, rounded corners=6pt, thin,align=center, fill=gray!20,
                   text width=15em},
  level 3/.style = {basic, thin, align=left, fill=gray!20,rectangle, text width=10em}
}

\begin{figure*}[t]
\begin{center}
\includegraphics[width=1\linewidth]{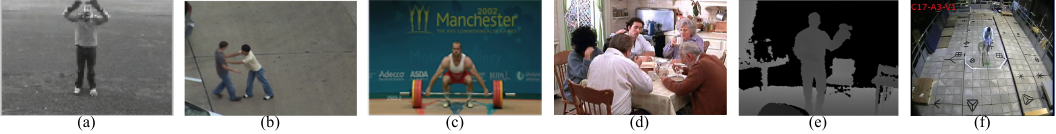}
\end{center}
\caption{Example frames of action videos used in computer vision research. (a) single person's action; (b) human interaction; (c) human-object interaction; (d) group action; (e) RGB-D action; (f) multi-view action.}
\label{fig:fig1}
\end{figure*}

The term \textit{human action} studied in computer vision research ranges from the simple limb movement to joint complex movement of multiple limbs and the human body. This process is dynamic, and thus is usually conveyed in a video lasting a few seconds. Though it might be difficult to give a formal definition of human action studied in the computer vision community, we provide some examples used in the community. Typical example actions are, 1) an individual action in KTH dataset \cite{SchuldtICPR2004} (Fig.~\ref{fig:fig1}(a)), which contains simple daily actions such as ``clapping'' and ``running''; 2) a human interaction in UT-Interaction dataset \cite{RyooICCV2009} (Fig.~\ref{fig:fig1}(b)), which consists of human interactions including ``handshake'' and ``push''; 3) a human-object interaction in UCF Sports dataset \cite{RodriguezCVPR2008} (Fig.~\ref{fig:fig1}(c)), which comprises of sport actions and human-object interactions;  4) a group action in Hollywood 2 dataset \cite{MarszalekCVPR2009} (Fig.~\ref{fig:fig1}(d)); 5) an action captured by a RGB-D sensor in UTKinect dataset \cite{XiaCVPRW2012} (Fig.~\ref{fig:fig1}(e)); and 6) a multi-view action in Multicamera dataset \cite{Singh2010} (Fig.~\ref{fig:fig1}(f)) capturing human actions from multiple camera views. In all these examples, a human action attempts to achieve a certain goal, in which some of them can be achieved by simply moving arms, and the others need to be accomplished in several steps. 

Technology advances in computer science and engineering have been enabling machines to understand human actions in videos. There are two basic topics in the computer vision community, vision-based human action recognition and prediction:
\begin{enumerate}
    \item \textbf{Action recognition}: recognize a human action from a video containing complete action execution.
    \item \textbf{Action prediction}: reason a human action from temporally incomplete video data.
\end{enumerate}

\begin{figure}[!t]
\begin{center}
\subfigure[Action recognition]{\includegraphics[width=0.48\linewidth]{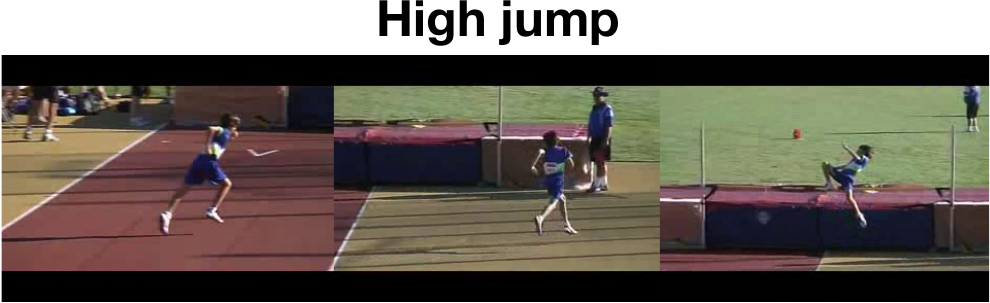}}
\subfigure[Action prediction]{\includegraphics[width=0.48\linewidth]{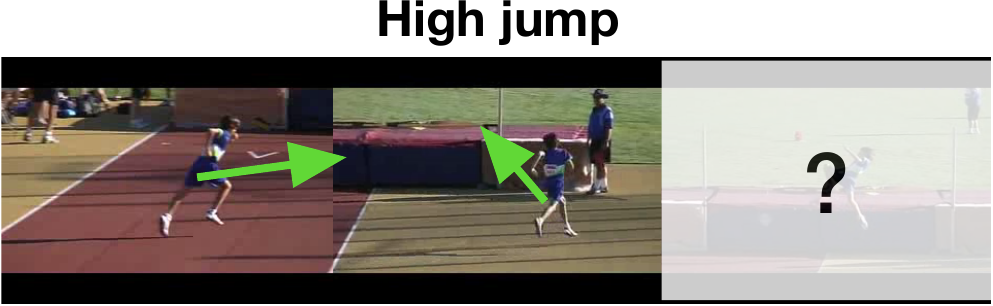}}
\end{center}
\caption{(a) Action recognition task infers an action category from a video containing complete action execution, while (b) action prediction task infers a label from temporally incomplete video. The label could be an action category (early action classification), or a motion trajectory (trajectory prediction).}
\label{fig:recpred}
\end{figure}

\textit{Action recognition} is a fundamental task in the computer vision community that recognizes human actions based on the complete action execution in a video (see Figure~\ref{fig:recpred}(a)) \cite{BobickTPAMI2001,EfrosICCV2003,WeinlandCVIU2006,LaptevIJCV2005,LiuCVPR2009,TangCVPR2012,TranICCV2015}. 
It has been studied for decades and is still a very popular topic due to broad real-world applications including video retrieval \cite{CiptadiECCV2014}, visual surveillance \cite{HuTIP2007,Singh2010}, etc. Researchers have made great efforts to create an intelligent system mimicking humans' capability that can recognize complex human actions in cluttered environments. However, to a machine, an action in a video is just an array of pixels. The machine has no idea about how to convert these pixels into an effective representation, and how to infer human actions from the representation. These two problems are considered as \textit{action representation} and \textit{action classification} in action recognition, and many attempts \cite{LaptevIJCV2005,RaptisCVPR2013,JiTPAMI2013,CarreiraCVPR2017} have been proposed to address these two problems.

On the contrary, \textit{action prediction} is a before-the-fact video understanding task and is focusing on the future state. In some real-world scenarios (\eg, vehicle accidents and criminal activities), intelligent machines do not have the luxury of waiting for the entire action execution before having to react to the action contained in it. For example, being able to predict a dangerous driving situation before it occurs; opposed to recognizing it thereafter. This is referred to as the action prediction task where approaches that can recognize and infer a label from a temporally incomplete video (see Figure~\ref{fig:recpred}(b))  \cite{RyooICCV2011,KongECCV2014,KongCVPR2017}, different to action recognition approaches that expect to see the entire set of action dynamics extracted from a full video.

The major difference between action recognition and action prediction lies in \textit{when to make a decision}. Human action recognition is to infer the action label \textit{after} the entire action execution has been observed. This task is generally useful in non-urgent scenarios, such as video retrieval, entertainment, etc. Nevertheless, action prediction is to infer \textit{before} fully observing the entire execution, which is of particular important in certain scenarios. For example, it would be very helpful if an intelligent system on a vehicle can predict a traffic accident before it happens; opposed to recognizing the dangerous accident event thereafter.


We will mainly discuss recent advance in action recognition and prediction in this survey. To ease the navigation of this paper, Fig. \ref{fig:IJCVframework} illustrates the topics discussed in this paper and the representative works are also included. Different from recent survey papers \cite{HerathIVC2017,PoppeIVC2010}, studies in action prediction are also described in this paper. Human action recognition and prediction are closely related to other computer vision tasks such as human gesture analysis, gait recognition, and event recognition. In this survey, we focus on the vision-based recognition and prediction of actions from videos that usually involve one or more people. The input is a series of video frames and the output is an action label. We are also interested in learning human actions from RGB-D videos. Some of existing studies \cite{YaoPAMI2012,YaoECCV2012} aim at learning actions from static images, which is not the focus of this paper. This paper will first give an overview of recent studies in action recognition and prediction, describe popular human actions datasets, and will then discuss several interesting future directions in details.



\subsection{Real-World Applications}
\begin{figure}[t]
\begin{center}
\includegraphics[width=1\linewidth]{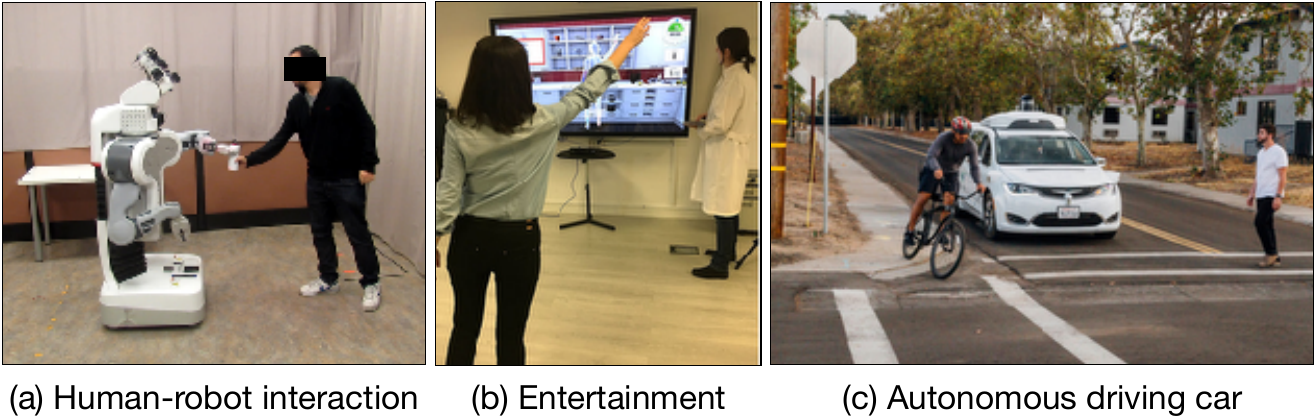}
\end{center}
\caption{Examples of real-world applications using action recognition techniques.}
\label{fig:app}
\end{figure}

Action recognition and prediction algorithms empower many real-world applications (examples are shown in Figure~\ref{fig:app}). State-of-the-art algorithms \cite{WangECCV2016,FeichtenhoferCVPR2017,KongAAAI2018,MaCVPR2016} remarkably reduce the human labor in analyzing a large-scale of video data and provide understanding on the current state and future state of ongoing video data. 

\subsubsection{Visual Surveillance}
Security issue is becoming more important in our daily life, and it is one of the most frequently discussed topics nowadays. Places under surveillance typically allow certain human actions, and other actions are not allowed \cite{HuTIP2007}. With the input of a network of cameras \cite{WeinlandCVIU2006,Singh2010}, a visual surveillance system powered by action recognition \cite{JiTPAMI2013,SimonyanNIPS2014,KarpathyCVPR2014} and prediction \cite{RyooICCV2011,KongECCV2014,KongCVPR2017} algorithms may increase the chances of capturing a criminal on video, and reduce the risk caused by criminal actions. For example, in Boston marathon bombing site, if we had such an intelligent visual surveillance system that can forewarn the public by looking at the criminal's suspicious action, the victims' lives could be saved. The cameras also make some people feel more secure, knowing the criminals are being watched.

\subsubsection{Video Retrieval}
Nowadays, due to the fast growth of technology, people can easily upload and share videos on the Internet. However, managing and retrieving videos according to video content is becoming a tremendous challenge as most search engines use the associated text data to manage video data \cite{Ramezani2016}. The text data, such as tags, titles, descriptions, and keywords, can be incorrect, obscure, and irrelevant, making video retrieval unsuccessful \cite{Zhai2013}. An alternative method is to analyze human actions in videos, as the majority of these videos contain such a cue. For example, in \cite{CiptadiECCV2014}, researchers created a video retrieval framework by computing the similarity between action representations, and used the proposed framework to retrieve videos of children with autism in a classroom setting. Compared to conventional human action recognition task, the video retrieval task relies on the retrieval ranking instead of classification \cite{Ramezani2016}. 

\subsubsection{Entertainment}
The gaming industry in recent years has attracted an increasingly large and diverse group of people. A new generation of games based on full body play such as dance and sports games have increased the appeal of gaming to family members of all ages. To enable accurate perception of human actions, these games use cost-effective RGB-D sensors (\eg, Kinect \cite{ShottonPAMI2013}) which provide an additional depth channel data \cite{XiaCVPR2013,YangCVPR2014,HadfieldCVPR2013}. This depth data encode rich structural information of the entire scene, and facilitate action recognition task as it simplifies intra-class motion variations and reduces cluttered background noise \cite{KongCVPR2015,KongIJCV2017,JiaMM2014,LiuIJCAI2013}.

\subsubsection{Human-Robot Interaction}
Human-robot interaction is popularly applied in home and industry environment. Imagine that a person is interacting with a robot and asking it to perform certain tasks, such as ``passing a cup of water'' or ``performing an assembling task''. Such an interaction requires communications between robots and humans, and visual communication is one of the most efficient ways \cite{RyooHRI2015,KoppulaTPAMI2016}. 

\subsubsection{Autonomous Driving Vehicle}
Action prediction algorithms \cite{RyooIJCV2011,KongTPAMI2016} could be one of the potentials and maybe most important building components in an autonomous driving vehicle. Action prediction algorithms can predict a person's intention \cite{PeiICCV2011,LiTPAMI2014,KoppulaTPAMI2016} in a short period of time. In an urgent situation, a vehicle equipped with an action prediction algorithm can predict a pedestrian's future action or motion trajectory in the next few seconds, and this could be critical to avoid a collision. By analyzing human body motion characteristics at an early stage of an action using so-called interest points or convolutional neural network \cite{KongCVPR2017}, action prediction algorithms \cite{KongCVPR2017,KongTPAMI2016} can understand the possible actions by analyzing the action evolution without the need to observe the entire action execution. 

\subsection{Research Challenges}
Despite significant progress has been made in human action recognition and prediction, state-of-the-art algorithms still misclassify actions due to several major challenges in these tasks.

\subsubsection{Intra- and inter-class Variations}
\begin{figure}[t]
\begin{center}
\includegraphics[width=1\linewidth]{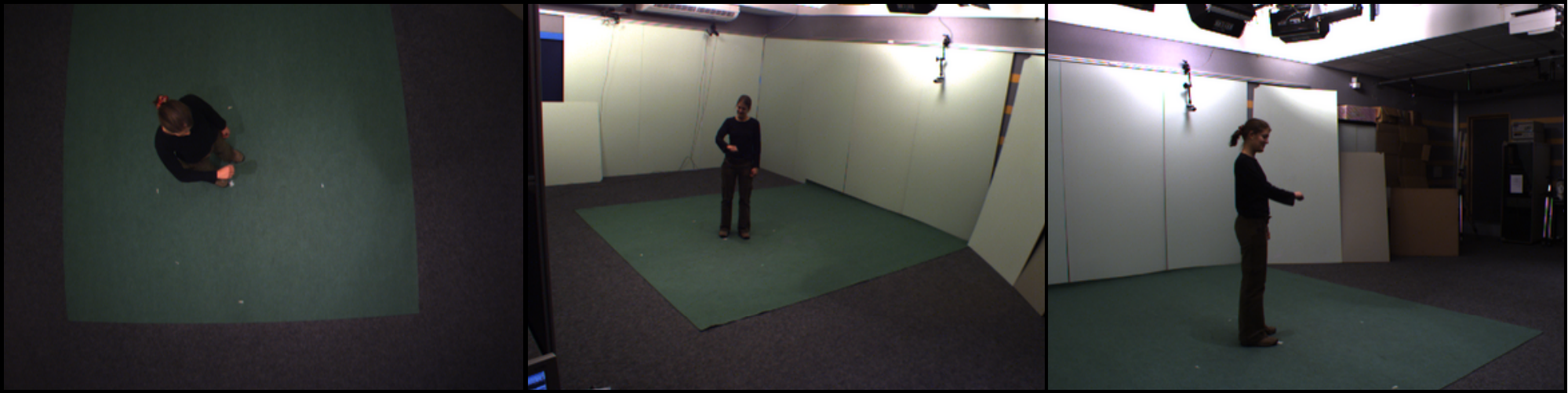}
\end{center}
\caption{Appearance variations in different camera views.}
\label{fig:multiview}
\end{figure}

As we all know, people behave differently for the same actions. For a given semantic meaningful action, for example, ``running'', a person can run fast, slow, or even jump and run. That is to say, one action category may contain multiple different styles of human movements. In addition, videos in the same action can be captured from various viewpoints. They can be taken in front of the human subject, on the side of the subject, or even on top of the subject, showing appearance variations in different views (see Figure~\ref{fig:multiview}). Furthermore, different people may show different poses in executing the same action. All these factors will result in large intra-class appearance and pose variations, which confuse a lot of existing action recognition algorithms. These variations will be even larger on real-world action datasets \cite{KarpathyCVPR2014,caba2015activitynet}. This triggers the investigation of more advanced action recognition algorithms that can be deployed in real-world scenarios. Furthermore, similarities exist in different action categories. For instance, ``running'' and ``walking'' involve similar human motion patterns. These similarities would also be challenging to differentiate for intelligent machines, and consequently contribute to misclassifications.


\subsubsection{Cluttered Background and Camera Motion}
It is interesting to see that a number of human action recognition algorithms work very well in indoor controlled environments but not in outdoor uncontrolled environments. This is mainly due to the background noise. In fact, most of the existing activity features such as histograms of oriented gradient \cite{LaptevCVPR2008} and interest points \cite{DollarICCVW2005} also encode background noise, and thus degrade the recognition performance. Camera motion is another factor that should be considered in real-world applications. Due to significant camera motion, action features cannot be accurately extracted. In order to better extract action features, camera motion should be modeled and compensated \cite{WangICCV2013}. Other environment-related issues such as illumination conditions, viewpoint changes, dynamic background will also be the challenges that prohibit action recognition algorithms from being used in practical scenarios.

\subsubsection{Insufficient Annotated Data}
Even though existing action recognition approaches \cite{KlaserBMVC2008,LiuCVPR2011,NieblesECCV2010} have shown impressive performance on small-scale datasets in laboratory settings, it is really challenging to generalize them to real-world applications due to their inability of training on large-scale datasets. Recent deep approaches \cite{WangECCV2016,FeichtenhoferCVPR2017} have shown promising results on datasets captured in uncontrolled settings, but they normally require a large amount of annotated training data. Action datasets such as HMDB51 \cite{KuehneICCV2011} and UCF-101 \cite{Soomro2012} contain thousands of videos, but still far from enough for training deep networks with millions of parameters. Although Youtube-8M \cite{abu2016youtube} and Sposrts-1M datasets \cite{KarpathyCVPR2014} provide millions of action videos, their annotations are generated by a retrieval method, and thus may not be accurate. Training on such datasets would hurt the performance of action recognition algorithms that do not have a tolerance to inaccurate labels. However, it is possible that some of the data annotations are available, which would result in a training setting with a mixture of labeled data and unlabeled data. Therefore, it is imperative to design action recognition algorithms that can learn actions from both labeled data and unlabeled data.

\subsubsection{Action Vocabulary}
Actions could be categorized into different levels, movements, atomic actions, composite actions, events, etc. This defines an action hierarchy, and complex actions at high levels of the hierarchy can be decomposed into a combination of actions at a lower level. How to define and analyze these different kinds of actions is very important.

\subsubsection{Uneven Predictability}
Not all frames are equally discriminative. As shown in \cite{RaptisCVPR2013,Vahdat2011}, a video can be effectively represented by a small set of key frames. This indicates that lots of frames are redundant, and discriminative frames may appear anywhere in the video. However, action prediction methods \cite{RyooICCV2011,KongECCV2014,MaCVPR2016,LanECCV2014} require the beginning portions of the video to be discriminative in order to maximize predictability. To solve this problem, context information is transferred to the beginning portions of the videos \cite{KongCVPR2017}, but the performance is still limited due to the insufficient discriminative information.

In addition, actions differ in their predictabilities \cite{LiTPAMI2014,KongCVPR2017}. As shown in \cite{KongCVPR2017}, some actions are instantly predictable while the other ones need more frames to be observed. However, in practical scenarios, it is necessary to predict any actions as early as possible. This requires us to create general action prediction algorithms that can make accurate and early predictions for most of or all actions.

\section{Human Perception of Actions}
Human actions, particularly those involving whole-body and limb (\eg, arms and legs) movements, and interactions with their environment contain rich information about the performer's intention, goal, mental status, etc. Understanding the actions and intentions of other people is one of the most important social skills we have, and the human vision system provides a particularly rich source of information in support of this skill \cite{Blake2007}. Compared to static images, human actions in videos provide even more reliable and more expressive information, and thus speak louder than images when it comes to understanding what others are doing \cite{Darwin1872}. There are a number of information we can tell from human actions, including the action categories \cite{Mass1971}, emotional implication \cite{Clarke2005}, identity \cite{Cutting1977,Troje2005}, gender \cite{Sumi2000,Troje2002}, etc. The human visual system is finely optimized for the perception of human movements \cite{Decety1999}. 

Action understanding by humans is a complex cognitive capability performed by a complex cognitive mechanism. Such a mechanism can be decomposed into three major components, including action recognition, intention understanding, and narrative understanding \cite{Keestra2015}. Ricoeur \cite{Ricoeur1992} suggested that actions can be approached with a set of interrelated questions including, who, what, why, how, where, and when. Three questions are prioritized, which offer different perspectives on the action: what is the action, why is the action being done, and who is the agent. Computational models for the first two questions have been extensively investigated in action recognition \cite{BlankICCV2005,PerezBMVC2010,ChoiICCVW2009,MarszalekCVPR2009,KuehneICCV2011,JiTPAMI2013,TranICCV2015} and prediction \cite{RyooICCV2011,KongECCV2014,MaCVPR2016,CaoCVPR2013} research in the computer vision community. The last question ``who is the agent'' refers to the agent's identity, or social role, which provides a more thoroughgoing understanding of the ``who'' behind it, and thus is referred to as narrative understanding \cite{Ricoeur1992}. Few work in the computer vision community studies this question \cite{LanCVPR2012,RamanathanCVPR2013}.


Some of the human actions are goal-oriented, i.e., a goal is completed by performing one or a series of actions. Understanding such actions is crucial for predicting the effects or outcomes of the actions. As humans, we make inferences about the action goals of an individual by evaluating the end state that would be caused by their actions, given particular situational or environmental constraints. The inference is possibly made by a direct matching process of a mirror neuron system, which maps the observed action onto our own motor representation of that action \cite{Rizzolatti2004,Rizzolatti2010}. According to the direct matching hypothesis, the prediction of one's action goal is heavily relying on the observer's action vocabulary or knowledge. Another cue for making action prediction is from emotional or attentional information, such as the facial expression and gaze or the other individuals. Such referential information makes the observer pay attention to the specific objects because of the particular relations that link these cues to their referents. These psychological and cognitive findings would be helpful for designing action prediction approaches.


\section{Action Recognition}
A typical action recognition flowchart generally contains two major components \cite{SchuldtICPR2004,WangIJCV2013,PoppeIVC2010}, action representation and action classification. The action representation component basically converts an action video into a feature vector \cite{LaptevIJCV2005,DollarICCVW2005,WangIJCV2015,Scovanner2007} or a series of vectors \cite{NieblesECCV2010,KongCVPR2017,MorencyCVPR2007}, and the action classification component infers an action label from the vector \cite{LiuCVPR2011,SminchisescuICCV2005,ShiIJCV2011}. Recently, deep networks \cite{JiTPAMI2013,TranICCV2015,FeichtenhoferCVPR2017} merge these two components into a unified end-to-end trainable framework, which further enhance the classification performance in general. In this section we will discuss recent work in action representation, action classification, and deep networks.

\begin{figure}[t]
\begin{center}
\includegraphics[width=0.9\linewidth]{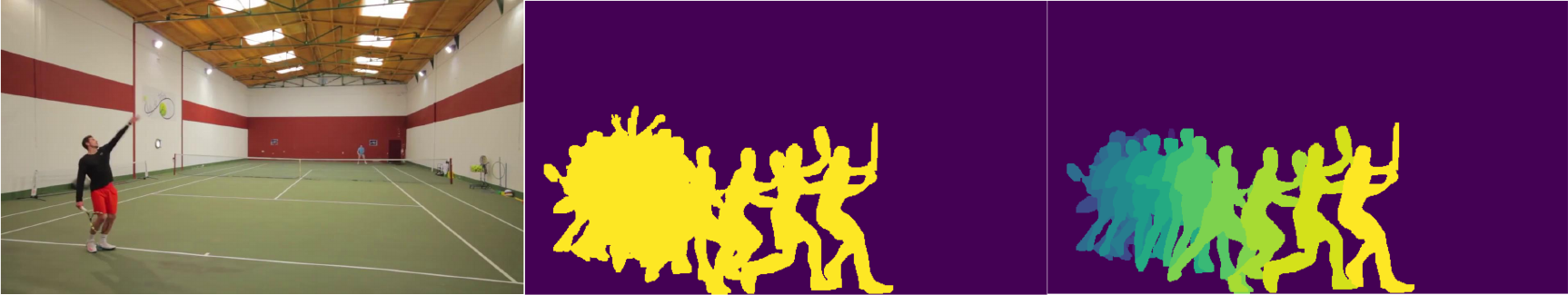}
\end{center}
\caption{Examples of an input video frame, the corresponding motion energy image and motion history image computed by \cite{BobickTPAMI2001}.}
\label{fig:meimhi}
\end{figure}
\subsection{Shallow Approaches}

\subsubsection{Action Representation}
The first and the foremost important problem in action recognition is \textit{how to represent an action in a video}. Human actions appearing in videos differ in their motion speed, camera view, appearance and pose variations, etc, making action representation a really challenging problem. A successful action representation method should be efficient to compute, effective to characterize actions, and can maximize the discrepancy between actions, in order to minimize the classification error.

One of the major challenges in action recognition is large appearance and pose variations in one action category, making the recognition task difficult. The goal of action representation is to convert an action video into a feature vector, extract representative and discriminative information of human actions, and minimize the variations, thereby improving the recognition performance. Action representation approaches can be roughly categorized into holistic features and local features, which will be discussed next.

Many attempts have been made in action recognition to convert action videos into discriminative and representative features, in order to minimize with-in class variations and maximize between class variations. Here, we focus on \textit{hand-crafted} action representation methods, which means the parameters in these methods are pre-defined by experts. This differs from deep networks, which can automatically learn parameters from data.

\paragraph{Holistic Representations}
Human action in a video generates a space-time shape in the 3D volume. This space-time shape encodes both spatial information of the human pose at various times, and dynamic information of the human body. Holistic representation methods capture the motion information of the entire human subject, providing rich and expressive motion information for action recognition. However, holistic representations tend to be sensitive to noise. It captures the information in a certain rectangle region, and thus may introduce irrelevant information and noise from the human subject and cluttered background.

One pioneering work in \cite{BobickTPAMI2001} presented Motion Energy Image (MEI) and Motion History Image (MHI) to encode dynamic human motion into a single image. As shown in Figure~\ref{fig:meimhi}, the two methods work on the silhouettes. The MEI method shows ``where'' the motion is occurring: the spatial distribution of motion is represented and the highlighted region suggests both the action occurring and the viewing condition. In addition to MEI, the MHI method shows both ``where'' and ``how'' the motion is occurring. Pixel intensity on a MHI is a function of the motion history at that location, where brighter values correspond to more recent motion.

Although MEI and MHI showed promising results in action recognition, they are sensitive to viewpoint changes. To address this problem, \cite{WeinlandCVIU2006} generalized \cite{BobickTPAMI2001} to 3D motion history volume (MHV) to remove the viewpoint dependency in the final action representation. MHV relies on the 3D voxels obtained from multiple camera views, and shows the 3D occupancy in the resulting volume. Fourier transform is then used to create features invariant to locations and rotations. 

\begin{figure}[t]
\begin{center}
\includegraphics[width=0.95\linewidth]{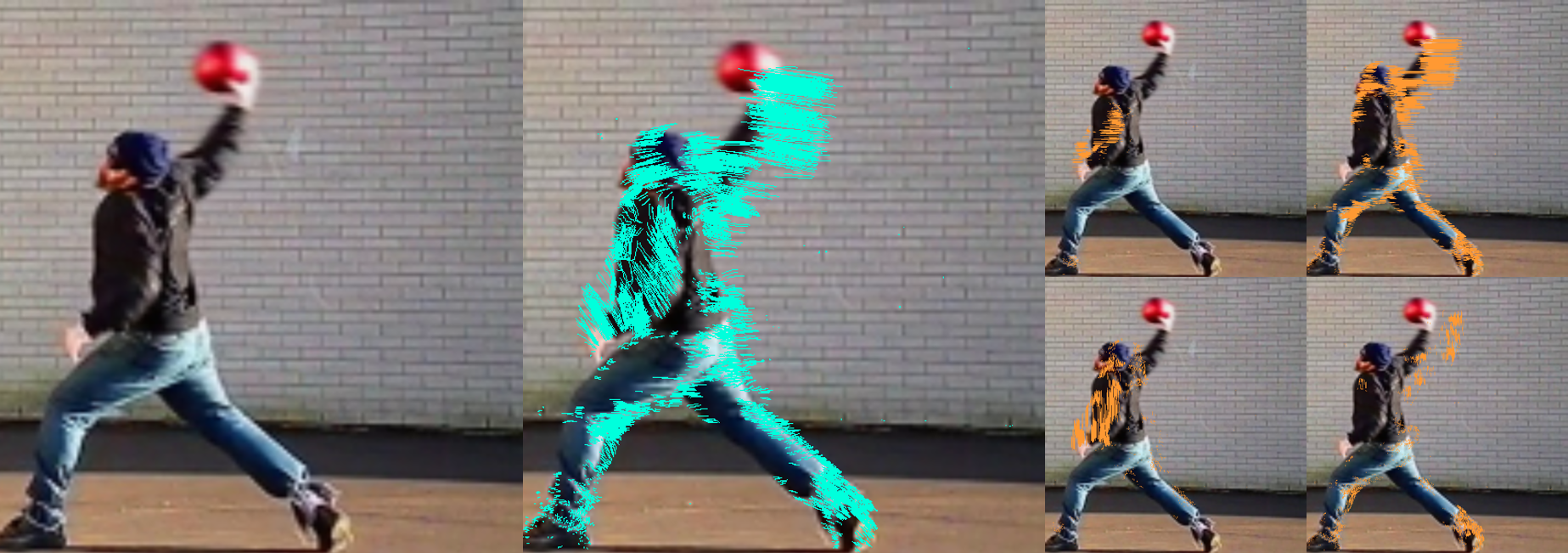}
\end{center}
\caption{Examples of the original frame, optical flow, and the flow field in four channels computed by \cite{EfrosICCV2003}.}
\label{fig:flow}
\end{figure}
To capture space-time information in human actions, \cite{GorelickPAMI2007,BlankICCV2005} utilized the Poisson equation to extract various shape properties for action representation and classification. Their method takes a space-time volume as input. Then the method discovers space-time saliency of moving body parts, and locally computes the orientation using the Poisson equation. These local properties are finally converted into a global feature by weighted averaging each point inside the volume. Another method to describe shape and motion was presented in \cite{YilmazCVPR2005}. In this method, a spatio-temporal volume is first generated by computing correspondences between frames. Then, spatio-temporal features by analyzing differential geometric surface properties from the volume.

Instead of computing silhouette or shape for action representation, motion information can also be computed from videos. One typical motion information is computed by the so-called optical flow algorithms \cite{Lucas1981,Horn1981,SunCVPR2010}, which indicate the pattern of apparent motion of objects on two consecutive frames. Under the assumption that illumination conditions do not change on the frames, optical flow computes the motion in the horizontal and vertical axis. An early work by Efros \etal \cite{EfrosICCV2003} split the flow field into four channels (see Figure~\ref{fig:flow}) capturing the horizontal and vertical motion in successive frames. This method was then used in \cite{WangPAMI2010} to describe the features of both the human body and the body parts. 


\paragraph{Local Representations}
Local representations only identify local regions having salient motion information, and thus inherently overcome the problem in holistic representations. Successful methods such as space-time interest points \cite{DollarICCVW2005,LaptevICCV2003,KlaserBMVC2008,BregonzioCVPR2009} and motion trajectory \cite{WangCVPR2011,WangIJCV2013} have shown their robustness to translation, appearance variation, etc. Different from holistic features, local features describe the local motion of a person in space-time regions. These regions are detected since the motion information within the regions is more informative and salient than the surrounding areas. After detection, the regions are described by extracting features in the regions. 

\begin{figure}[!t]
\begin{center}
\includegraphics[width=0.9\linewidth]{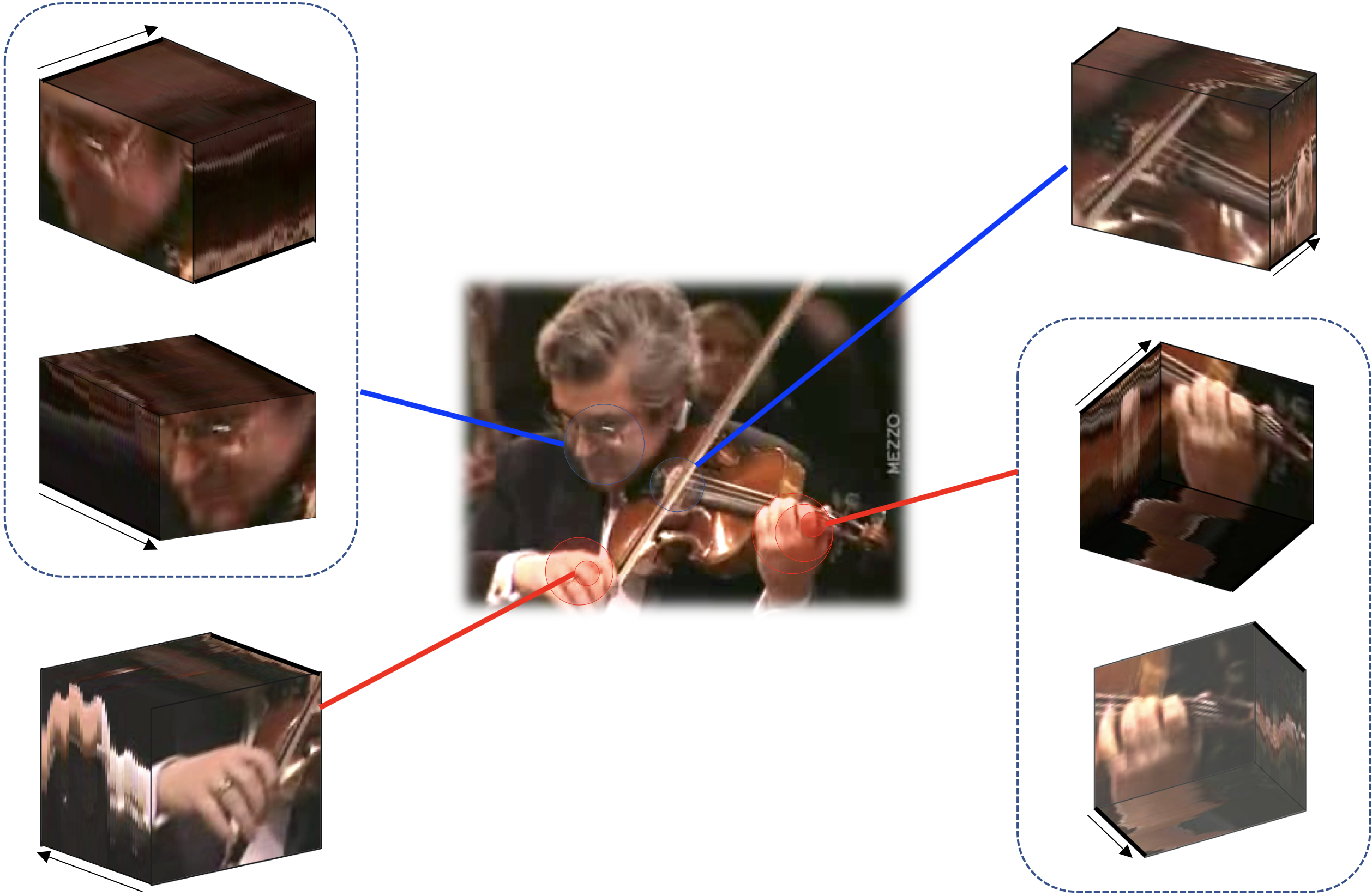}
\end{center}
\caption{Illustration of interest points detected on human body. Revised based on the original figure in \cite{HerathIVC2017}.}
\label{fig:interestpoint}
\end{figure}
Space-time interest points (STIPs) \cite{LaptevICCV2003,LaptevIJCV2005}-based approaches is one of the most important local representations. Laptev's seminal work \cite{LaptevICCV2003,LaptevIJCV2005} extended the Harris corner detector \cite{Harris1988} to space-time domain. A spatio-temporal separable Gaussian kernel is applied on a video to obtain its response function for finding large motion changes in both spatial and temporal dimensions  (see Figure~\ref{fig:interestpoint}). An alternative method was proposed in \cite{DollarICCVW2005}, which detects dense interest points. 2D Gaussian smoothing kernel is applied only along the spatial dimension, and the 1D Gabor filter is applied to the temporal dimension. Around each interest point, raw pixel values, gradient, and optical flow features are extracted and concatenated into a long vector. The principal component analysis is applied on the vector to reduce the dimensionality, and a k-means clustering algorithm is then employed to create the codebook of these feature vectors and generate one vector representation for a video \cite{SchuldtICPR2004}. Bregonzio \etal \cite{BregonzioCVPR2009} detected spatial-temporal interest points using Gabor filters. Spatiotemporal interest points can also be detected by using the spatiotemporal Hessian matrix \cite{WillemsECCV2008}. Other detection algorithms detect spatiotemporal interest points by extending their counterparts of 2D detectors to spatiotemporal domains, such as 3D SIFT \cite{Scovanner2007}, HOG3D \cite{KlaserBMVC2008}, local trinary patterns \cite{YeffetCVPR2009}, etc. Several descriptors have been proposed to describe the motion and appearance information within the small region of the detected interest points such as optical flow and gradient. Optical flow feature computed in a local neighborhood is further aggregated in histograms, called histograms of optical flow (HOF) \cite{LaptevCVPR2008}, and combined with HOG features \cite{DalalCVPR2005,KlaserBMVC2008} to represent complex human activities \cite{KlaserBMVC2008,LaptevCVPR2008,WangBMVC2009}. Gradients over optical flow fields are computed to build the so-called motion boundary histograms (MBH) for describing trajectories \cite{WangBMVC2009}.

\begin{figure}[!t]
\begin{center}
\includegraphics[width=1\linewidth]{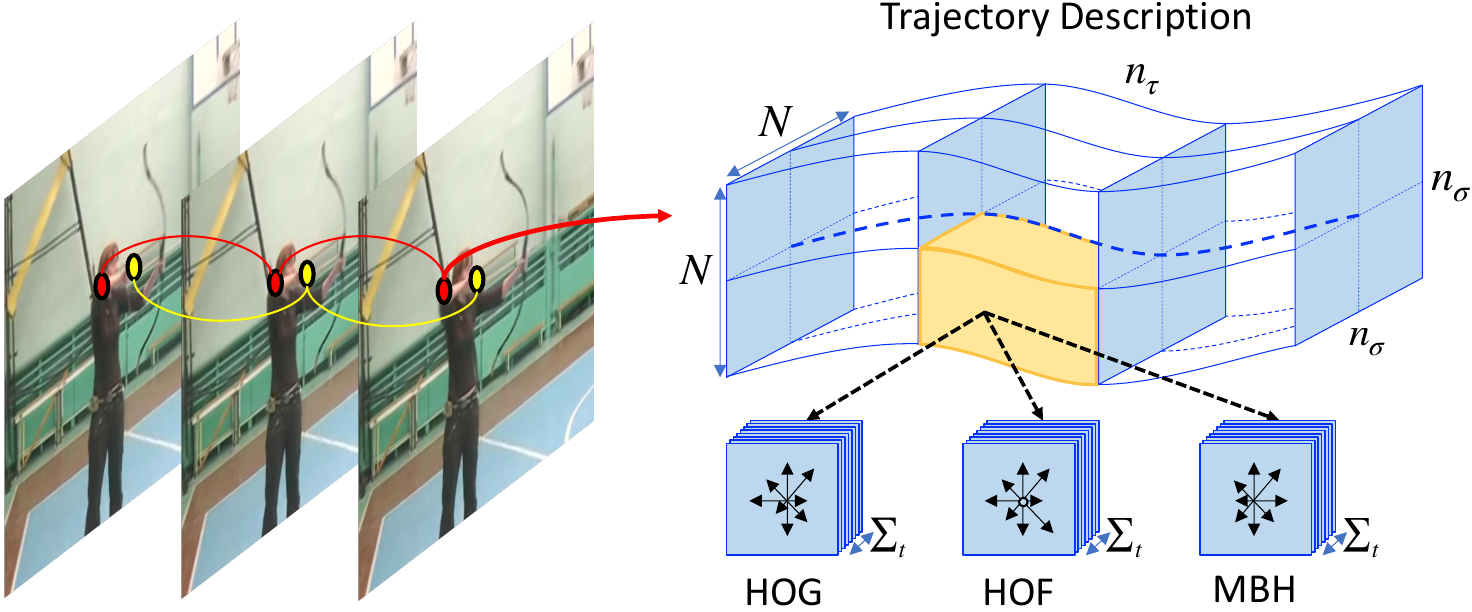}
\end{center}
\caption{Tracked point trajectories over frames, and are described by HOG, HOF and MBH features. Revised based on the original figure in \cite{WangIJCV2013}.}
\label{fig:traj}
\end{figure}

However, spatiotemporal interest points only capture information within a short temporal duration and cannot capture long-term duration information. It would be better to track these interest points and describe their changes of motion properties. Feature trajectory is a straightforward way of capturing such long-duration information \cite{WangBMVC2009,WangCVPR2011,RaptisECCV2010}. To obtain features for trajectories, in \cite{MessingICCV2009}, interest points are first detected and tracked using Harris3D interest points with a KLT tracker \cite{Lucas1981}. The method in \cite{SunCVPR2009} finds trajectories by matching corresponding SIFT points over consecutive frames. Hierarchical context information is captured in this method to generate more accurate and robust trajectory representation. Trajectories are described by a concatenation of HOG, HOF and MBH features \cite{WangCVPR2011,WangIJCV2013,JainCVPR2013} (see Figure~\ref{fig:traj}), intra- and inter-trajectory descriptors \cite{SunCVPR2009}, or HOG/HOF and averaged descriptors \cite{RaptisECCV2010}. In order to reduce the side effect of camera motion, \cite{WangICCV2013,WangIJCV2015} find correspondences between two frames first and then use RANSAC to estimate the homography. 

\subsubsection{Action Classifiers}
After action representations have been computed, action classifiers should be learned from training samples that determine the class boundaries for various action classes. Action classifiers can be roughly divided into the following categories:
\paragraph{Direct Classification}
This type of approaches summarize an action video into a feature vector, and then directly classify the vector into action categories using off-the-shelf classifiers such as support vector machine \cite{SchuldtICPR2004,LaptevCVPR2008,MarszalekCVPR2009}, k-nearest neighbor (k-NN) \cite{BlankICCV2005,LaptevICCV2007,TranECCV2008}, etc. In these methods, action dynamics are characterized in a holistic way using action shape \cite{GorelickPAMI2007,BlankICCV2005}, or using the so-called bag-of-words model, which captures local motion patterns using a histogram of visual words \cite{BlankICCV2005,LaptevICCV2007,SchuldtICPR2004,LaptevCVPR2008,MarszalekCVPR2009}. 

\begin{figure}[!t]
\begin{center}
\includegraphics[width=1\linewidth]{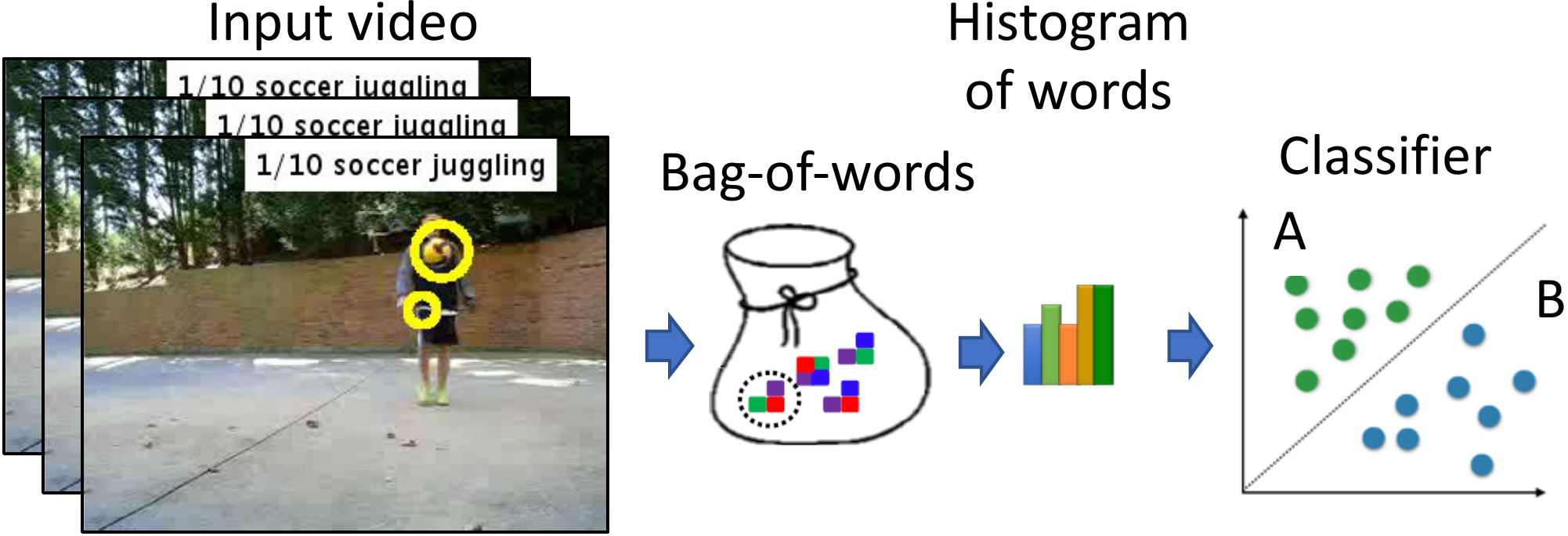}
\end{center}
\caption{A typical flowchart of the so-called bag-of-words methods. Local features detected on the input video are shown in yellow circles.}
\label{fig:direct}
\end{figure}

In fact, bag-of-words approaches received lots of attention in the last few years. As shown in Figure~\ref{fig:direct}, these approaches first detect local salient regions using the spatiotemporal interest point detectors \cite{DollarICCVW2005,SchuldtICPR2004,LaptevIJCV2005,KlaserBMVC2008}. Features such as gradient and optical flow are extracted around each 3D interest point. The principal component analysis is adopted to reduce the dimensionality of the features. Then the so-called visual words can be computed by k-means clustering \cite{SchuldtICPR2004}, or Fisher vector \cite{PerronninCVPR2006}. Finally, an action can be represented by a histogram of visual words, and can be recognized by a classifier such as the support vector machine. The bag-of-words model has been shown to be insensitive to appearance and pose variations \cite{WangBMVC2009}. However, it does not consider the temporal characteristics of human actions, as well as their structural information, which can be addressed by sequential approaches \cite{ShiIJCV2011,RaptisCVPR2013} and space-time approaches \cite{RyooICCV2009}, respectively.

\paragraph{Sequential Approaches}
This line of work captures temporal evolution of appearance or pose using sequential state models such as hidden Markov models (HMMs) \cite{DuongCVPR2005,RajkoCVPR2007,IkizlerCVPR2007}, conditional random fields (CRFs) \cite{SminchisescuICCV2005,WangCVPR2006,WangCVPR2007,MorencyCVPR2007} and structured support vector machine (SSVM) \cite{NieblesECCV2010,WangCVPR2012,TangCVPR2012,ShiIJCV2011}. These approaches treat a video as a composition of temporal segments or frames. The work in \cite{DuongCVPR2005} considers human routine trajectory in a room, and use a two-layer HMMs to model the trajectory. Recent work in \cite{RaptisCVPR2013} shows that representative key poses can be learned to better represent human actions. This method discards a number of non-informative poses in a temporal sequence, and builds a more compact pose sequence for classification. Nevertheless, these sequential approaches mainly use holistic features from frames, which are sensitive to background noise and generally do not perform well on challenging datasets.

\paragraph{Space-time Approaches}
Although direct approaches have shown promising results on some action datasets \cite{SchuldtICPR2004,LaptevCVPR2008,MarszalekCVPR2009}, they do not consider the spatiotemporal correlations between local features, and do not take the potentially
valuable information about the global spatio-temporal distribution of interest points into account. This problem was addressed in \cite{WuCVPR2011}, which learns a global Gaussian mixture model (GMM) using the relative coordinates features, and uses multiple GMMs to describe the distribution of interest points over local regions at multiple scales. A global feature on top of interest points was proposed in \cite{yuan20133d} to capture the detailed geometrical distribution of interest points. The feature is computed by extended 3D discrete Radon transform. Such a feature captures the geometrical information of the interest points, and is robust to geometrical transformation and noise. The spatiotemporal distribution of interest points is described by a Directional Pyramid Co-occurrence Matrix in (DPCM) \cite{yuan2014modeling}. DPCM characterizes the co-occurrence statistics of local features as well as the spatio-temporal positional relationships among the concurrent features. Graph is a powerful tool for modeling structured objects, and it was used in \cite{wu2014human} to capture the spatial and temporal relationships among local features. Local features are used as the vertices of the two-graph model and the relationships among local features in the intra-frames and inter-frames are characterized by the edges. A novel family of context-dependent graph kernels (CGKs) was proposed in \cite{wu2014human} to measure the similarity between the two-graph models. Although the above methods have achieved promising results, they are limited to small datasets as the correlations between interest points in their methods which are explosive on large datasets.

\begin{figure}[t]
\begin{center}
\includegraphics[width=0.9\linewidth]{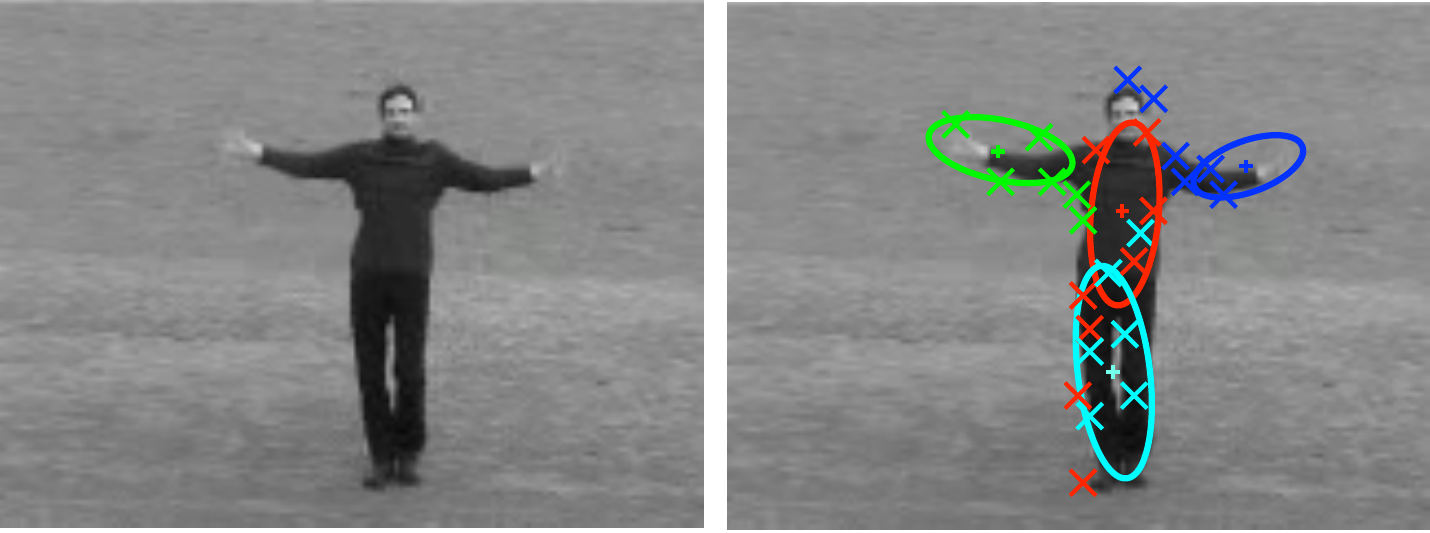}
\end{center}
\caption{Example of body parts detected by the constellation model in \cite{NieblesCVPR2007}. Revised based on the original figure in \cite{NieblesCVPR2007}.}
\label{fig:parts}
\end{figure}

\paragraph{Part-based Approaches}
Human bodies are structured objects, and thus it is straightforward to model human actions using motion information from body parts. Part-based approaches consider motion information from both the entire human body as well as body parts. The benefit of this line of approaches is it inherently captures the geometric relationships between body parts, which is an important cue for distinguishing human actions. A constellation model was proposed in \cite{FantiCVPR2005}, which models the position, appearance and velocity of body parts. Inspired by \cite{FantiCVPR2005}, a part-based hierarchical model was presented in \cite{NieblesCVPR2007}, in which a part is generated by the model hypothesis and local visual words are generated from a body part (see Figure~\ref{fig:parts}). 

The method in \cite{WongCVPR2007} considers local visual words as parts, and models the structure information between parts. This work was further extended in \cite{NieblesIJCV2008}, where the authors assume an action is generated from a multinomial distribution, and then each visual word is generated from distribution conditioned on the action. These part-based generated models were further improved by discriminative models for better classification performance \cite{WangNIPS2008,WangPAMI2010}. In \cite{WangNIPS2008,WangPAMI2010}, a part is considered as a hidden variable in their models. It is corresponding to a salient region with the most positive energy.

\paragraph{Manifold Learning Approaches}
Human action videos can be described by temporally variational human silhouettes. However, the representation of these silhouettes is usually high-dimensional and prevents us from efficient action recognition. To solve this problem, manifold learning approaches were proposed in  \cite{WangCVPR2007,JiaCVPR2008} to reduce the dimensionality of silhouette representation and embed them on nonlinear low-dimensional dynamic shape manifolds. The method in \cite{WangCVPR2007} adopts kernel PCA to perform dimensionality reduction, and discover the nonlinear structure of actions in the manifold. Then, a two-chain factorized CRF model is used to classify silhouette features in the low-dimensional space into human actions. A novel manifold embedding method was presented in \cite{JiaCVPR2008}, which finds the optimal embedding that maximizes the principal angles between temporal subspaces associated with silhouettes of different classes. Although these methods tend to achieve very high performance in action recognition, they heavily rely on clean human silhouettes which could be difficult to obtain in real-world scenarios.

\paragraph{Mid-Level Feature Approaches}
Bag-of-words models have shown to be robust to background noise but may not be expressive enough to describe actions in the presence of large appearance and pose variations. In addition, they may not well represent actions due to the large semantic gap between low-level features and high-level actions. To address these two problems, hierarchical approaches \cite{WangPAMI2010,ChoiCVPR2011,LiuCVPR2011,KongPAMI2014} are proposed to learn an additional layer of representations, and expect to better abstract the low-level features for classification. 

\begin{figure}[!t]
\begin{center}
\includegraphics[width=1\linewidth]{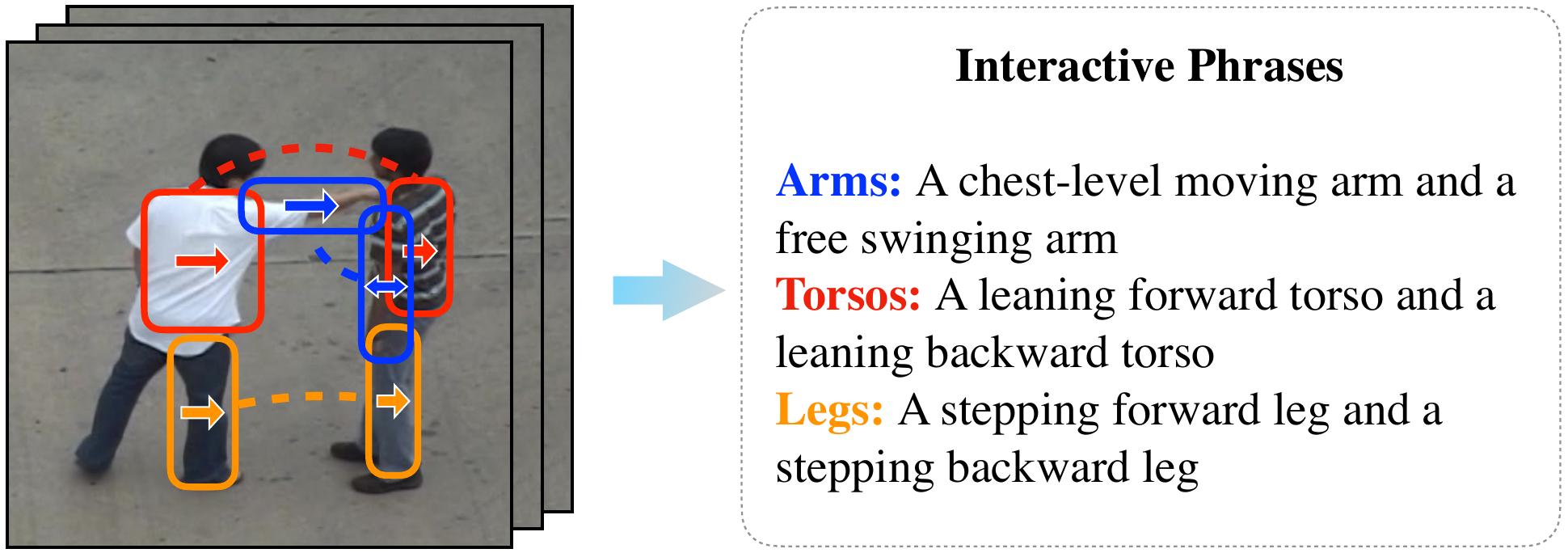}
\end{center}
\caption{Interaction recognition by learning semantic descriptions from videos. Revised based on the original figure in \cite{KongPAMI2014}.}
\label{fig:tpami2014}
\end{figure}

Hierarchical approaches learn mid-level features from low-level features, which are then used in the recognition task. The learned mid-level features can be considered as knowledge discovered from the same database used for training or being specified by experts. Recently, semantic descriptions or attributes (see Figure\ref{fig:tpami2014}) are popularly investigated in action recognition. These semantics are defined and further introduced into the activity classifiers in order to characterize complex human actions \cite{KongECCV2012,KongPAMI2014,LiuCVPR2011}. Other hierarchical approaches such as \cite{RaptisCVPR2013,Vahdat2011} select key poses from observed frames, which also learn better action representations during model learning. These approaches have shown superior results due to the use of human knowledge, but require extra annotations which is labor-intensive.

\paragraph{Feature Fusion Approaches}
Fusing multiple types of features from videos is a popular and effective way for action recognition. Since these features are generated from the same visual inputs, they are inter-related. However, the inter-relationship is complicated and is usually ignored in the existing fusion approaches. This problem was addressed in \cite{luo2014learning}, in which the maximum margin distance learning method is used to combine global temporal dynamics and local visual spatio-temporal appearance features for human action recognition. A Multi-Task Sparse Learning (MTSL) model was presented in \cite{yuan2013multi-task} to fuse multiple features for action recognition. They assume multiple learning tasks share priors, one for each type of features, and exploit the correlations between tasks to better fuse multiple features. A multi-feature max-margin hierarchical Bayesian model (M3HBM) was proposed in \cite{yang2015multi-feature} to learn a high-level representation by combining a hierarchical generative model (HGM) and discriminative max-margin classifiers in a unified Bayesian framework. HGM represents actions by distributions over latent spatial temporal patterns (STPs) learned from multiple feature modalities.
This work was further extended in \cite{yuan2016fusing} to combine spatial interest points with context-aware kernels for action recognition. Specifically, a video set is modeled as an optimized probabilistic hypergraph, and a robust context-aware kernel is used to measure high order relationships among videos.

\subsubsection{Classifiers for Human Interactions}
Human interaction is typical in daily life. Recognizing human interactions focuses on the actions performed by multiple people, such as ``handshake'', ``talking'', etc. Even though some of the early work such as \cite{LaptevCVPR2008,RyooICCV2009,YuBMVC2010,MarszalekCVPR2009,LiuCVPR2009} used action videos containing human interactions, they recognize actions in the same way as single-person action recognition. Specifically, interactions are treated as a whole and are represented as a motion descriptor including all the people in a video. Then an action classifier such as a linear support vector machine is adopted to classify interactions. Despite reasonable performance has been achieved, these approaches do not explicitly consider the intrinsic methods of interactions, and fail to consider the co-occurrence information between interacting people. Furthermore, they do not extract the motion of each person from the group, and thus their methods can not infer the action label of each interacting person.

Action co-occurrence of individual person is a piece of valuable information in human interaction recognition. In \cite{OliverPAMI2000}, action co-occurrence is captured by coupling motion state of one person with the other interaction person. Human interactions such as ``hug'', ``push'', and ``hi-five'' usually involve frequent close physical contact, and thus some body parts may be occluded. To robustly find body parts, Ryoo and Aggarwal \cite{RyooCVPR2006} utilized body part tracker to extract each individual in videos and then applied context-free grammar to model spatial and temporal relationships between people. A human detector is adopted in \cite{PerezPAMI2012} to localize each individual. Spatial relationships between individuals are captured using the structured learning technique \cite{FelzenszwalbCVPR2008}. Spatiotemporal context of a group of people including human pose, velocity and spatiotemporal distribution of individuals is captured in \cite{ChoiCVPR2011} to recognize human interactions. Their method shows promising results on collective actions without close physical contact such as ``crossing the road'', ``talking'', or ``waiting''. They further extended their work that can simultaneously track and recognize human interactions \cite{ChoiECCV2012}. A hierarchical representation of interactions is proposed in \cite{ChoiECCV2012} that models atomic action, interaction, and collective action. The method in \cite{LanPAMI2012} also utilizes the idea of hierarchical representation, and studies the collective activity recognition problem using crowd context. Different from these methods, the work in \cite{Vahdat2011} represents individuals in interactions as a set of key poses, and models spatial and temporal relationships of the key poses for interaction recognition. In our earlier work \cite{KongPAMI2014,KongECCV2012}, a semantic description-based approach is proposed to represent complex human interactions by learned motion relationships (see Figure~\ref{fig:tpami2014}). Instead of directly modeling action co-occurrence, we propose to learn phrases that describe the motion relationships between body parts. This will describe complex interactions in more details, and introduce human knowledge into the model. All these methods may not perform well in interactions with close physical contact due to the ambiguities in feature-to-person assignments. To address this problem, a patch-aware model was proposed in \cite{KongECCVW2014} to learn discriminative patches for interaction recognition, and determine the assignments at a patch level.




\subsubsection{Classifiers for RGB-D Videos}
Action recognition from RGB-D videos has been receiving a lot of attentions \cite{WangECCV2012,WangJCVPR2012,HadfieldCVPR2013,XiaCVPR2013,LiuIJCAI2013,OreifejCVPR2013} due to the advent of the cost-effective Kinect sensor \cite{ShottonPAMI2013}. RGB-D videos provide an additional depth channel compared with conventional RGB videos, allowing us to capture 3D structural information that is very useful in reducing background noise and simplifying intra-class motion variations \cite{Ni2011,WangCVPR2012,OreifejCVPR2013,HadfieldCVPR2013,Ofli2013}. 

Effective features have been proposed for the recognition task using depth data, such as histogram of oriented 4D normals \cite{OreifejCVPR2013,YangCVPR2014} and depth spatiotemporal interest points \cite{XiaCVPR2013,HadfieldCVPR2013}. Features from depth sequences can be encoded by \cite{LuoICCV2013}, or be used to build actionlets \cite{WangCVPR2012} for recognition. An efficient binary range-sample feature for depth data was proposed in \cite{LuCVPR2014}. This binary depth feature is fast, and has shown to be invariant to changes in scale, viewpoint, and background. 
The work in \cite{SungICRA2012,KoppulaICML2013} built layered action graph structures to model actions and subactions in a RGB-D video. Recent work \cite{LiuIJCAI2013} also showed that features of RGB-D data can be learned using deep learning techniques.

The methods in \cite{LiCVPRW2010,OreifejCVPR2013,YangCVPR2014,HadfieldCVPR2013,WangECCV2012,LuoICCV2013} only use depth data, and thus would fail if depth data were missing. Joint use of both RGB and depth data for action recognition is investigated in \cite{HuCVPR2015,JiaMM2014,LinCVPR2014,LiuIJCAI2013,WangCVPR2012,KongCVPR2015}. However, they only learn features shared between the two modalities and do not learn modality-specific or private features. To address this problem, shared features and privates features are jointly learned in \cite{KongIJCV2017}, which learns extra discriminative information for classification, and demonstrate superior performance than \cite{HuCVPR2015,JiaMM2014,LinCVPR2014,LiuIJCAI2013,WangCVPR2012,KongCVPR2015}. The methods in \cite{KongCVPR2015,KongIJCV2017} also show that they can achieve high recognition performance even though one modality is missing in training or testing. 

Auxiliary information has also shown to be useful in RGB-D action recognition. Skeleton data provided by a Kinect sensor was used in \cite{HuCVPR2015,WangCVPR2012,KongIJCV2017}, and has shown to be very effective in action recognition. The method in \cite{HuCVPR2015} learns a shared feature space for various types of features including skeleton features and local HOG features, and project these features onto the shared space for action recognition.
Different from this work, the method in \cite{KongIJCV2017} jointly learns RGB-D and skeleton features and action classifiers. The projection matrices in \cite{KongIJCV2017} are learned by minimizing the noise after projection and classification error using the projected features. Using auxiliary databases to improve the recognition performance was studied in \cite{JiaMM2014,LinCVPR2014}, in which actions are assumed to be reconstructed by entries in the auxiliary databases.

\subsection{Deep Architectures}
Although great success has been made by global and local features, these hand-crafted features require heavy human labor and domain expert knowledge to develop effective feature extraction methods. In addition, they normally do not generalize very well on large datasets. In recent years, feature learning using deep learning techniques has been receiving increasing attention due to their capability of learning powerful features that can be generalized very well \cite{JiTPAMI2013,TranICCV2015,DonahueCVPR2015,SimonyanNIPS2014}. The success of deep networks in action recognition can also be attributed to scaling up the networks to tens of millions of parameters and massive labeled datasets. Recent deep networks \cite{VarolTPAMI2017,TranICCV2015,FeichtenhoferCVPR2017,KarCVPR2017} have achieved surprisingly high recognition performance on a variety of action datasets.

Action features learned by deep learning techniques has been popularly investigated \cite{YangECCV2012,WangMM2014,TaylorECCV2010,SunCVPR2014,PlotzIJCAI2011,LeCVPR2011,KarpathyCVPR2014,JiTPAMI2013,JiICML2010,HasanECCV2014,BengioPAMI2013,SimonyanNIPS2014} in recent years. The two major variables in developing deep networks for action recognition are the convolution operation and temporal modeling, leading to a few lines of networks. 

\begin{figure}[!t]
\begin{center}
\subfigure[2D convolution]{\includegraphics[width=0.48\linewidth]{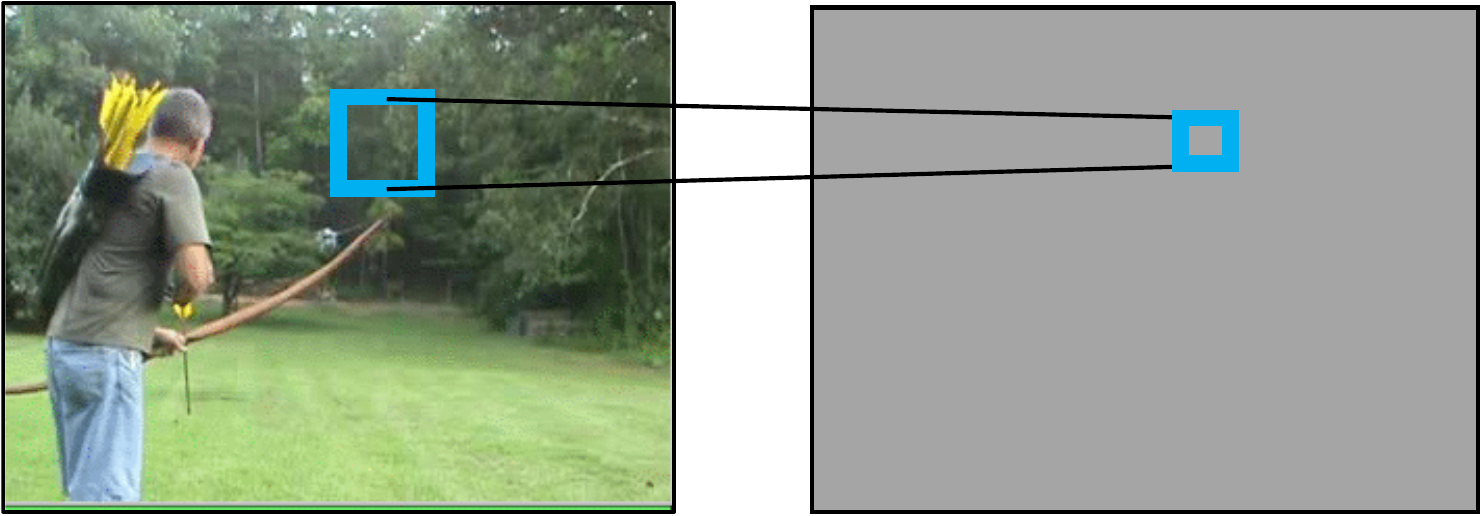}}
\subfigure[3D convolution]{\includegraphics[width=0.48\linewidth]{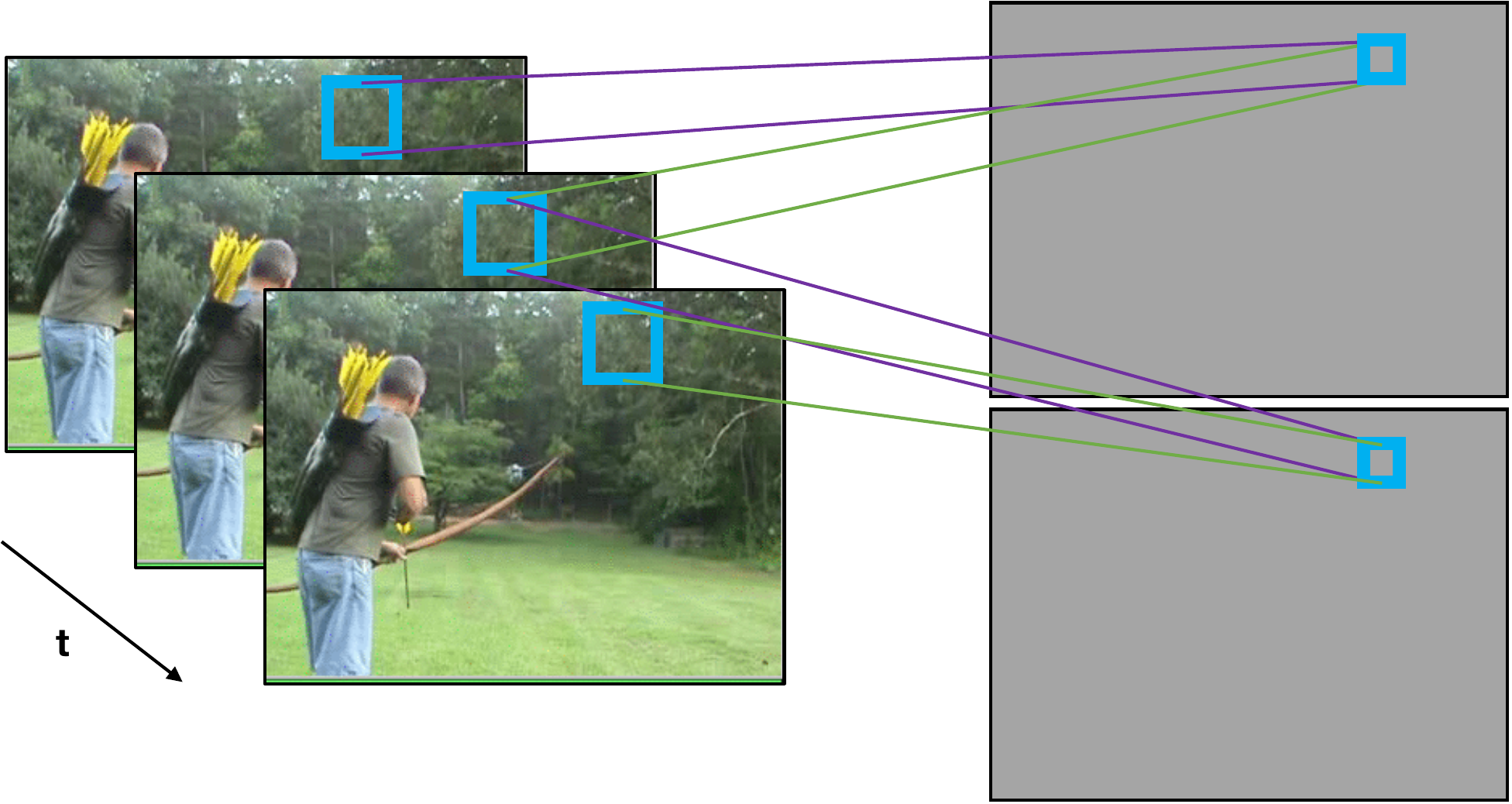}}
\end{center}
\caption{Illustration of (a) 2D convolution and (b) 3D convolution.}
\label{fig:2d3dconv}
\end{figure}
The convolution operation is one of the fundamental components in deep networks for action recognition, which aggregates pixel values in a small spatial (or spatiotemporal) neighborhood using a kernel matrix. \textbf{2D vs 3D Convolution:} 2D convolution over images (Figure~\ref{fig:2d3dconv}(a)) is one of the basic operation in deep networks, and thus it is straightforward to use 2D convolution on video frames. The work in \cite{KarpathyCVPR2014} presented a single-frame architecture based on a 2D CNN model, and extracted a feature vector for each frame. Such a 2D convolution network (2D ConvNet) also enjoys the benefit of using the networks pre-trained on large-scale image datasets such as ImageNet. However, 2D ConvNets do not inherently model temporal information, and requires an additional aggregation or modeling of such information.

As multiple frames are presenting in videos, 3D convolution (Figure~\ref{fig:2d3dconv}(b)) is more intuitive to capture temporal dynamics in a short period of time. Using 3D convolution, 3D convolutional networks (3D ConvNets) directly create hierarchical representations of spatio-temporal data \cite{JiICML2010,JiTPAMI2013,TaylorECCV2010,TranICCV2015}. However, the issue is they have many more parameters than 2D ConvNets, making them hard to train. In addition, they are prevented from enjoying the benefits of ImageNet pre-training.


Another key variable in designing deep networks is \textbf{Temporal Modeling}.
Generally, there are roughly three methods in temporal modeling. One straightforward way is to directly apply 3D convolution to several consecutive frames \cite{JiICML2010,JiTPAMI2013,TaylorECCV2010,TranICCV2015,CarreiraCVPR2017}. As a result, the temporal dimension in the 3D convolution kernel will capture the temporal dynamics in these frames. One of the limitations of these approaches is they may not be able to reuse the 2D ConvNets pre-trained on large-scale image datasets. Another line of approaches model temporal dynamics by using multiple streams \cite{SimonyanNIPS2014,FeichtenhoferCVPR2016,CarreiraCVPR2017,GirdharCVPR2017,KarCVPR2017}. A stream named flow net in the networks trains on optical flow frames, which essentially capture motion information in the adjacent two frames. However, these approaches largely disregard the long-term temporal structure of videos. 2D convolution is usually used in these approaches, and thus they can easily exploit the new ultra-deep architectures and models pre-trained for still images. The third category of approaches uses temporal pooling \cite{KarCVPR2017,GirdharCVPR2017} or aggregation to capture temporal information in a video. The aggregation can be performed by using a LSTM model on top of 2D ConvNets \cite{DonahueCVPR2015,NgCVPR2015}.



%

\subsubsection{Space-time Networks}
Space-time networks are straightforward extensions of 2D ConvNets as they capture temporal information using 3D convolutions. 

The method in \cite{JiICML2010} was one of the pioneering works in using convolution neural networks (CNN) for action recognition. They perform 3D convolutions over adjacent frames, and thus extract features from both spatial and temporal dimensions. Their 3D CNN network architecture starts with $5$ hardwired kernels including gray, gradient-x, gradient-y, optflow-x, and optflow-y, resulting in $33$ feature maps. Then the network repeats 3D convolution and subsampling, and uses a fully-connected layer to generate a 128-dimensional feature vector for action classification. In a later extension \cite{JiTPAMI2013}, the authors regularized the network to encode long-term action information by encouraging the network to learn feature vector close to high-level motion features such as the bag-of-words representation of SIFT features.

\begin{figure}[!t]
\begin{center}
\includegraphics[width=0.9\linewidth]{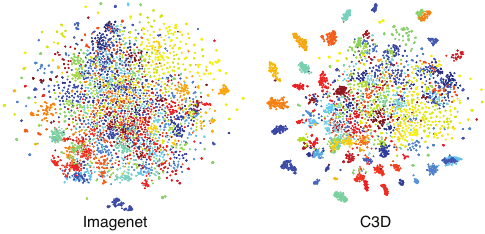}
\end{center}
\caption{Feature embedding by Imagenet and C3D. C3D features show better class separation than Imagenet, indicating its capability in learning better features for videos. Originally shown in \cite{TranICCV2015}.}
\label{fig:embedding}
\end{figure}
The 3D ConvNet \cite{JiICML2010,JiTPAMI2013} was later extended to a modern deep architecture called C3D \cite{TranICCV2015} that learns on large-scale datasets. The C3D network contains $5$ convolution layers, $5$ max-pooling layers, $2$ fully-connected layers, and a softmax loss layer, subject to the machine memory limit and computation affordability. Their work demonstrated that C3D learns a better feature embedding for videos (see Figure~\ref{fig:embedding}). Results showed that the C3D method with a linear classifier can outperform or approach the state-of-the-art methods on a variety of video analysis benchmarks including action recognition and object recognition.

Still, 3D ConvNets \cite{JiICML2010,JiTPAMI2013,TranICCV2015} for action recognition are relatively shallow with up to $8$ layers. To further improve the generalization power of 3D ConvNets, \cite{CarreiraCVPR2017} inflated very deep networks for image classification into spatio-temporal feature extractors by repeating 2D filters along the time dimension, allowing the network to reuse 2D filters pre-trained on ImageNet. This work also shows that pre-training on the Kinetics dataset achieves better recognition accuracy on UCF-101 and HMDB51 datasets. Another solution to build a deep 3D ConvNet was proposed in \cite{QiuICCV2017}, which uses a combination of one $1\times 3\times 3$ convolutional layer and one $3\times 1\times 1$ convolutions to take the place of a standard 3D convolution. 

One limitation of 3D ConvNets is that they typically consider very short temporal intervals, such as 16 frames in \cite{TranICCV2015}, thereby failing to capture long-term temporal information. To address this problem,  \cite{VarolTPAMI2017} increases the temporal extent in the 3D convolutions, and empirically shows that they can significantly improve the recognition performance.


\subsubsection{Multi-Stream Networks}

\begin{figure}[!t]
\begin{center}
\includegraphics[width=0.9\linewidth]{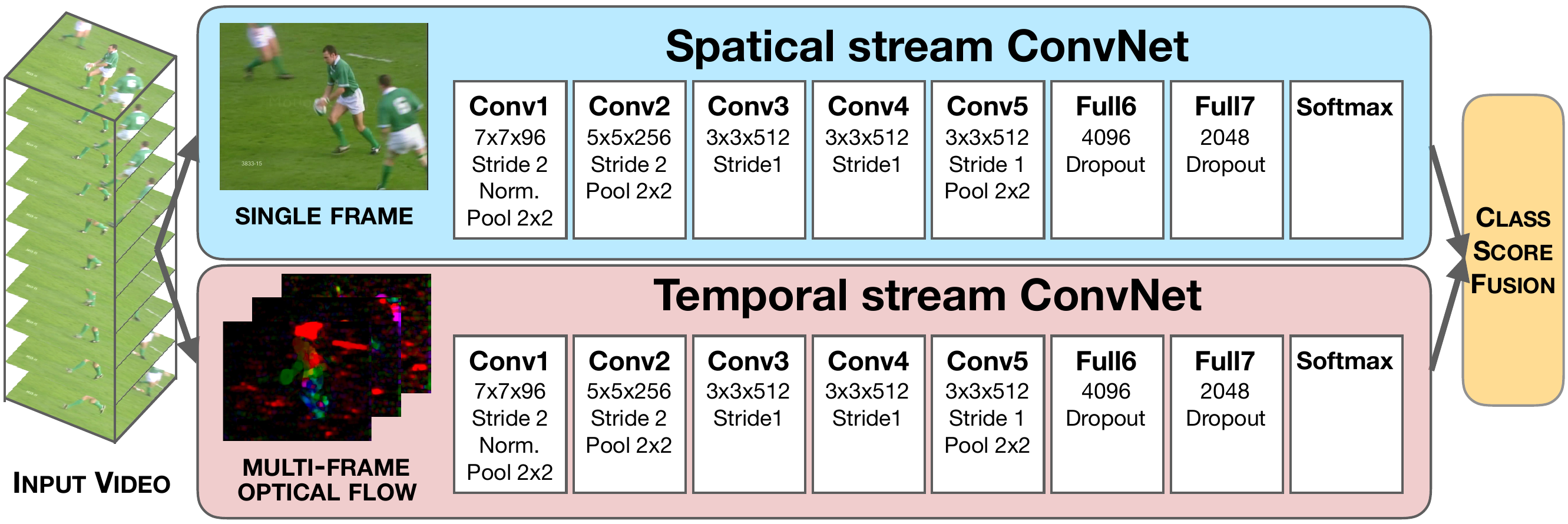}
\end{center}
\caption{Two-stream network proposed in \cite{SimonyanNIPS2014} contains a spatial network and a temporal network, which are used for modeling static information in still frames and motion information in optical flow images, respectively. Revised based on the original figure in \cite{SimonyanNIPS2014}.}
\label{fig:twostream}
\end{figure}

Multi-stream networks utilize multiple convolutional networks to model both appearance and motion information in action videos. Even though the network in \cite{KarpathyCVPR2014} achieved great success, its results were significantly worse than those of the best hand-crafted shallow representations \cite{WangIJCV2015,WangIJCV2013}. To address this problem, a successful work by \cite{SimonyanNIPS2014} explored a new architecture related to the two-stream hypothesis \cite{Goodale1992}. Their architecture contains two separate streams, a spatial ConvNet and a temporal ConvNet (see Figure~\ref{fig:twostream}). The former one learns actions from still images, and the latter one performs recognition based on the optical flow field.

The two-stream network \cite{SimonyanNIPS2014} directly fuses the outputs of the two streams generated by their respective softmax function, which may not be appropriate for gathering information over a long period of time. An improvement was proposed in \cite{WangCVPR2015}, which used the two-stream network to obtain multi-scale convolutional feature maps, and pooled the feature maps together with the detected trajectories to compute ConvNet responses centered at the trajectories. Such a scheme encodes deep features into effective descriptors constrained by sampled trajectories. Temporal feature pooling in the two-stream network was investigated in \cite{NgCVPR2015}, which is capable of making video-level predictions after the pooling layer. The work in \cite{GirdharCVPR2017} also presented a novel pooling layer named ActionVLAD that aggregates convolutional feature descriptors in different image portions and temporal spans. They also used ActionVLAD to combine appearance and motion streams together. The network named temporal linear encoding \cite{DibaCVPR2017} aggregates temporal features sampled from a video, and then projects onto a low-dimensional feature space. By doing so, long-range temporal structure in different frames can be captured and be encoded into a compact representation. AdaScan proposed in \cite{KarCVPR2017} evaluated the importance of the next frame, so that only informative frames will be pooled, and non-informative frames will be disregarded in the video-level representation. Their AdaScan method uses a multilayer perceptron to compute the importance for the next frame given temporally pooled features up to the current frame. The importance score will then be used as a weight for the feature pooling operation for aggregating the next frame. Despite effective, most of the feature encoding methods lack of considering spatio-temporal information. To address this problem, the work in \cite{DutaCVPR2017} proposed a new feature encoding method for deep features. More specifically, they proposed locally max-pooling that groups features according to their similarity and then performs max-pooling. In addition, they performed max-pooling and sum-pooling over the positions of features to achieve spatio-temporal encoding. 

Temporal sampling in the two-stream network was proposed in Temporal Segment Networks (TSN) \cite{WangECCV2016}. In TSN long-range dynamics are gathered by analyzing short video snippets formed from randomly sampled frames from segments of the full video.  The idea here is that directly analyzing densely sampled video sequence makes no sense since the consecutive frames in the video contain a lot of redundancy. Moreover, some actions reveal them-self at different temporal scales, such as sprinting, which requires multiple actions over a long span of time, compared to just crouching. The original TSN network \cite{WangECCV2016}, was based on two-stream architecture from \cite{SimonyanNIPS2014}. The prediction from temporal segments was summaries by applying consensus function to frame features extracted with pre-trained Deep CNN classification network. As for consensus function was used a simple pooling operation. The advantage of this network is that it can enjoy the benefits of using big pre-trained classification networks for feature extraction. To improve the performance of temporal sampling in \cite{zhou2018temporal} was suggested to perform sampling at different temporal scales, and substitute pooling operation with a fully connected network, which should encode the temporal ordering of frames. The TSN can be also incorporated into another action recognition frameworks as illustrated in \cite{qiu2019learning}. Recently, Liu \etal{}~\cite{no_frames} attempted to use all video frames for classification by clustering the activations along the temporal dimension based on the assumption that similar frames should have similar activation values. However, this method is limited in its ability of dynamically selecting the number of clusters. Wang \etal{}~\cite{tdn} proposed Temporal Difference Network (TDN) which aims to recognize actions from the entire video. TDN contains short-term temporal difference modules to encode local motion information and long-term temporal difference modules to capture motion across segments.

One of the major problems in the two-stream networks \cite{SimonyanNIPS2014,WangCVPR2015,NgCVPR2015} is that they do not allow interactions between the two streams. However, such an interaction is really important for learning spatiotemporal features. To address this problem, Feichtenhofer \etal \cite{FeichtenhoferCVPR2016} proposed a series of spatial fusion functions that make channel responses at the same pixel position be in the same correspondence. These fusion layers are placed in the middle of the two-streams allowing interactions between them. They further injected residual connections between the two streams \cite{FeichtenhoferNIPS2016,FeichtenhoferCVPR2017}, and allow a stream to be multiplicatively scaled by the opposing stream's input \cite{FeichtenhoferCVPR2017}. Such a strategy bridges the gap between the two streams, and allows information transfer in learning spatiotemporal features.

\begin{figure}[t]
\begin{center}
\includegraphics[width=1\linewidth]{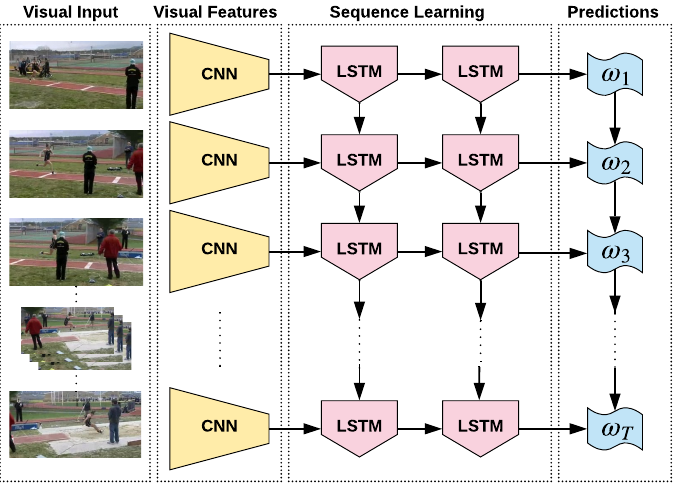}
\end{center}
\caption{Network architecture of LRCN \cite{DonahueCVPR2015} with a hybrid of ConvNets and LSTMs. Revised based on the original figure in \cite{DonahueCVPR2015}.}
\label{fig:lrcn}
\end{figure}
\subsubsection{Hybrid Networks}
Another solution to aggregate temporal information is to add a recurrent layer on top of the CNNs, such as LSTMs, to build hybrid networks \cite{DonahueCVPR2015,NgCVPR2015}. Such hybrid networks take the advantages of both CNNs and LSTMs, and thus have shown promising results in capturing spatial motion patterns, temporal orderings and long-range dependencies \cite{WangCVPR2015,DibaCVPR2017,KarCVPR2017}. 

Donahue \etal \cite{DonahueCVPR2015} explored the use of LSTM in modeling time series of frame features generated by 2D ConvNets. As shown in Figure~\ref{fig:lrcn}, the recurrence nature of LSTMs allows their network to generate textual descriptions of variable lengths, and recognize human actions in the videos.  Ng \etal \cite{NgCVPR2015} compared temporal pooling and using LSTM on top of CNNs. They discussed six types of temporal pooling methods including slow pooling and Conv pooling, and empirically showed that adding a LSTM layer generally outperforms temporal pooling by a small margin because it capture the temporal orderings of the frames. A hybrid network using CNNs and LSTMs was proposed in \cite{WuMM2015}. They used two-stream CNN \cite{SimonyanNIPS2014} to extract motion features from video frames, and then fed into a bi-directional LSTM to model long-term temporal dependencies. A regularized fusion scheme was proposed in order to capture the correlations between appearance and motion features. 

Hybrid networks have also been applied to skeleton-based action recognition. Skeleton data can be easily obtained by depth sensors such as Kinect or pose estimation algorithms. In these methods, hybrid deep neural networks \cite{Shahroudy_2016_NTURGBD,zhu2016co,liu2016spatio,ke2017new,yan2018spatial} are developed to model the structure information of various body joints as well as temporal information of body movement. Recurrent neural networks are widely used to capture the features consisting of ordered joints ~\cite{Shahroudy_2016_NTURGBD,zhu2016co,liu2016spatio}. Temporal CNN~\cite{ke2017new} is also applied to capture the features of structured body joints. Recently, graph convolution networks have shown superior performance over RNNs and Temporal CNNs, and become the backbone for capturing the structural information of joints. Yan \etal~\cite{yan2018spatial} proposed a spatio-temporal graph convolution to learn the structural and temporal information at the same time. Si \etal~\cite{si2019attention} applied GCN-LSTM to model the temporal dependencies of skeleton and proposed an attention model to learn the importance of each joint.

\begin{table*}[]
\centering
\caption{Pros and cons of action recognition approaches.}
\label{tab:comp}
\begin{tabular}{|l|l|l|l|}
\hline
                     & Approaches            & Pros                                                  & Cons                                                           \\
\hline
\multirow{7}{*}{\rotatebox{90}{Shallow}} & Direct \cite{SchuldtICPR2004,WangICCV2013}  & Easy and quick to use.                                & Performance is limited.                     \\
                         & Sequential \cite{MorencyCVPR2007,ShiIJCV2011}& Models temporal evolution.                            & Sensitive to noise.      \\
                         & Space-time \cite{wu2014human,WuCVPR2011}& Captures spatiotemporal structures.                   & Limited to small datasets.                                     \\
                         & Part-based \cite{WangPAMI2010,NieblesECCV2010}& Models body parts at a finer level.                   & Limited to small datasets.                     \\
                         & Manifold \cite{WangCVPR2007,JiaCVPR2008}     & Tend to achieve high performance.                     & Rely on human silhouettes.                  \\
                         & Mid-level feature \cite{ChoiCVPR2011,LiuCVPR2011}    & Introduce knowledge to models.              & Require extra annotations.                        \\
                         & Feature fusion \cite{yuan2013multi-task,yuan2016fusing}       & Tend to achieve high performance.                     & Slow in feature extraction.               \\\hline
\multirow{3}{*}{\rotatebox{90}{Deep}}    & Space-time \cite{JiTPAMI2013,TranICCV2015}  & Natural extension of 2D convolution.                  & Short temporal interval. \\
                         & Multi-stream \cite{SimonyanNIPS2014,FeichtenhoferCVPR2017}& Able to use pre-trained 2D ConvNets.                  & Int. b/w networks is difficult.                     \\
                         & Hybrid \cite{DonahueCVPR2015,NgCVPR2015}      & Easy to build using existing networks. & Difficult to fine-tune.\\
\hline
\end{tabular}
\end{table*}

\subsection{Learning with Limited Data/Label}
Due to the necessity of training deep neural networks, recent video are becoming extremely large. For example, Youtube-8M dataset \cite{abu2016youtube} consists of over $8$ million videos. For such large-scale datasets, it is expensive and almost impossible to annotate all the video data. Even though search engines were given action labels and were used to retrieve videos, they also make mistakes and thus the compiled video data could be noisy. One solution is to learn action models in a weakly-supervised fashion or an unsupervised fashion. Therefore, the models do not necessarily require fully-annotated video data and can learn under very limited or no supervisory signals. Few-shot learning was also recently introduced to learn in the low-sample regime.

\subsubsection{Weakly-supervised Action Learning}
Weakly-supervised learning methods \cite{laptev2008learning,bojanowski2014weakly,ghadiyaram2019large} are developed to deal with the scenarios where each of the videos is not fully annotated. One promising application scenario is to understand human actions in untrimmed videos, in which the temporal boundaries of various actions in the videos are not annotated. Such a learning capability enhances most of the existing action recognition methods \cite{TranICCV2015,KongTPAMI2018,SimonyanNIPS2014,WangCVPR2015,NgCVPR2015}, as require all the action videos to be trimmed which is expensive and time-consuming to achieve.

Movie with script data is a typical scenario to evaluate weakly-supervised action learning methods. Pioneering work made by Laptev \etal~\cite{laptev2008learning} presented a novel realistic action dataset from movies. Annotations were made using movie scripts. Duchenel \etal~\cite{duchenne2009automatic} followed this work and addressed the problem of weakly-supervised learning of action models and localizing action instances in videos given the corresponding movie scripts. 

Another type of work is weak-supervised action understanding given a temporally ordered list of action classes that will appear in the video. For example, Bojanowski \etal~\cite{bojanowski2014weakly} formulated the problem as a weakly supervised temporal assignment and proposed a clustering method that assigns the action labels to the temporal segments in videos. Huang \etal~\cite{huang2016connectionist} adapted the Connectionist Temporal classification model from speech recognition to perform weakly-supervised action labeling.

Recent works have extended weakly-supervised action representation learning to untrimmed videos with unordered action lists. Wang \etal~\cite{WangCVPR2017} proposed the UntrimmedNet for untrimmed video understanding by learning action models and reasoning temporal duration of action instances in an end-to-end framework. Ghadiyaram \etal ~\cite{ghadiyaram2019large} took advantage of large-scale noisy labeled web videos to learn a pre-trained model for video action recognition. 

\subsubsection{Unsupervised and Self-supervised Action Learning}
Unsupervised or self-supervised representation learning is becoming popular in recent years as it allows deep neural networks to be pre-trained utilizing the supervisory signals within the training data, rather than given by humans. Such pre-trained models can be beneficial for downstream tasks, such as action recognition and localization. Many attempts have leveraged the temporal coherence, motion consistency and temporal continuity as supervision, which will be discussed below.

The chronological order of frames is a typical free supervision signal for videos. Action models learn to tell whether the frame sequence is ordered or not, given either shuffled or unshuffled videos~\cite{misra2016shuffle,fernando2017self}. Another related task is training the model to tell the actual order of the shuffled video frames~\cite{lee2017unsupervised}. Xu \etal~\cite{xu2019self} extended the order prediction tasks from frames to clips. This helps to train a 3D CNN framework using chronological order supervision. Buchler \etal~\cite{buchler2018improving} applied deep reinforcement learning to sample new permutations according to their expected utility to adapts to the state of the network.

The motion of objects in videos can also be used as supervision. Wang \etal~\cite{wang2015unsupervised} found the corresponding pairs using visual tracking, based on Siamese-Triplet network. Purushwalkam \etal~\cite{purushwalkam2016pose} utilized pose as free supervision since similar pose should have similar motion.
Wang \etal~\cite{wang2017transitive} explored different self-supervised methods to learn the representations invariant to the variations between the object patches, which is extracted by motion cues.
Gan \etal~\cite{gan2018geometry} used geometry cues flow field and disparity maps to learn the video representations.

\subsubsection{Few-shot Learning}

Few-shot learning aims at learning reliable models from minimalist data sets. In extreme cases, there could be no training sample for some categories which is called the zero-shot learning. Majority of few-shot works target at recognising images, while only a few address the video action recognition challenge. \cite{CMN_few_shot} proposed a compound memory network (CMN) which predicts the unseen video by retrieving a similar video stored in the memory of the CMN architecture. ProtoGAN \cite{protoGAN} learns the class-prototype vectors through a feature aggregator network called \textit{Class Prototype Transfer Network} (CPTN), then generates additional video features for the recognition classifier. Neural Graph Matching (NGM) network \cite{NGMN_few_shot} is a graph-based approach that generates graph representations for 3D action videos and match unseen videos and seen videos by the similarity of their graph representations. \cite{ashish_few_shot} proposes a framework for zero-shot action recognition which models each action class as a probability distribution and the distribution parameters are a linear combination of the attributes of the action class. The weights of the attributes are learnt from the labeled samples. One challenge in few-shot action recognition is the variation of temporal lengths. Temporal Attentive Relation Network (TARN) \cite{TARN} uses attention modules to align video segments and learns a distance measure between the aligned representations for few-shot and zero-shot learning. Action Relation Network (ARN) \cite{ARN_few_shot} encodes the video clips features of the query set and support set into a Power Normalized Autocorrelation Matrix (AM) from which a relation network learns to captures the relations. Similar to ARN, Ordered Temporal Alignment Module (OTAM) \cite{OTAM_few_shot} extracts per-frame feature through an embedding network, then computes an alignment score of the distance matrix. Temporal-Relational CrossTransformers (TRX) \cite{TRX} classifies the query video by matching each sub-sequence to all sub-sequences in the support set using CrossTransformer attention modules.

\subsection{Summary}
Deep networks are dominant in action recognition research but shallow methods are still useful. Compared with deep networks, shallow methods are easy to train, and generally perform well on small datasets. Recent shallow methods such as improved dense trajectory with linear SVM \cite{WangICCV2013} have also shown promising results on large datasets, and thus they are still popularly used recently in the comparison with deep networks \cite{TranICCV2015,VarolTPAMI2017,FeichtenhoferCVPR2017}. It would be helpful to use shallow approaches first if the datasets are small, or each video exhibits complex structures that need to be modeled. However, there are lots of pre-trained deep networks on the Internet such as C3D \cite{TranICCV2015} and TSN \cite{WangECCV2016} that can be easily employed. It would be also helpful to try these methods and fine-tune the models to particular datasets. Table~\ref{tab:comp} summarizes the pros and cons of action recognition approaches.

\section{Action Localization and Detection}
In order to recognize and predict an action, the machine needs to know where is the action in a video. This is achieved by action localization and detection, which find out the spatiotemporal regions containing certain human actions in videos. Both of the two tasks have attracted a large amount of research in recent years.  As an analogy to object localization and detection in the image domain, action detection is additionally required to identify the action type of each action that occurs in the video compared to the action localization. Based on the feature learning paradigms, related work can be categorized into shallow and deep learning methods, for which we will make a comprehensive literature review. Table \ref{THUMOSresults} summarizes some recent detection methods and compares results on thresholds of 0.3, 0.4, and 0.5. The mAP@$\alpha$ denotes the mean Average Precision at different IOU threshold which measures the average prevision on each action category.

\begin{table}[]
\caption{Results of action detection methods on THUMOS'14 \cite{THUMOS14}. The mAP@$\alpha$ denotes the mean Average Precision at different threshold, $\alpha$. ``-'' indicates the result is not reported.}
\begin{tabular}{llll}
\hline
\multicolumn{1}{|l|}{}                                                                    & \multicolumn{1}{l|}{mAP@0.5} & \multicolumn{1}{l|}{mAP@0.4} & \multicolumn{1}{l|}{mAP@0.3} \\ \hline
\multicolumn{1}{|l|}{End-to-End \cite{yeung2016end}}                                                               & \multicolumn{1}{l|}{17.1}    & \multicolumn{1}{l|}{26.4}    & \multicolumn{1}{l|}{36.0}    \\ \hline
\multicolumn{1}{|l|}{Multi-stage \cite{ShouCVPR2016}}                                                                & \multicolumn{1}{l|}{19.0}    & \multicolumn{1}{l|}{28.7}    & \multicolumn{1}{l|}{36.3}    \\ \hline
\multicolumn{1}{|l|}{TURN \cite{GaoICCV2017}}                                                                & \multicolumn{1}{l|}{24.5}    & \multicolumn{1}{l|}{35.3}    & \multicolumn{1}{l|}{46.3}    \\ \hline
\multicolumn{1}{|l|}{\begin{tabular}[c]{@{}l@{}}Temporal Context \\Network \cite{Dai2017TemporalCN}\end{tabular}}                                                                 & \multicolumn{1}{l|}{25.6}    & \multicolumn{1}{l|}{33.3}    & \multicolumn{1}{l|}{-}       \\ \hline
\multicolumn{1}{|l|}{\begin{tabular}[c]{@{}l@{}}Single-stream \\ R-C3D \cite{xu2017r}\end{tabular}}      & \multicolumn{1}{l|}{28.9}    & \multicolumn{1}{l|}{35.6}    & \multicolumn{1}{l|}{44.8}    \\ \hline
\multicolumn{1}{|l|}{SSN \cite{ZhaoICCV2017}}                                                                 & \multicolumn{1}{l|}{29.8}    & \multicolumn{1}{l|}{41.0}    & \multicolumn{1}{l|}{51.9}    \\ \hline
\multicolumn{1}{|l|}{\begin{tabular}[c]{@{}l@{}}Single-stream \\ R-C3D+OHEM \cite{xu2019two}\end{tabular}} & \multicolumn{1}{l|}{35.8}    & \multicolumn{1}{l|}{43.1}    & \multicolumn{1}{l|}{51.1}    \\ \hline
\multicolumn{1}{|l|}{\begin{tabular}[c]{@{}l@{}}Two-stream \\ R-C3D \cite{xu2019two}\end{tabular}}         & \multicolumn{1}{l|}{36.1}    & \multicolumn{1}{l|}{43.0}    & \multicolumn{1}{l|}{51.2}    \\ \hline
\multicolumn{1}{|l|}{BSN \cite{lin2018bsn}}                                                                 & \multicolumn{1}{l|}{36.9}    & \multicolumn{1}{l|}{45.0}    & \multicolumn{1}{l|}{53.5}    \\ \hline
\multicolumn{1}{|l|}{MGG UNet \cite{liu2019multi}}                                                            & \multicolumn{1}{l|}{37.4}    & \multicolumn{1}{l|}{46.8}    & \multicolumn{1}{l|}{53.9}    \\ \hline
\multicolumn{1}{|l|}{BMN \cite{lin2019bmn}}                                                                 & \multicolumn{1}{l|}{38.8}    & \multicolumn{1}{l|}{47.4}    & \multicolumn{1}{l|}{56.0}    \\ \hline
                                                                                          &                              &                              &                             
\end{tabular}

\label{THUMOSresults}
\end{table}

\subsection{Shallow Approaches}
Early work~\cite{KaramanTHUMOS2014,WangTHUMOS2014} formulated action detection as a classification task by firstly using temporal segmentation or sliding window methods. In these work, the untrimmed video is segmented into short video clips and the multiple features are extracted for classifiers such as support vector machine (SVM) to recognize the action types. Eventually, the actions that appear in the video as well as their temporal locations are determined. Jian \etal{}~\cite{JainCVPR2014} proposed to generate a set of bounding boxes from the video which are called tubelets for action localization. However, these methods suffer from handcraft feature engineering and multi-stage model tuning, leading to quite inaccurate detection results. 

\subsection{Deep Architectures}
Recent approaches to action localization and detection make full use of deep neural networks for learning better video feature representation. To this end, Shou \etal{}~\cite{ShouCVPR2016} proposed to first generate action proposals from the long videos. Then, a localization network is introduced to fine-tune the trained action classification network to recognize the action labels. The idea of their action proposals inspired many later research~\cite{EscorciaECCV2016,ShouCVPR2017,WangCVPR2017,ZhaoICCV2017,xu2017r,GaoICCV2017,ChaoCVPR2018}. For these methods, Escorcia \etal{}~\cite{EscorciaECCV2016} proposed a deep action proposals (DAP) method which achieves high efficiency and demonstrates to have good generalization capability. To detect human actions in frame-level granularity, Shou \etal{}~\cite{ShouCVPR2017} proposed an end-to-end learning framework in which a CDC convolutional filter is designed on top of 3D ConvNet. To model the temporal structure of each action instance, Zhao \etal{}~\cite{ZhaoICCV2017} proposed a structured segment network (SSN) with a temporal pyramid and a dubbed temporal actionness grouping (TAG) model for action proposals generation. As the action detection is similar to the object detection, Chao~\etal{}~\cite{ChaoCVPR2018} revisited the most widely-used object detection method Faster R-CNN and propose a temporal action localization network (TAL-Net) to address the unsolved challenges, including the large variation of action durations, temporal context modeling, and multi-stream feature fusion. Song \etal{}~\cite{SongCVPR2019} noted that the ambiguous transition states of an action and long-term temporal context are critical for accurate action detection. Thus, they propose a transition-aware context network and it is demonstrated to be significantly effective for untrimmed video dataset. To modeling the relations among action proposals, Zeng \etal{}~\cite{ZengICCV2019} recently proposed to introduce the graph convolutional neural networks (GCN) for temporal action localization. Song \etal{}~\cite{SongTMM2019} introduced the action pattern tree (AP-Tree) in which the temporal information can be utilized. Inspired by the conventional idea of coarse-to-fine detection, Yang \etal{}~\cite{YangCVPR2019} proposed a spatio-temporal progressively learning method for video action detection, achieving remarkable performance on existing benchmarks. Recently, Xu \etal{}~\cite{XuICCV2019} raised the importance of online action detection and propose a temporal recurrent network (TRN) by simultaneously performing online action detection and anticipation, significantly outperforming the state-of-the-art. Chen \etal{}~\cite{woo} unified the tasks of actor localization and action classification into the same backbone, which reduces model complexity and improves efficiency compared to SOTA methods. Li \etal{}~\cite{three_birds} designed two auxiliary pretext
tasks to recycle the limited labeled data and benefit both features extraction as well as prediction.

Different from previous full-supervised methods that require large-scale frame-level annotations of action instances, weakly-supervised methods need only the video- or clip-level action annotations so that they are more promising in practice. Wang \etal{}~\cite{WangCVPR2017} proposed a weakly-supervised action detection model that is directly learned on the untrimmed video data, achieving performance on-par-with those of the full-supervised action detection methods. Recently, Yu \etal{}~\cite{YuICCV2019} introduced the temporal structure mining (TSM) approach to the weakly-supervised action detection problem. In their method, an action instance is modeled as a multi-phase process so that the phase filters can be utilized to compute the confidence score, indicating the action occurrence probability. For weakly-supervised action localization problem, it also attracts much attention in recent years. Gao \etal{}~\cite{gao2021woad} proposed a weakly supervised framework that consists of two modules, one module generates the pseudo ground truth of action boundaries which are used to supervise the action recognition module. Yang \etal{}~\cite{uncertainty_action_detection} proposed to incorporate the uncertainty for reducing the noise in the generated pseudo labels. To handle the challenge of limited temporal annotations, Yang \etal{}~\cite{YangCVPR2018} used an one-shot learning technique of matching network for temporal action localization. Narayan \etal{}~\cite{NarayanICCV2019} introduced a novel loss function comprising the action classification loss, multi-label center loss, and the counting loss, setting the new state-of-the-art on weakly-supervised action localization. 

In addition to using visual data, other data modalities such as skeleton and RGB-D data can also be utilized for temporal action localization and detection. To learn the features of discriminative skeleton joints, Song~\etal{}~\cite{SongTIP2018} introduced a spatio-temporal attention LSTM model for action recognition and detection. To handle the modality discrepancy in a multi-modal setting, Luo \etal{}~\cite{LuoECCV2018} proposed a graph distillation method that privileged information is learned from a large-scale multi-modal dataset in the source domain and their model can be effectively deployed to the modality-scarce target domain. For the continuous action stream scenario, Dawar \etal{}~\cite{DawarSensors2018} designed a multimodal fusion system to incorporate depth camera data and wearable inertial sensor signals for action detection.

\section{Action Prediction}
After-the-fact action recognition has been extensively studied in the last few decades, and fruitful results have been achieved. State-of-the-art methods \cite{DonahueCVPR2015,GirdharCVPR2017,WangECCV2016} are capable of accurately giving action labels after observing the entire action executions. However, in many real-world scenarios (\eg, vehicle accident and criminal activity), intelligent systems do not have the luxury of waiting for the entire video before having to react to the action contained in it. For example, being able to predict a dangerous driving situation before it occurs; opposed to recognizing it thereafter. In addition, it would be great if an autonomous driving vehicle could predict the motion trajectory of a pedestrian on the street and avoid the crash, rather than identify the trajectory after the crash into the pedestrian. Unfortunately, most of the existing action recognition approaches are unsuitable for such early classification tasks as they expect to see the entire set of action dynamics from a full video, and then make decisions.

Different from action recognition approaches, action or motion prediction\footnote{In this paper, action prediction refers to the task of predicting action category, and motion prediction refers to the task of predicting motion trajectory. Video prediction is not discussed in this paper as it focuses on motion in videos rather than motion of human.} approaches reason about the future and infer labels before action executions end. These labels could be the discrete action categories, or continuous positions on a motion trajectory. The capability of making a prompt reaction makes action/motion prediction approaches more appealing in time-sensitive tasks. However, action/motion prediction is really challenging because accurate decisions have to be made on partial action videos.





\begin{figure}[!t]
\begin{center}
\includegraphics[width=1\linewidth]{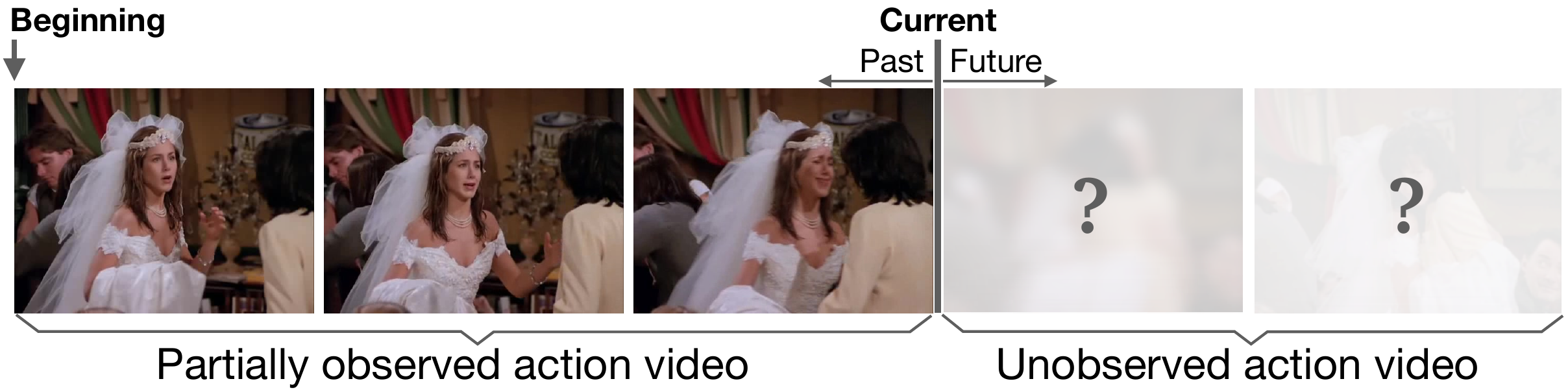}
\end{center}
\caption{Early action classification methods predicts action label given a partially observed video. Revised based on the original figure in \cite{KongECCV2014}.}
\label{fig:eccv2014}
\end{figure}
\subsection{Action Prediction}
Action prediction tasks can be roughly categorized into two types, \textit{short-term prediction} and \textit{long-term prediction}. The former one, short-term prediction focuses on short-duration action videos, which generally last for several seconds, such as action videos in UCF-101 and Sports-1M datasets. The goal of this task is to infer action labels based upon temporally incomplete action videos. Formally, given an incomplete action video $\mathbf{x}_{1:t}$ containing $t$ frames, i.e., $\mathbf{x}_{1:t}=\{f_1,f_2,\cdots,f_t\}$, the goal is to infer the action label $y$: $\mathbf{x}_{1:t}\rightarrow y$. Here, the incomplete action video $\mathbf{x}_{1:t}$ contains the beginning portion of a complete action execution $\mathbf{x}_{1:T}$, which only contains one single action. The latter one, long-term prediction or intention prediction, infers the future actions based on current observed human actions. It is intended for modeling action transition, and thus focuses on long-duration videos that last for several minutes. In other words, this task predicts the action that is going to happen in the future. More formally, given an action video $\mathbf{x}_a$, where $\mathbf{x}_a$ could be a complete or an incomplete action execution, the goal is to infer the next action $\mathbf{x}_b$. Here, $\mathbf{x}_a$ and $\mathbf{x}_b$ are two independent, semantically meaningful, and temporally correlated actions.

\begin{table*}
\setlength{\tabcolsep}{1pt}
\centering
\caption{Results of early action classification methods on various datasets. X@Y denotes the prediction results at Y dataset when observation ratio is set to X. ``-'' indicates the result is not reported. }
 \label{tab:pred}
 \begin{tabular}{|c||c|c|c|c|c|c|c|c|c|}
    \hline
    Methods & Year & 0.1@BIT & 0.5@BIT &0.1@UTI-1 & 0.5@UTI-1 & 0.1@UCF-101 & 0.5@UCF-101 & 0.1@Sports-1M & 0.5@Sports-1M\\
    \hline\hline
Integral BoW \cite{RyooICCV2011}  & 2011 & $22.66\%$ & $48.44\%$ & $18.00\%$ & $48.00\%$ & $36.29\%$ & $74.39\%$ & $43.47\%$ & $55.99\%$ \\
MSSC \cite{CaoCVPR2013}   & 2013         & $21.09\%$ & $48.44\%$ & $28.00\%$ & $70.00\%$ & $34.05\%$ & $61.79\%$ & $46.70\%$ & $57.16\%$ \\
Poselet \cite{RaptisCVPR2013}  & 2013     & -         & -         & -         & $73.33\%$ & -         &           &           &           \\
HM \cite{LanECCV2014}  & 2014             & -         & -         & $38.33\%$ & $83.10\%$ & -         &           & -         &           \\
MTSSVM \cite{KongECCV2014}  & 2014        & $28.12\%$ & $60.00\%$ & $36.67\%$ & $78.33\%$ & $40.05\%$ & $82.39\%$ & $49.92\%$ & $66.90\%$ \\
MMAPM \cite{KongTPAMI2016}  & 2016        & $32.81\%$ & $67.97\%$ & $46.67\%$ & $78.33\%$ & -         & -         & -         &           \\
DeepSCN \cite{KongCVPR2017}  & 2017       & $37.50\%$ & $78.13\%$ & -         & -         & $45.02\%$ & $85.75\%$ & $55.02\%$ & $70.23\%$ \\
GLTSD \cite{LaiTIP2018}  & 2018           & $26.60\%$ & $79.40\%$ & -         & -         & -         &           & -         & -         \\
mem-LSTM \cite{KongAAAI2018}  & 2018      & -         & -         & -         & -         & $51.02\%$ & $88.37\%$ & $57.60\%$ & $71.63\%$\\
  \hline
\end{tabular}
\end{table*}
\subsubsection{Early Action Classification}
This task aims at recognizing a human action at an early stage, i.e., based on a temporally incomplete video (see Figure~\ref{fig:eccv2014}). The goal is to achieve high recognition accuracy when only the beginning portion of a video is observed. The observed video contains an unfinished action, and thus making the prediction task challenging. Although this task may be solved by action recognition methods \cite{RaptisCVPR2013,Vahdat2011,YaoPAMI2012,YaoECCV2012}, they were developed for recognizing complete action executions, and were not optimized for partial action observations, making action recognition approaches unsuitable for predicting actions at an early stage. Table \ref{tab:pred} provides some results of early action classification on four datasets.

\begin{figure}[!t]
\begin{center}
\includegraphics[width=1\linewidth]{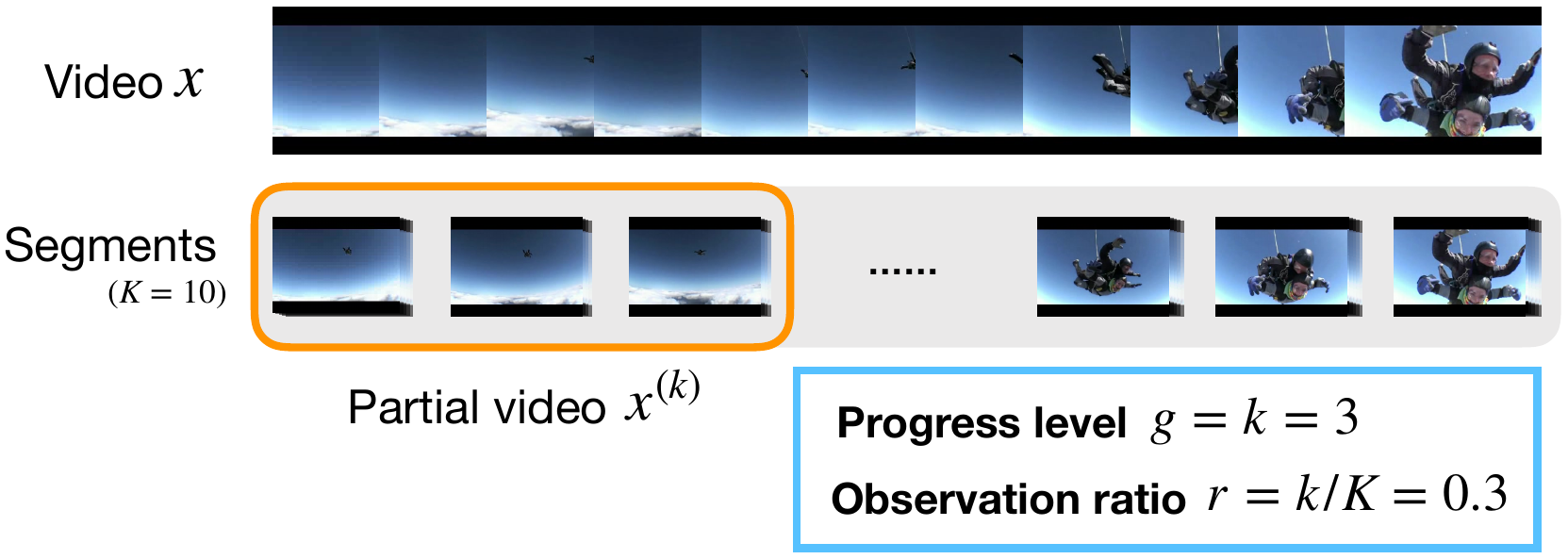}
\end{center}
\caption{Example of a temporally partial video, and graphical illustration of progress level and observation ratio. Revised based on the original figure in \cite{KongCVPR2017}.}
\label{fig:setting}
\end{figure}
Most of the short-term action prediction approaches follow the problem setup described in \cite{KongECCV2014} shown in Figure~\ref{fig:setting}. To mimic sequential data arrival, a complete video $\mathbf{x}$ with $T$ frames is segmented into $K=10$ segments. Consequently, each segment contains $\frac{T}{K}$ frames. Video lengths $T$ may vary for different videos, thereby causing different lengths in their segments. For a video of length $T$, its $k$-th segment ($k\in\{1,\cdots, K\}$) contains frames starting from the $[(k-1)\cdot \frac{T}{K}+1]$-th frame to the $(\frac{kT}{K})$-th frame. A temporally \textit{partial video} or \textit{partial observation} $\mathbf{x}^{(k)}$ is defined as a temporal subsequence that consists of the beginning $k$ segments of the video. The \textit{progress level} $g$ of the partial video $\mathbf{x}^{(k)}$ is defined by the number of the segments contained in the partial video $\mathbf{x}^{(k)}$: $g=k$. The \textit{observation ratio} $r$ of a partial video $\mathbf{x}^{(k)}$ is $\frac{k}{K}$: $r=\frac{k}{K}$. 

Action prediction approaches aim at recognizing unfinished action videos. Ryoo \cite{RyooICCV2011} proposed the integral bag-of-words (IBoW) and dynamic bag-of-words (DBoW) approaches for action prediction. The action model of each progress level is computed by averaging features of a particular progress level in the same category. However, the learned model may not be representative if the action videos of the same class have large appearance variations, and it is sensitive to outliers. To overcome these two problems, Cao \etal \cite{CaoCVPR2013} built action models by learning feature bases using sparse coding and used the reconstruction error in the likelihood computation. Li \etal \cite{LiECCV2012} explored long-duration action prediction problem. However, their work detects segments by motion velocity peaks, which may not be applicable to complex outdoor datasets. Compared with \cite{CaoCVPR2013,LiECCV2012,RyooICCV2011}, \cite{KongECCV2014} incorporates an important prior knowledge that informative action information is increasing when new observations are available. In addition, the method in \cite{KongECCV2014} models label consistency of segments, which is not presented in their methods. From a perspective of interfering social interaction, Lan \textit{et al.}~\cite{LanECCV2014} developed ``hierarchical movements'' for action prediction, which is able to capture the typical structure of human movements before an action is executed. An early event detector \cite{HoaiCVPR2012} was proposed to localize the starting and ending frames of an incomplete event. Their method first introduces a monotonically increasing scoring function in the model constraint, which has been popularly used in a variety of action prediction methods \cite{KongECCV2014,KongTPAMI2016,MaCVPR2016}. Different from the aforementioned methods, \cite{RyooHRI2015} studied the action prediction problem in a first-person scenario, which allows a robot to predict a person's action during human-computer interactions.

\begin{figure}[!t]
\begin{center}
\includegraphics[width=1\linewidth]{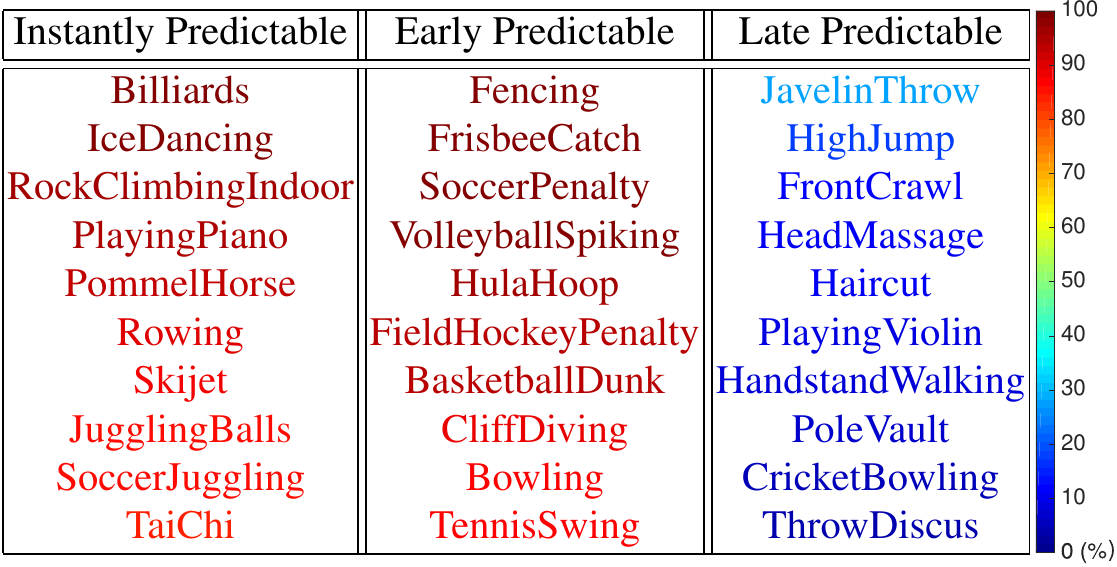}
\end{center}
\caption{Top 10 instantly, early, and late predictable actions in UCF101 dataset. Action names are colored and sorted according to the percentage of their testing samples falling in the category of instant predictable, early predictable, or late predictable. Originally shown in \cite{KongCVPR2017}.}
\label{fig:actionpredictability}
\end{figure}
Deep learning methods have also shown in action prediction. The work in \cite{MaCVPR2016} proposed a new monotonically decreasing loss function in learning LSTMs for action prediction. Inspired by that, the work in \cite{KongCVPR2017} adopted an autoencoder to model sequential context information for action prediction. This method learns such information from fully-observed videos, and transfer it to partially observed videos. We enforced that the amount of the transferred information is temporally ordered for the purpose of modeling the temporal orderings of inhomogeneous action segments. We demonstrated that actions differ in their predictability, and show the top 10 instantly, early, and late predictable actions in Figure~\ref{fig:actionpredictability}. We also studied the action prediction problem following the popular two-stream framework \cite{SimonyanNIPS2014}. In \cite{KongAAAI2018}, we proposed to use memory to store hard-to-predict training samples in order to improve the prediction performance at the early stage. The memory module used in \cite{KongAAAI2018} measures the predictability of each training sample, and will store those challenging ones. Such a memory retains a large pool of samples, and allows us to create complex classification boundaries, which are particularly useful for discriminating partial videos at the beginning stage.



\subsubsection{Action Anticipation}

Action anticipation aims to anticipate future actions from a history of actions \cite{red}. This task is fundamental to many real world applications. For example, surveillance cameras can raise an alarm before a road accident happens, robots can make better plans and decisions by anticipating human actions \cite{robotic_antic}. Action anticipation is a challenging task because the models not only need to detect the actions, but also infer future actions from the seen actions. RED \cite{red} uses an encoder-decoder LSTM structure to predict the future video representations from the extracted representations of the historical video frames. Similarly, two LSTMs were used in \cite{furnari2020rulstm} to summarize the past and infer the future for egocentric videos. The work in \cite{Vondrick} trains a CNN to regress the future representations from the past ones in an unsupervised way.
Three similarity metrics between the past and future video representations were presented in \cite{fernando}, namely Jaccard vector similarity, Jaccard cross-correlation, and Jaccard Frobenius inner product over covariances for early action anticipation. 
Future actions are predicted in \cite{vae_anticip} by learning a distribution of future actions using Variational Auto-Encoder. The work in \cite{Qiuhong} predicts actions at different future timestamps in one-shot by incorporating a temporal parameter and skip connections. 
Hyperbolic space is used in \cite{suris2021learning} to predict future actions because it can represent actions through a compact hierarchy.
In \cite{ke2021future}, the authors proposed a model that consists of a conditional VAE for modeling the uncertainty of the action starting time and a MLP to predict whether the action will happen. Recently, ~\cite{Rohit21} presents a new model called Anticipative Video Transformer and a self-supervised future prediction loss for action anticipation.

\subsubsection{Intention Prediction}
In practice, there are certain types of actions that contain several primitive action patterns and exhibit complex temporal arrangements, such as ``make a dish''. Typically, the length of these complex actions is longer than that of short-term actions. Prediction of these long-term actions is receiving a surge of interest as it allows us to understand ``what is going to happen'', including the final goal of complex human action and the person's plausible intended action in the near future.

However, long-term action prediction is extremely challenging due to the large uncertainty in human future actions. Cognitive science shows that context information is critical to action understanding, as they typically occur with certain object interactions under particular scenes. Therefore, it would be helpful to consider the interacting objects together with the human actions, in order to achieve accurate long-term action prediction. Such knowledge can provide valuable clues for two questions ``what is happening now?'' and ``what is going to happen next?''. It also limits the search space for potential actions using the interacting object. For example, if an action ``a person grabbing a cup'' is observed, most likely the person is going to ``drink a beverage'', rather than going to ``answering a phone''. Therefore, a prediction method considering such context is expected to provide opportunities to benefit from contextual constraints between actions and objects.

\begin{figure}[t]
\begin{center}
\includegraphics[width=1\linewidth]{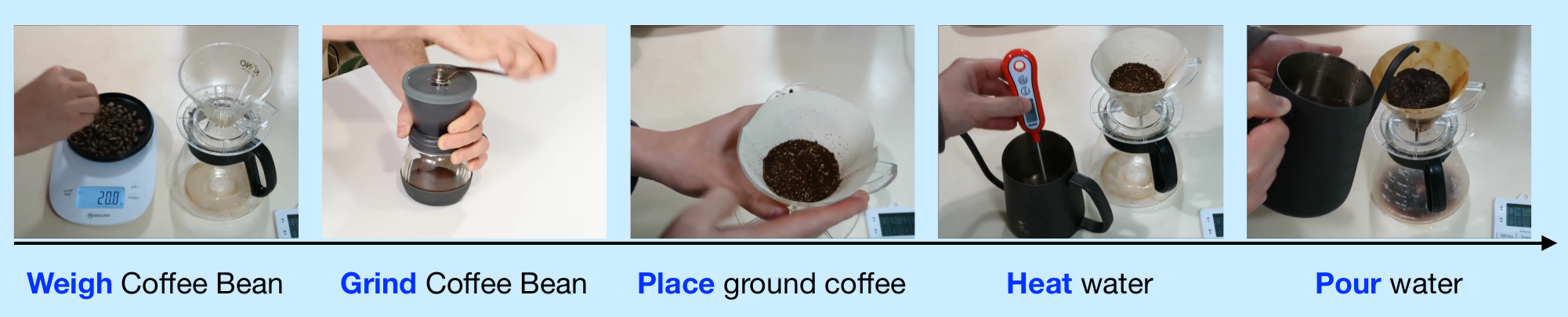}
\end{center}
\caption{A complex action can be decomposed into a series of action primitives. Revised based on the original figure in \cite{LiTPAMI2014}.}
\label{fig:longterm}
\end{figure}

Pei \etal \cite{PeiICCV2011} addressed the problem of goal inference and intent prediction using an And-Or-Graph method, in which the Stochastic Context Sensitive Grammar is embodied. They modeled agent-object interactions, and generated all possible parse graphs of a single event. Combining all the possibilities generates the interpretation of the input video and achieves the global maximum posterior probability. They also show that ambiguities in the recognition of atomic actions can be reduced largely using hierarchical event contexts. Li \etal \cite{LiECCV2012} proposed a long-term action prediction method using Probabilistic Suffix Tree (PST), which captures variable Markov dependencies between action primitives in complex action. For example, as shown in Figure~\ref{fig:longterm}, a wedding ceremony can be decomposed into primitives of ``hold-hands'', ``kneel'', ``kiss'', and ``put-ring-on''. In their extension \cite{LiTPAMI2014}, object context is added to the prediction model, which enables the prediction of human-object interactions occurring in actions such as ``making a dish''. Their work first introduced a concept ``predictability'', and used the Predictive Accumulative Function (PAF) to show that some actions can be early predictable while others cannot be early predicted. Prediction of human action and object affordance was investigated in \cite{KoppulaTPAMI2016}. They proposed an anticipatory temporal conditional random field (ATCRF) to model three types of context information, including the hierarchical structure of action primitives, the rich spatial-temporal correlations between objects and their affordances, and motion anticipation of objects and humans. In order to find the most likely motion, ATCRFs are considered as particles, which are propagated over time to represent the distribution of possible actions in the future. 
The work in \cite{girase2021loki} introduces a new dataset called LOKI (LOng term and Key Intentions) for autonomous driving. The authors also proposed a long-term goal proposal network and a scene graph refinement and trajectory decoder module for jointly predicting the future trajectory and intention of pedestrians. In ~\cite{bhattacharyya2021euro}, the authors provided a new dataset for pedestrian trajectory prediction in dense urban scenarios. A Joint-$\beta$-cVAE is further designed to effectively model the interaction between pedestrians and vehicles, the model is trained by optimizing the ELBO The authors in \cite{rasouli2021bifold} proposed a multi-task learning framework which predicts both trajectories and actions of pedestrians conditioned on multi-modal data. They proposed a bi-fold feature fusion to effectively fuse multiple modalities, also a semantic map as an additional input to the model for categorical interaction modeling during training.

\subsection{Summary}
The availability of big data and recent advance in computer vision and machine learning enable the reasoning about the future. The key in this research is how to discover temporal correlations in large-scale data and how to model such correlations. Results shown in Table~\ref{fig:actionpredictability} demonstrate the predictability of actions that can be used as a prior and inspiring more powerful action prediction methods. There are still some unexplored opportunities in this research, such as interpretability of temporal extent, how to model long-term temporal correlations, and how to utilize multi-modal data to enrich the prediction model, which will be discussed in Section~\ref{sec:future}.

\section{Motion Trajectory Prediction}
Besides predicting human actions, the other key aspect in human-centered prediction is motion trajectory prediction, which aims at predicting a pedestrian's moving path. Motion trajectory prediction, an inherent capability of us, reasons the possible destination and motion trajectory of the target person. We can predict with high confidence that a person is going to walk on sidewalks than streets, and will avoid any obstacles during walking. Therefore, it is interesting to study how to make machines do the same job. Table \ref{ETH} shows the ADE and FDE results on ETH/UCF dataset. ADE and FDE are standard metrics on motion trajectory prediction. Some works do not report the FDE result.

\begin{table}[]
\caption{Results of motion trajectory prediction on ETH/UCF datasets. ADE is the minimum average displacement error, and FDE denotes the final displacement error. ``-'' indicates the result is not reported.}
\begin{tabular}{|l|l|l|}

\hline
                                      & ADE & FDE \\ \hline
Social GAN \cite{GuptaCVPR2018}                   & 0.58     &  -        \\ \hline
Sophie \cite{sadeghian2019sophie}             & 0.54     &  -        \\ \hline
CGNS \cite{li2019conditional}                 & 0.49     &  -        \\ \hline
Social BiGAT \cite{kosaraju2019social}        & 0.48     &  -        \\ \hline
Next \cite{liang2019peeking}                  & 0.46     &  -        \\ \hline
Social-STCNN \cite{mohamed2020social}         & 0.44     &  -        \\ \hline
MANTRA \cite{marchetti2020mantra}             & 0.32     & 0.65     \\ \hline
Transformer TF \cite{giuliari2021transformer} & 0.31     &  -        \\ \hline
PECNet \cite{mangalam2020not}                 & 0.29     & 0.48     \\ \hline
Social-NCE \cite{liu2020social}               & 0.19     & 0.40     \\ \hline
SGNet \cite{wang2021stepwise}                  & 0.18     & 0.35     \\ \hline
AgentFormer \cite{yuan2021agentformer}        & 0.18     & 0.29     \\ \hline
Y-Net \cite{mangalam2020goals}                & 0.18     & 0.27    \\ \hline
\end{tabular}

\label{ETH}
\end{table}

\begin{figure}[t]
\begin{center}
\includegraphics[width=1\linewidth]{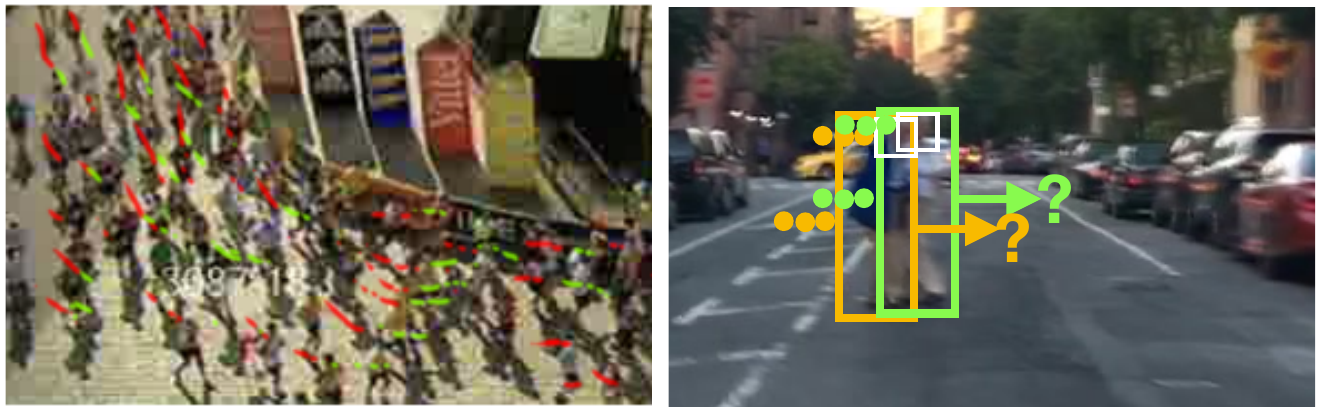}
\end{center}
\caption{Motion trajectory prediction is essential for practical applications such as visual surveillance and self-driving cars.}
\label{fig:trajpred}
\end{figure}

Vision-based motion trajectory prediction is essential for practical applications such as visual surveillance and self-driving cars (see Figure~\ref{fig:trajpred}), in which reasons about the future motion patterns of a pedestrian is critical. A large body of work learns motion patterns by clustering trajectories~\cite{ZhouCVPR2011,MorrisandTPAMI2011,KimICCV2011,HuTIP2007}. However, forecasting future motion trajectory of a person is really challenging as the prediction cannot be predicted in isolation. In a crowded environment, humans adapt their motion according to the behaviors of neighboring people. They may stop, or alter their paths to accommodate other people or the environment in the vicinity. Jointly modeling such complex dependencies is really difficult in dynamic environments. In addition, the predicted trajectories should not only be \textit{physically acceptable}, but also \textit{socially acceptable} \cite{GuptaCVPR2018}. Pedestrians always respect personal space while walking, and thus yield the right-of-way. Human-human and human-object interactions are typically subtle and complex in crowded environments, making the problem even more challenging. Furthermore, there are multiple future predictions in a crowded environment, which are all socially acceptable. Thus uncertainty estimation for the multimodal predictions is desired. 

Forecasting trajectory and destination by understanding the physical scene was investigated in \cite{KitaniECCV2012}, which was one of the pioneering work in trajectory prediction in the computer vision community. The proposed method models the effect of the physical environment on the choice of human actions. 
The authors integrate state-of-the-art semantic scene understanding with the ideas from inverse optimal control (IOC) or inverse reinforcement learning \cite{AbbeelICML2004,ZiebartAAAI2008}.
In this work, human motion is modeled as a sequence of decision-making process, and a prediction is made by maximizing the reward. Lee and Kitani \cite{LeeWACV2016} extends \cite{KitaniECCV2012} to a dynamic environment. The state reward function is extended to a linear combination of static and dynamic state functions to update the forecasting distribution in a dynamic environment. However, IOC is limited to controlled settings as the goal state of the pedestrian's destination requires a priori. To relax this assumption, the concept of \textit{goal set} was introduced in \cite{Mainprice2016,DraganICRA2011}, which defines a target task space. The work in \cite{AlahiCVPR2014} introduced a large-scale dataset of $42$ million trajectories and studied the problem of trajectory prediction by modeling the social interactions of pedestrians. They captured the spatial positions of the neighboring trajectories of a person by a so-called \textit{social affinity map}. The trajectory prediction task is formulated as a maximum a-posterior estimation problem, and the origin and destination prior knowledge is introduced to the model. The method in \cite{BallanECCV2016} takes a step further and generalizes trajectory prediction by considering human-scene interactions. Instead of just using semantic labels of the scene (\eg, grass, street, etc), functional properties of a scene map \cite{TurekECCV2010} are learned in \cite{BallanECCV2016}, which allows the prediction model to understand how agents of the same class move from one patch to another. This provides us with rich navigation patterns to the final destination. Scene semantics was also used to predict the dynamics of multiple objects \cite{FouheyCVPR2014,HuangECCV2008,KooijECCV2014,KretzschmarICRA2014}. Kooji \etal \cite{KooijECCV2014} focused on predicting pedestrians' path intention of crossing the street from the viewpoint of an approaching vehicle. Their method is built upon the dynamic Bayesian network (DBN), which considers the pedestrian's decision to stop by three cues, including the existence of an approaching vehicle, the pedestrian's awareness, and the spatial layout of the scene. Walker \etal in \cite{WalkerCVPR2014} predicted the behavior of agents (\eg, a car) in a visual scene. Ziebart \etal \cite{ZiebartIROS2009} presented a planning-based approach for trajectory prediction.

Thanks to the recent advance in deep networks, motion trajectory prediction problem can be solved using RNN/LSTM networks \cite{AlahiCVPR2016,LeeCVPR2017,SuIJCAI2017,GuptaCVPR2018}, which have the capability of generating long sequences. More specifically, a single LSTM model was used to account for one single person's trajectory, and a social pooling layer in LSTMs was proposed to  model dependencies between LSTMs, and preserve the spatial information \cite{AlahiCVPR2016}. Compared to previous work \cite{KitaniECCV2012,LeeWACV2016,AlahiCVPR2014,BallanECCV2016,KooijECCV2014}, the method in \cite{AlahiCVPR2016} is end-to-end trainable, and generalizes well in complex scenes. An encoder-decoder framework was proposed in \cite{LeeCVPR2017} for path prediction in more natural scenarios where agents interact with each other and dynamically adapt their future behaviors. Past trajectories are encoded in a RNN and then future trajectory hypotheses are generated using another decoder implemented by a separate RNN. This method also extends inverse optimal control (IOC) \cite{LeeWACV2016,KitaniECCV2012} to a deep model, which has shown promising results in robot control \cite{Finn2016} and driving \cite{Wulfmeier2016} tasks. The proposed Deep IOC is used to rank all the possible hypotheses. The scene context is captured using a CNN model, which is part of the input to the RNN encoder. A Social-GAN network in \cite{GuptaCVPR2018} was proposed to address the limitation of L2 loss in \cite{LeeCVPR2017}. Using an adversarial loss, \cite{GuptaCVPR2018} can potentially learn the distribution of multiple socially acceptable trajectories, rather than learning the average trajectories in the training data. 
The work in \cite{dendorfer2021mg} proposes a Multi-Generator Model (MGM) to address the problem of out-of-distribution samples generated using a single generator. A categorical distribution over different trajectory types is first predicted by a Path Module Network, from which the generator is chosen to sample the future trajectories. Thus, the model can select scene-specific generators and deactivate unsuitable ones. A divide and conquer method was proposed in \cite{narayanan2021divide} which prevents mode collapse problems in trajectory prediction under the winner-takes-all objective. The work in \cite{zhao2021you} proposes a model for goal-conditioned trajectory prediction which exploits nearest examples for goal position query and considers multi-modality and physical constraints.

\section{Datasets}
This section discusses some of the popular action video datasets, including actions captured in a controlled and uncontrolled environment. A detailed list is shown in Table~\ref{tab:dataset}. These datasets differ in the number of human subjects, background noise, appearance and pose variations, camera motion, etc., and have been widely used for the comparison of various algorithms.

\begin{table*}[htp]
\caption{A list of popular action video datasets used in action recognition research.}
\begin{center}
\begin{tabular}{|c|c|c|c|c|c|c|c|}
\hline
Datasets & Year & $\#$Videos & $\#$Views & $\#$Actions & $\#$Subjects & $\#$Modality & Env.\\
\hline
KTH \cite{SchuldtICPR2004} 	 		& 2004 & 600	& 1 	& 6  	& 25 & RGB &Controlled\\
Weizmann \cite{BlankICCV2005} 		& 2005 & 90		& 1		& 10	& 9  & RGB & Controlled\\
INRIA XMAS \cite{WeinlandCVIU2006}  & 2006 & 390 	& 5 	& 13 	& 10(3 times) & RGB & Controlled\\
IXMAS \cite{YuanCVPR2009}	   		& 2006 & 1,148	& 5 	& 11 	& -  & RGB & Controlled\\
UCF Sports \cite{RodriguezCVPR2008}	& 2008 & 150  	& - 	& 10 	& - & RGB & Uncontrolled\\
Hollywood \cite{LaptevCVPR2008}		& 2008 & -   	& - 	& 8 	& - & RGB & Uncontrolled\\
Hollywood2 \cite{MarszalekCVPR2009}	& 2009 & 3,669  & -		& 12 	& 10 & RGB & Uncontrolled\\
UCF 11 \cite{LiuCVPR2009a}			& 2009 & 1,100+ & -		& 11 	& - & RGB & Uncontrolled\\
CA \cite{ChoiICCVW2009}				& 2009 & 44  	& - 	& 5 	& - & RGB & Uncontrolled\\
MSR-I \cite{YuanCVPR2009}	 	 	& 2009 & 63		& - 	& 3 	& 10 & RGB & Controlled\\
MSR-II \cite{YuanPAMI2010}		    & 2010 & 54   	& - 	& 3 	& - & RGB & Crowded\\
MHAV \cite{MuHAVi2010}			    & 2010 & 238 	& 8 	& 17 	& 14 & RGB & Controlled\\
UT-I \cite{UT-Interaction-Data}		& 2010 & 60   	& 2 	& 6 	& 10 & RGB & Uncontrolled\\
TV-I \cite{PerezBMVC2010} 			& 2010 & 300  	& - 	& 4 	& - & RGB & Uncontrolled\\
MSR-A \cite{LiCVPRW2010}	 		& 2010 & 567 	& - 	& 20 	& 1 & RGB-D & Controlled\\
Olympic	\cite{NieblesECCV2010}		& 2010 & 783   	& - 	& 16	& - & RGB & Uncontrolled \\
HMDB51 \cite{KuehneICCV2011} 		& 2011 & 6849   & - 	& 51 	& - & RGB & Uncontrolled\\
CAD-60 \cite{SungAAAIW2011} 		& 2011 & 60 	& - 	& 12 	& 4 & RGB-D & Controlled\\
BIT-I \cite{KongECCV2012}	 		& 2012 & 400   	& - 	& 8 	& 50 & RGB & Controlled\\
LIRIS \cite{wolf2014evaluation} 	& 2012 & 828   	& 1 	& 10 	& - & RGB & Controlled\\
MSRDA \cite{WangJCVPR2012}  		& 2012 & 320 	& - 	& 16 	& 10 & RGB-D & Controlled\\
UCF50 \cite{ReddyMVA2012}	 		& 2012 & 50   	& - 	& 50 	& - & RGB & Uncontrolled\\
UCF101 \cite{Soomro2012}	 		& 2012 & 13,320 & -		& 101 	& - & RGB & Uncontrolled\\
MSR-G \cite{Kurakin2012}			& 2012 & 336 	& - 	& 12 	& 1 & RGB-D & Controlled\\
UTKinect-A \cite{XiaCVPRW2012}		& 2012 & 200    & - 	& 10 	& - & RGB-D & Controlled\\
ASLAN \cite{GrossPAMI2012}	 		& 2012 & 3,698  & -		& 432 	& - & RGB & Uncontrolled\\
MSRAP  \cite{OreifejCVPR2013} 		& 2013 & 360 	& - 	& 6 pairs& 10 & RGB-D & Controlled\\
CAD-120  \cite{Koppula2013}		 	& 2013 & 120 	& - 	& 12	& 4 & RGB-D & Controlled\\
THUMOS'14 \cite{THUMOS14} 	 		& 2014 & 413	& 1 	& 20  	& -- & RGB &Uncontrolled\\
Sports-1M \cite{KarpathyCVPR2014} 	& 2014 & 1,133,158  & -	& 487 	& - & RGB & Uncontrolled\\
3D Online  \cite{YuACCV2014}		& 2014 & 567 	& - 	& 20 	& - & RGB-D & Uncontrolled\\
FCVID \cite{FCVID}			 		& 2015 & 91,233 & - 	& 239 	& - & RGB & Uncontrolled\\
ActivityNet \cite{caba2015activitynet} & 2015 & 28,000 & - 	& 203	& - & RGB & Uncontrolled\\
YouTube-8M \cite{abu2016youtube}	& 2016 & 8,000,000 & - 	& 4,716 & - & RGB & Uncontrolled\\
Charades \cite{nturgb60} 	& 2016 & 9,848 	& 2 	& 157 	& - & RGB & Controlled\\
NTU-RGB+D~\cite{Shahroudy_2016_NTURGBD} & 2016 & 56,680  & - & 120 & 106 & RGB+D+IR+Skeleton & Controlled\\
PKU-MMD (Phase 1) \cite{PKUMMD2017} 		& 2017 & 1076		& 3		& 51	& 66  & RGB+D+IR+Skeleton & Uncontrolled\\
PKU-MMD (Phase 2) \cite{PKUMMD2017} 		& 2017 & 2000		& 3		& 49	& 13  & RGB+D+IR+Skeleton & Uncontrolled\\
NEU-UB 		& 2017    & 600 	& - 	& 6 	& 20 & RGB-D & Controlled\\
Kinetics \cite{kay2017kinetics} 	& 2017 & 500,000& - 	& 600	& - & RGB & Uncontrolled\\
AVA \cite{gu2017ava}		 		& 2017 & 57,600 & - 	& 80 	& - & RGB & Uncontrolled\\
20BN-Something-Something \cite{goyal2017something} & 2017 & 108,499 	& - & 174 & - & RGB & Uncontrolled\\
SLAC \cite{zhao2017slac}	 		& 2017 & 520,000& - 	& 200 	& - & RGB & Uncontrolled\\
Moments in Time \cite{monfortmoments} & 2017 & 1,000,000&-& 339 	& - & RGB & Uncontrolled\\
EPIC-Kitchens~\cite{damen2018scaling} & 2018 & 90,000+ & - & 397 & 32 & RGB & Uncontrolled\\
COIN \cite{COIN2019} 		& 2019 & 11,827		& 1		& 180	& --  & RGB & Uncontrolled\\
HACS Segments \cite{HACS2019} 		& 2019 & 50,000+		& 1		& 200	& --  & RGB & Uncontrolled\\
HAA00 \cite{haa500} & 2021 & 10,000 & - & 500 & - & RGB & Uncontrolled  \\
MultiSports \cite{multi-sports} & 2021 & 3200 & - & 4 & - & RGB & Uncontrolled \\

\hline
\end{tabular}
\end{center}
\label{tab:dataset}
\end{table*}%

\subsection{Controlled Action Video Datasets}
We first describe individual action datasets captured in controlled settings, and then list datasets with two or more people involved in actions. We also discuss some of the RGB-D action datasets captured using a cost-effective Kinect sensor.

\subsubsection{Individual Action Datasets}
\textbf{Weizmann dataset} \cite{BlankICCV2005} is a popular video dataset for human action recognition. The dataset contains $10$ action classes such as ``walking'', ``jogging'', ``waving'' performed by $9$ different subjects, to provide a total of $90$ video sequences. The videos are taken with a static camera under a simple background.

\textbf{KTH dataset} \cite{SchuldtICPR2004} consists of $6$ types of human actions (boxing, hand clapping, hand waving, jogging, running and walking) repeated several times by $25$ different subjects in $4$ scenarios (outdoors, outdoors with scale variation, outdoors with different clothes and indoors). There are 600 action videos in the dataset.

\textbf{INRIA XMAS multiview dataset} \cite{WeinlandCVIU2006} was complied for multi-view action recognition. It contains videos captured from $5$ views including a top-view camera. This dataset consists of $13$ actions, each of which is repeated $3$ times by $10$ actors.





\subsubsection{Group Action Datasets}
\textbf{UT-Interaction dataset} \cite{UT-Interaction-Data} is comprised of $2$ sets of $10$ videos with different background and camera settings. The videos contain 6 classes of human-human interactions: handshake, hug, kick, point, punch, and push.


\textbf{BIT-Interaction dataset} \cite{KongECCV2012} consists of $8$ classes of human interactions (bow, boxing, handshake, high-five, hug, kick, pat, and push), with $50$ videos per class. Videos are captured in realistic scenes with cluttered backgrounds, partially occluded body parts, moving objects, and variations in subject appearance, scale, illumination condition, and viewpoint.

\textbf{TV-Interaction dataset} \cite{PerezBMVC2010} contains $300$ videos clips with human interactions. These videos are categorized into $4$ interaction categories: handshake, high five, hug, and kiss, and annotated with the upper body of people, discrete head orientation and interaction. 

\textbf{MultiSports dataset} \cite{multi-sports} is a multi-person dataset that contains $3200$ video clips of $4$ sport classes. The dataset contains $37701$ action instances with $902,000$ bounding boxes, which helps for more fine-grained spatio-temporal action detection and localization.




\subsection{Unconstrained Datasets}
Although the aforementioned datasets lay a solid foundation for action recognition research, they were captured in controlled settings, and may not be able to train approaches that can be used in real-world scenarios. To address this problem, researchers collected action videos from the Internet, and compiled large-scale action datasets, which will be discussed in the following.

\textbf{UCF101 dataset} \cite{Soomro2012} has been widely used in action recognition research. It comprises of realistic videos collected from Youtube. It contains $101$ action categories, with $13320$ videos in total. UCF101 gives the largest diversity in terms of actions and with the presence of large variations in camera motion, object appearance and pose, object scale, viewpoint, cluttered background, illumination conditions, etc. The dataset can be roughly divided into 5 categories: 1) Human-Object Interaction 2) Body-Motion Only 3) Human-Human Interaction 4) Playing Musical Instruments 5) Sports. It should be noted that many clips are collected from the same video. Consequently, different clips may have the same person or the same scenario, or the same lighting, etc. This seems different from practical scenarios, and thus its difficulty is limited.

\textbf{HMDB51 dataset} \cite{KuehneICCV2011} contains a total of about 6849 video clips distributed in a large set of $51$ action categories. Each category contains a minimum of $101$ video clips. In addition to the label of the action category, each clip is annotated with an action label as well as a meta-label describing the property of the clip, such as visible body parts, camera motion, camera viewpoint, number of people involved in the action, and video quality. The actions can be grouped into five categories, including general facial actions (\eg smile, chew, talk), Facial actions with object manipulation (\eg smoke, eat, drink), General body movements (\eg cartwheel, clap hands, climb), Body movements with object interaction (\eg brush hair, catch, draw sword), Body movements for human interaction (\eg fencing, hug, kick someone). The dataset also has two distinct categories namely ``no motion'' and ``camera motion''. The dataset is extremely challenging mainly due to the presence of a significant camera/background motion. To remove camera motion, standard image stitching techniques to can be used to align frames of a clip.

\textbf{Kinetics} \cite{CarreiraCVPR2017} dataset comprises of $700$ human action classes and approximately $650,000$ video clips, including human-object interactions and human-human interactions. The videos were compiled from YouTube by matching its title and the prepared Kinetics actions list. After that, the videos were segmented by tracking actions on Google Image Search, and then labeled by Amazon's Mechanical Turk(AMT). In the end, this dataset is cleaned and de-noised using machine learning techniques. Different from previous datasets, one clip in this dataset may contain several different actions in sequence, but it is only classified into one action category. This means these clips don't have complete action labels. As described in their work, the top-5 measure is supposed to be used because of the incomplete labels. Within the same action category, clips are captured from different videos, including TV and film videos. Consequently, there is a large appearance variation, for example, people in clips may have different age, height, clothes, \etc, and there are various types of camera motion/shake, background clutter. Besides, each clip lasts around 10s and has a variable resolution.


\textbf{Sports-1M dataset} \cite{KarpathyCVPR2014} contains $1,133,158$ video URLs, which have been annotated automatically with 487 labels. It is one of the largest video datasets. Very diverse sports videos are included in this dataset, such as Shaolin Kung Fu, Wing Chun, etc. The dataset is extremely challenging due to very large appearance and pose variations, significant camera motion, noisy background motion, etc. 

\textbf{THUMOS'14 dataset \cite{THUMOS14}} contains more than 20 hours of sport videos. Though the training sets are trimmed videos labeled with 20 action classes, the validation and testing sets include 200 and 213 untrimmed videos, respectively. This dataset has been the most widely used dataset for action detection and localization.

\textbf{ActivityNet dataset \cite{ActivityNet2015}} has two versions for action detection and localization. The first is Activity v1.2, which covers 100 activity classes and contains 4,819 training videos and 2,383 videos for validation. The other version is Activity v1.3, which consists of 10,024 videos for training and 4,926 videos for validation with 200 activity classes. 

\textbf{PKU-MMD dataset \cite{PKUMMD2017}} is a large-scale multi-modal datasets focusing on long continuous sequences action detection and multi-modality action analysis. The first phase contains 51 action categories, performed by 66 distinct subjects in 3 camera views. Each video lasts about $3\sim 4$ minutes and contains approximately 20 action instances. The second phase contains 2,000 short video sequences in 49 action categories, performed by 13 subjects in 3 camera views. Each video lasts about $1\sim 2$ minutes and contains approximately 7 action instances.

\textbf{AVA dataset \cite{AVA2018}} provides audio-visual annotations for about 15 minute long movie clips. For the AVA Action subsets, it contains 430 videos split into 235 for training, 64 for validation, and 131 for test. Each video has 15 minutes annotated in 1-second intervals.

\textbf{COIN dataset \cite{COIN2019}} is a recently released large-scale dataset to address instruction video analysis problems. It contains 11,827 daily activity videos of 180 different classes. Different from other action datasets, human actions in COIN dataset are hierarchically structured with practical semantics.

\textbf{HACS dataset \cite{HACS2019}} is also a recently released large-scale dataset for action localization and recognition. For the HACS Segments subset, it contains 139K action segments densely annotated in 50K untrimmed videos spanning 200 action categories.

\textbf{20BN-SOMETHING-SOMETHING dataset} \cite{goyal2017something} is a dataset shows human interaction with everyday objects. In the dataset, human performs pre-defined action with a daily object. It contains $108,499$ video clips across $174$ classes. The dataset enables the learning of visual representations for the physical properties of the objects and the world.

\textbf{Moments-in-Time dataset} \cite{monfortmoments} is a large-scale video dataset for action understanding. It contains over $1,000,000$ 3-second labeled video clips distributed in $339$ categories. The visual elements of the videos include people, animals, objects or natural phenomena. The dataset is dedicated to building models that are capable of abstracting and reasoning complex human actions.

\textbf{EPIC-Kitchens} dataset \cite{damen2018scaling} is one of the largest first-person vision dataset. It consists of $55$ hours videos and $125$ verb classes and $300$ noun classes recorded by head-mounted camera. These videos are shot at different cities and different styles kitchens and divided to $39,600$ action segments with object bounding boxes. Besides, these videos contain human doing different kitchen tasks at the same time. To better annotate these actions, voice notes for the actions are collected in the dataset.

\textbf{HAA500} dataset \cite{haa500} is a human-centric atomic action dataset. It consists of $500$ atomic classes, where $212$ are sport/athletics, $51$ are playing musical instruments, $82$
are games and hobbies, and $155$ are daily actions.

\subsection{RGB-D Action Video Datasets}
All the datasets described above were captured by RGB video cameras. Recently, there is an increasing interest in using cost-effective Kinect sensors to capture human actions due to the extra depth data channel. Compared to RGB data channels, the extra depth data channel elegantly provides scene structure, which can be used to simplify intra-class motion variations and reduce cluttered background noise \cite{KongIJCV2017}. Popular RGB-D action datasets are listed in the following.

\textbf{MSR Daily Activity dataset} \cite{WangJCVPR2012}: there are $16$ categories of actions: drink, eat, read book, call cellphone, write on a paper, use laptop, use vacuum cleaner, cheer up, sit still, toss paper, play game, lie down on sofa, walk, play guitar, stand up, sit down. All these actions are performed by $10$ subjects. There are $320$ RGB samples and $320$ depth samples available.

\textbf{3D Online Action dataset} \cite{YuACCV2014} was compiled for three evaluation tasks: same-environment action recognition, cross-environment action recognition and continuous action recognition. The dataset contains human action or human-object interaction videos captured from RGB-D sensors. It contains $7$ action categories, such as drinking, eating, and reading cellphone.


\textbf{CAD-120 dataset} \cite{Koppula2013} comprises of $120$ RGB-D action videos of long daily activities. It is also captured using the Kinect sensor. Action videos are performed by $4$ subjects. The dataset consists of 12 action types, such as rinsing mouth, talking on the phone, cooking,  writing on whiteboard, etc. Tracked skeletons, RGB images, and depth images are provided in the dataset. 


\textbf{UTKinect-Action dataset} \cite{utkinect-action3d}  was captured by a Kinect device. There are $10$ high-level action categories contained in the dataset such as walk, sit down, etc. The dataset comprises 200 action vidos and three channels were recorded: RGB, depth and skeleton joint locations.

\textbf{NTU-RGB+D} \cite{nturgb60,nturgb120} dataset contains $60$ action classes and $56,880$ video samples. Recently, it has been extended to $120$ action classes and another $114,480$ video samples in \cite{nturgb120}. All the samples were collected from $106$ distinct subjects by Kinect sensors. RGB videos, depth map sequences, 3D skeletal data, and infrared (IR) videos are provided for each sample. There is higher variation of environmental conditions compared with previous datasets, including $96$ different backgrounds with illumination variations. 

\section{Evaluation Protocols for Action Recognition and Prediction}
Due to different application purposes, action recognition and prediction techniques are evaluated in different ways.

Shallow action recognition methods such as \cite{SchuldtICPR2004,NieblesCVPR2007,WuCVPR2011} were usually evaluated on small-scale datasets, for example, Weizmann dataset \cite{BlankICCV2005}, KTH dataset \cite{SchuldtICPR2004}, UCF Sports dataset \cite{RodriguezCVPR2008}. The leave-one-out training scheme is popularly used on these datasets, and the confusion matrix is usually adopted to show the recognition accuracy of each action category. For sequential approaches such as \cite{WangNIPS2008,WangPAMI2010}, per-frame recognition accuracy is often used. In \cite{MarszalekCVPR2009,TangCVPR2012}, average precision that approximates the area under the precision-recall curve is also adopted for each individual action class. Deep networks \cite{CarreiraCVPR2017,TranICCV2015,VarolTPAMI2017} are generally evaluated on large-scale datasets such as UCF-101 \cite{Soomro2012} and HMDB51 \cite{KuehneICCV2011} and thus can only report overall recognition performance on each dataset. Please refer to \cite{HerathIVC2017} for a list of performance of recent action recognition methods on various datasets.

Most of action prediction methods \cite{RyooICCV2011,CaoCVPR2013,KongECCV2014,KongCVPR2017} were evaluated on existing action datasets. Different from the evaluation method used in action recognition, recognition accuracy at each observation ratio (ranging from $10\%$ to $100\%$) is reported for action prediction methods. As described in \cite{KongCVPR2017}, the goal of these methods is to achieve high recognition accuracy at the beginning stage of action videos, in order to accurately recognize actions as early as possible. Table~\ref{tab:pred} summarizes the performance of action prediction methods on various datasets.

There are several popular metrics for evaluating motion trajectory prediction methods, including \textit{Average Displacement Error} (ADE), \textit{Final Displacement Error} (FDE), and \textit{Average Non-linear Displacement Error} (ANDE). ADE is the mean square error computed over all estimated points of a trajectory and the ground-truth points. FDE is defined as the distance between the predicted final destination and the ground-true final destination. ANDE is the MSE at the non-linear turning regions of a trajectory arising from human-human interactions.

Vairous metrics exist to evaluate action detection and localization methods. Recall that Recall (R) measures the number of true positives over the total number of true positives and false negatives. Average Recall (AR) is the average of recalls over multiple Intersection over Union (IoU) values. Area under the AR vs. AN curve (AUC) measures how well the detection method is able to distinguish between positive and negative proposals. Another metric called mean Average Precision (mAP) @ $\alpha$ where $\alpha$ denotes different IoU threshold which measures the Average Prevision (AP) on each action category.

\section{Future Directions}\label{sec:future}
In this section, we discuss some future directions in action recognition and prediction research that might be interesting to explore. 

\textbf{Dataset.} Significant efforts have been made to collect different types of action video datasets in recent years in order to advance the research of action recognition and prediction. Nevertheless, existing action recognition and prediction models trained on these datasets are still difficult to be generalized to real-world scenarios, possibly because the incapability of these datasets in covering all the aspects that may happen in practical scenarios. First of all, majority of the video datasets were collected under good lighting and weather conditions. However, this assumption may not hold in practice. A visual surveillance system may need to run 24 hours a day whatever the weather is. Unfortunately, existing methods are still difficult to be generalized to poor lighting conditions or extreme weather. Second of all, some datasets are restricted to certain scenarios, for example, UCF101 contains sports videos and EPIC-Kitchens dataset captured in kitchens. Although one well-trained model may perform well in one scenario, it may perform poorly in a new scenario. This could be attributed to the new environment, camera motion, appearance changes, \etc that have not been seen in the previous scenario. Last but not least, existing deep neural networks based methods require a significant amount of data for training. However, video data could be limited in some research areas, such as biomedical research or human rehabilitation research. Is it possible to create and render virtual training video data using game engines such as UnReal \cite{unreal,unrealcv} based on existing small-scale data? This could serve as an alternative solution to directly generalizing deep neural networks to small-scale data. All of these challenges bring new problems to action recognition research and prompt us to collect new datasets to advance the research.

\textbf{Benefitting from image models.}
Deep architectures are dominating the action recognition research lately like the trend of other developments in the computer vision community. However, training deep networks on videos is difficult, and thus benefiting from deep models pre-trained on images or other sources would be a better solution to explore. In addition, image models have done a good job of capturing spatial relationships of objects, which could also be exploited in action understanding. It is interesting to explore how to transfer knowledge from image models to video models using the idea of inflation \cite{CarreiraCVPR2017} or domain adaptation \cite{TangNIPS2012}. 

\textbf{Interpretability on temporal extent.}
Interpretability of image models has been discussed but it has not been extensively discussed in video models. As shown in \cite{SatkinECCV2010,RaptisCVPR2013}, not all frames are equally important for action recognition; only few of them are critical. Therefore, there are a few things that require a deep understanding of the temporal interpretability of video models. First of all, actions, especially long-duration actions can be considered as a sequence of primitives. It would be interesting to have interpretability of these primitives, such as how are these primitives organized in the temporal domain in actions, how do they contribute to the classification task, can we only use few of them without sacrificing recognition performance in order to achieve fast training? In addition, actions differ in their temporal characteristics. Some actions can be understood at their early stage and some actions require more frames to be observed. It would be interesting to ask why these actions can be early predicted, and what are the salient signals that are captured by the machine. Such an understanding would be useful in developing more efficient action prediction models.

\textbf{Learning from multi-modal data.} Humans are observing multi-modal data everyday, including visual, audio, text, etc. These multi-modal data help the understanding of each type of data. For example, reading a book helps us to reconstruct the corresponding part of the visual scene. However, little work is paying attention to action recognition/prediction using multi-modal data. It is beneficial to use multi-modal data to help visual understanding of complex actions because the multi-modal data such as text data contain rich semantic knowledge given by humans. In addition to action labels, which can be considered as verbs, textual data may  include other entities such as nouns (objects), prepositions (spatial structure of the scene), adjectives and adverbs, etc. Although learning from nouns and prepositions have been explored in action recognition and human-object interaction, few studies have been devoted to learning from adjectives and adverbs. Such learning tasks provide more descriptive information about human actions such as motion strength, thereby making fine-grained action understanding into reality.

\textbf{Learning long-term temporal correlations.} 
Multi-modal data also enable the learning of long-term temporal correlations between visual entities from the data, which might be difficult to directly learn from visual data. Long-term temporal correlations characterize the sequential order of actions occurring in a long sequence, which is similar to what our brain stores. When we want to recall something, one pattern evokes the next pattern, suggesting the associations spanning in long-term videos. Interactions between visual entities are also critical to understanding long-term correlations. Typically, certain actions occur with certain object interactions under particular scene settings. Therefore, it needs to involve not only actions, but also an interpretation of objects, scenes and their temporal arrangements with actions, since this knowledge can provide a valuable clue for ``what's happening now'' and ``what's going to happen next''. This learning task also allows us to predict actions in a long-duration sequence.

\textbf{Physical aspect of actions.} Action recognition and prediction are tasks fairly targeting at high-level aspects of videos, and not focusing on finding action primitives that encode basic physical properties. Recently, there has been an increasing interest in learning the physical aspects of the world, which studies fine-grained actions. One example is the something-something dataset introduced in \cite{goyal2017something} that studies human-object interactions. Interestingly, this dataset provides labels or textual description templates such as ``Dropping [something] into [something]'', to describe the interaction between humans and objects, and an object and an object. This allows us to learn models that can understand physical aspects of the world including human actions, object-object interactions, spatial relationships, etc. 

Even though we can infer a large amount of information from action videos, there are still some physical aspects that are challenging to be inferred. We are wondering that can we make a step further, saying understanding more physical aspects, such as the motion style, force, acceleration, etc, from videos? Physics-101 \cite{WuNIPS2015} studied this problem in objects, but can we extend it to actions? A new action dataset containing such fine-grained information is needed. To achieve this goal, our ongoing work is providing a dataset containing human actions with EMG signals, which we hope to benefit fine-grained action recognition.

\textbf{Learning actions without annotations.} For increasingly large action datasets such as Something-Something \cite{goyal2017something} and Sports-1M \cite{KarpathyCVPR2014}, manual labeling becomes prohibitive. Automatic labeling using search engines \cite{KarpathyCVPR2014,abu2016youtube}, video subtitles and movie scripts \cite{MarszalekCVPR2009,LaptevCVPR2008} is possible in some domains, but still requires manual verification. Crowdsourcing \cite{goyal2017something} would be a better option but still suffers from labeling diversity problem, and may generate incorrect action labels. In addition, videos in almost all the action datasets are temporally segmented, with only one action in each of the videos. However, this assumption does not hold as videos may be streaming and it is difficult to know the exact starting and ending frames of an action execution in streaming videos. This prompts us to develop more robust and efficient action recognition/prediction approaches that can automatically learn from unlabeled videos or untrimmed videos.

\textbf{Actions in open-world.} 
Human action recognition in real-world is essentially an open set problem, which requires the model to simultaneously recognize the known action classes and reject the unknown actions~\cite{GengTPAMI2020,BaoICCV2021}. However, existing open set recognition (OSR) research works mainly focus on image modality~\cite{ScheirerTPAMI2012,ScheirerTPAMI2014,ZhangTPAMI2016,BendaleCVPR2016,OzaCVPR2019,PereraCVPR2020,ChenECCV2020}, except for a few works on videos~\cite{ShuICME2018,RoitbergIV2020} and other modalities~\cite{YangPR2019}. These works typically do not work well on video data due to the following challenges. First, the temporal nature of videos leads to high diversity of human actions, which is challenging for an OSR model to be aware of \emph{what it does not know} when given human actions with unknown temporal dynamics. Besides, the static bias (i.e., appearance of the video background and foreground actor) in video data could be easily over-fitted by deep learning models. The model finally could hardly identify unknown actions in an unbiased open vision world. These challenges motivate recent work~\cite{BaoICCV2021} to build an uncertainty-aware and unbiased model for open set action recognition (OSAR). Since open-world actions can be regarded as out-of-distribution (OOD) data, developing more advanced OOD detection methods to tackle the distributional shift of human actions under OSAR setting is promising in the future.

\section{Conclusion}
The availability of big data and powerful models diverts the research focus on human actions from understanding the present to reasoning the future. We have presented a complete survey of state-of-the-art techniques for action recognition and prediction from videos. These techniques became particularly interesting in recent decades due to their promising and practical applications in several emerging fields focusing on human movements. We investigate several aspects of the existing attempts including hand-crafted feature design, models and algorithms, deep architectures, datasets, and system performance evaluation protocols. Future research directions are also discussed in this survey.



\bibliographystyle{spmpsci}      
\bibliography{egbib,kong_eccv14,kong_eccvw14,kong_pami14,aaai18,yuan,deepfeature,egbib_ijcv,kong_interaction_pami,videopred,traj,actionpred,ActPredDan,wentao_locdet,xiwen,xinmiao}   

\begin{thebibliography}{100}
\providecommand{\url}[1]{{#1}}
\providecommand{\urlprefix}{URL }
\expandafter\ifx\csname urlstyle\endcsname\relax
  \providecommand{\doi}[1]{DOI~\discretionary{}{}{}#1}\else
  \providecommand{\doi}{DOI~\discretionary{}{}{}\begingroup
  \urlstyle{rm}\Url}\fi

\bibitem{unreal}
Unreal engine.
\newblock \url{https://www.unrealengine.com/}

\bibitem{unrealcv}
{UnrealCV}.
\newblock \url{https://unrealcv.org}

\bibitem{AbbeelICML2004}
Abbeel, P., Ng, A.: Apprenticeship learning via inverse reinforcement learning.
\newblock In: ICML (2004)

\bibitem{abu2016youtube}
Abu-El-Haija, S., Kothari, N., Lee, J., Natsev, P., Toderici, G., Varadarajan,
  B., Vijayanarasimhan, S.: Youtube-8m: A large-scale video classification
  benchmark.
\newblock arXiv preprint arXiv:1609.08675  (2016)

\bibitem{AlahiCVPR2014}
Alahi, A., Fei-Fei, V.R.L.: Socially-aware large-scale crowd forecasting.
\newblock In: CVPR (2014)

\bibitem{AlahiCVPR2016}
Alahi, A., Goel, K., Ramanathan, V., Robicquet, A., Fei-Fei, L., Savarese, S.:
  Social lstm: Human trajectory prediction in crowded spaces.
\newblock In: CVPR (2016)

\bibitem{BallanECCV2016}
Ballan, L., Castaldo, F., Alahi, A., Palmieri, F., Savarese, S.: Knowledge
  transfer for scene-specific motion prediction.
\newblock In: ECCV (2016)

\bibitem{BaoICCV2021}
Bao, W., Yu, Q., Kong, Y.: Evidential deep learning for open set action
  recognition.
\newblock In: ICCV (2021)

\bibitem{BendaleCVPR2016}
Bendale, A., Boult, T.E.: Towards open set deep networks.
\newblock In: CVPR (2016)

\bibitem{BengioPAMI2013}
Bengio, Y., Courville, A., Vincent, P.: Representation learning: A review and
  new perspectives.
\newblock IEEE Transactions on Pattern Analysis and Machine Intelligence
  (2013)

\bibitem{bhattacharyya2021euro}
Bhattacharyya, A., Reino, D.O., Fritz, M., Schiele, B.: Euro-pvi: Pedestrian
  vehicle interactions in dense urban centers.
\newblock In: CVPR (2021)

\bibitem{TARN}
Bishay, M., Zoumpourlis, G., Patras, I.: Tarn: Temporal attentive relation
  network for few-shot and zero-shot action recognition.
\newblock In: BMVC (2019)

\bibitem{Blake2007}
Blake, R., Shiffrar, M.: Perception of human motion.
\newblock Annu. Rev. Psychol. \textbf{58}, 47--73 (2007)

\bibitem{BlankICCV2005}
Blank, M., Gorelick, L., Shechtman, E., Irani, M., Basri, R.: Actions as
  space-time shapes.
\newblock In: Proc. ICCV (2005)

\bibitem{BobickTPAMI2001}
Bobick, A., Davis, J.: The recognition of human movement using temporal
  templates.
\newblock IEEE Trans Pattern Analysis and Machine Intelligence \textbf{23}(3),
  257--267 (2001)

\bibitem{bojanowski2014weakly}
Bojanowski, P., Lajugie, R., Bach, F., Laptev, I., Ponce, J., Schmid, C.,
  Sivic, J.: Weakly supervised action labeling in videos under ordering
  constraints.
\newblock In: European Conference on Computer Vision, pp. 628--643. Springer
  (2014)

\bibitem{BregonzioCVPR2009}
Bregonzio, M., Gong, S., Xiang, T.: Recognizing action as clouds of space-time
  interest points.
\newblock In: CVPR (2009)

\bibitem{buchler2018improving}
Buchler, U., Brattoli, B., Ommer, B.: Improving spatiotemporal self-supervision
  by deep reinforcement learning.
\newblock In: Proceedings of the European Conference on Computer Vision (ECCV),
  pp. 770--786 (2018)

\bibitem{OTAM_few_shot}
Cao, K., Ji, J., Cao, Z., Chang, C.Y., Niebles, J.C.: Few-shot video
  classification via temporal alignment.
\newblock In: CVPR (2020)

\bibitem{CaoCVPR2013}
Cao, Y., Barrett, D., Barbu, A., Narayanaswamy, S., Yu, H., Michaux, A., Lin,
  Y., Dickinson, S., Siskind, J., Wang, S.: Recognizing human activities from
  partially observed videos.
\newblock In: CVPR (2013)

\bibitem{CarreiraCVPR2017}
Carreira, J., Zisserman, A.: Quo vadis, action recognition? a new model and the
  kinetics dataset.
\newblock In: CVPR (2017)

\bibitem{ChaoCVPR2018}
Chao, Y.W., Vijayanarasimhan, S., Seybold, B., Ross, D.A., Deng, J.,
  Sukthankar, R.: Rethinking the {Faster R-CNN} architecture for temporal
  action localization.
\newblock In: CVPR (2018)

\bibitem{ChenECCV2020}
Chen, G., Qiao, L., Shi, Y., Peng, P., Li, J., Huang, T., Pu, S., Tian, Y.:
  Learning open set network with discriminative reciprocal points.
\newblock In: ECCV (2020)

\bibitem{woo}
Chen, S., Sun, P., Xie, E., Ge, C., Wu, J., Ma, L., Shen, J., Luo, P.: Watch
  only once: An end-to-end video action detection framework.
\newblock In: Proceedings of the IEEE/CVF International Conference on Computer
  Vision (ICCV), pp. 8178--8187 (2021)

\bibitem{ChoiECCV2012}
Choi, W., Savarese, S.: A unified framework for multi-target tracking and
  collective activity recognition.
\newblock In: ECCV, pp. 215--230. Springer (2012)

\bibitem{ChoiICCVW2009}
Choi, W., Shahid, K., Savarese, S.: What are they doing? : Collective activity
  classification using spatio-temporal relationship among people.
\newblock In: Computer Vision Workshops (ICCV Workshops), 2009 IEEE 12th
  International Conference on, pp. 1282 --1289 (2009)

\bibitem{ChoiCVPR2011}
Choi, W., Shahid, K., Savarese, S.: Learning context for collective activity
  recognition.
\newblock In: CVPR (2011)

\bibitem{haa500}
Chung, J., hsin Wuu, C., ru~Yang, H., Tai, Y.W., Tang, C.K.: Haa500:
  Human-centric atomic action dataset with curated videos.
\newblock In: ICCV (2021)

\bibitem{PKUMMD2017}
Chunhui, L., Yueyu, H., Yanghao, L., Sijie, S., Jiaying, L.: Pku-mmd: A large
  scale benchmark for continuous multi-modal human action understanding.
\newblock arXiv preprint arXiv:1703.07475  (2017)

\bibitem{CiptadiECCV2014}
Ciptadi, A., Goodwin, M.S., Rehg, J.M.: Movement pattern histogram for action
  recognition and retrieval.
\newblock In: D.~Fleet, T.~Pajdla, B.~Schiele, T.~Tuytelaars (eds.) Computer
  Vision -- ECCV 2014, pp. 695--710. Springer International Publishing, Cham
  (2014)

\bibitem{Clarke2005}
Clarke, T., Bradshaw, M., Field, D., Hampson, S., Rose, D.: The perception of
  emotion from body movement in point-light displays of interpersonal dialogue.
\newblock Perception \textbf{24}, 1171--80 (2005)

\bibitem{Cutting1977}
Cutting, J., Kozlowski, L.: Recognition of friends by their work: gait
  perception without familarity cues.
\newblock Bull. Psychon. Soc. \textbf{9}, 353--56 (1977)

\bibitem{Dai2017TemporalCN}
Dai, X., Singh, B., Zhang, G., Davis, L., Chen, Y.: Temporal context network
  for activity localization in videos.
\newblock 2017 IEEE International Conference on Computer Vision (ICCV) pp.
  5727--5736 (2017)

\bibitem{DalalCVPR2005}
Dalal, N., Triggs, B.: Histograms of oriented gradients for human detection.
\newblock In: CVPR (2005)

\bibitem{damen2018scaling}
Damen, D., Doughty, H., Farinella, G.M., Fidler, S., Furnari, A., Kazakos, E.,
  Moltisanti, D., Munro, J., Perrett, T., Price, W., Wray, M.: Scaling
  egocentric vision: The epic-kitchens dataset.
\newblock In: European Conference on Computer Vision (2018)

\bibitem{Darwin1872}
Darwin, C.: The Expression of the Emotions in Man and Animals.
\newblock London: John Murray (1872)

\bibitem{DawarSensors2018}
Dawar, N., Kehtarnavaz, N.: Action detection and recognition in continuous
  action streams by deep learning-based sensing fusion.
\newblock IEEE Sensors Journal \textbf{18}(23), 9660--9668 (2018)

\bibitem{Decety1999}
Decety, J., Grezes, J.: Neural mechanisms subserving the perception of human
  actions.
\newblock Neural mechanisms of perception and action \textbf{3}(5), 172--178
  (1999)

\bibitem{dendorfer2021mg}
Dendorfer, P., Elflein, S., Leal-Taix{\'e}, L.: Mg-gan: A multi-generator model
  preventing out-of-distribution samples in pedestrian trajectory prediction.
\newblock In: ICCV (2021)

\bibitem{DibaCVPR2017}
Diba, A., Sharma, V., Gool, L.V.: Deep temporal linear encoding networks.
\newblock In: CVPR (2017)

\bibitem{DollarICCVW2005}
Dollar, P., Rabaud, V., Cottrell, G., Belongie, S.: Behavior recognition via
  sparse spatio-temporal features.
\newblock In: ICCV VS-PETS (2005)

\bibitem{DonahueCVPR2015}
Donahue, J., Hendricks, L., Guadarrama, S., Rohrbach, M., Venugopalan, S.,
  Saenko, K., Darrell, T.: Long-term recurrent convolutional networks for
  visual recognition and description.
\newblock In: CVPR (2015)

\bibitem{DraganICRA2011}
Dragan, A., Ratliff, N., Srinivasa, S.: Manipulation planning with goal sets
  using constrained trajectory optimization.
\newblock In: ICRA (2011)

\bibitem{duchenne2009automatic}
Duchenne, O., Laptev, I., Sivic, J., Bach, F., Ponce, J.: Automatic annotation
  of human actions in video.
\newblock In: 2009 IEEE 12th International Conference on Computer Vision, pp.
  1491--1498. IEEE (2009)

\bibitem{DuongCVPR2005}
Duong, T.V., Bui, H.H., Phung, D.Q., Venkatesh, S.: Activity recognition and
  abnormality detection with the switching hidden semi-markov model.
\newblock In: CVPR (2005)

\bibitem{DutaCVPR2017}
Duta, I.C., Ionescu, B., Aizawa, K., Sebe, N.: spatio-temporal vector of
  locally max pooled features for action recognition in videos.
\newblock In: CVPR (2017)

\bibitem{protoGAN}
Dwivedi, S.K., Gupta, V., Mitra, R., Ahmed, S., Jain, A.: Protogan: Towards few
  shot learning for action recognition.
\newblock In: ICCVW (2019)

\bibitem{EfrosICCV2003}
Efros, A., Berg, A., Mori, G., Malik, J.: Recognizing action at a distance.
\newblock In: ICCV, vol.~2, pp. 726 --733 (2003)

\bibitem{EscorciaECCV2016}
Escorcia, V., Caba~Heilbron, F., Niebles, J.C., Ghanem, B.: {DAPs}: Deep action
  proposals for action understanding.
\newblock In: ECCV (2016)

\bibitem{caba2015activitynet}
Fabian Caba~Heilbron Victor~Escorcia, B.G., Niebles, J.C.: Activitynet: A
  large-scale video benchmark for human activity understanding.
\newblock In: Proceedings of the IEEE Conference on Computer Vision and Pattern
  Recognition, pp. 961--970 (2015)

\bibitem{FantiCVPR2005}
Fanti, C., Zelnik-Manor, L., Perona, P.: Hybrid models for human motion
  recognition.
\newblock In: CVPR (2005)

\bibitem{FeichtenhoferNIPS2016}
Feichtenhofer, C., Pinz, A., Wildes, R.P.: Spatiotemporal residual networks for
  video action recognition.
\newblock In: NIPS (2016)

\bibitem{FeichtenhoferCVPR2017}
Feichtenhofer, C., Pinz, A., Wildes, R.P.: Spatiotemporal multiplier networks
  for video action recognition.
\newblock In: 2017 IEEE Conference on Computer Vision and Pattern Recognition
  (CVPR), pp. 7445--7454. IEEE (2017)

\bibitem{FeichtenhoferCVPR2016}
Feichtenhofer, C., Pinz, A., Zisserman, A.: Convolutional two-stream network
  fusion for video action recognition.
\newblock In: CVPR (2016)

\bibitem{FelzenszwalbCVPR2008}
Felzenszwalb, P., McAllester, D., Ramanan, D.: A discriminatively trained,
  multiscale, deformable part model.
\newblock In: CVPR (2008)

\bibitem{fernando2017self}
Fernando, B., Bilen, H., Gavves, E., Gould, S.: Self-supervised video
  representation learning with odd-one-out networks.
\newblock In: Proceedings of the IEEE conference on computer vision and pattern
  recognition, pp. 3636--3645 (2017)

\bibitem{fernando}
Fernando, B., Herath, S.: Anticipating human actions by correlating past with
  the future with jaccard similarity measures.
\newblock In: CVPR (2021)

\bibitem{Finn2016}
Finn, C., Levine, S., Abbeel, P.: Guided cost learning: deep inverse optimal
  control via policy optimization.
\newblock In: arXiv preprint arXiv:1603.00448 (2016)

\bibitem{FouheyCVPR2014}
Fouhey, D.F., Zitnick, C.L.: Predicting object dynamics in scenes.
\newblock In: CVPR (2014)

\bibitem{furnari2020rulstm}
Furnari, A., Farinella, G.M.: Rolling-unrolling lstms for action anticipation
  from first-person video.
\newblock IEEE Transactions on Pattern Analysis and Machine Intelligence (PAMI)
   (2020)

\bibitem{gan2018geometry}
Gan, C., Gong, B., Liu, K., Su, H., Guibas, L.J.: Geometry guided convolutional
  neural networks for self-supervised video representation learning.
\newblock In: Proceedings of the IEEE Conference on Computer Vision and Pattern
  Recognition, pp. 5589--5597 (2018)

\bibitem{GaoICCV2017}
Gao, J., Yang, Z., Chen, K., Sun, C., Nevatia, R.: {TURN TAP}: Temporal unit
  regression network for temporal action proposals.
\newblock In: ICCV (2017)

\bibitem{GengTPAMI2020}
Geng, C., Huang, S.j., Chen, S.: Recent advances in open set recognition: A
  survey.
\newblock IEEE transactions on pattern analysis and machine intelligence
  (2020)

\bibitem{ghadiyaram2019large}
Ghadiyaram, D., Tran, D., Mahajan, D.: Large-scale weakly-supervised
  pre-training for video action recognition.
\newblock In: Proceedings of the IEEE Conference on Computer Vision and Pattern
  Recognition, pp. 12046--12055 (2019)

\bibitem{girase2021loki}
Girase, H., Gang, H., Malla, S., Li, J., Kanehara, A., Mangalam, K., Choi, C.:
  Loki: Long term and key intentions for trajectory prediction.
\newblock In: ICCV (2021)

\bibitem{GirdharCVPR2017}
Girdhar, R., Ramanan, D., Gupta, A., Sivic, J., Russell, B.: Actionvlad:
  Learning spatio-temporal aggregation for action classification.
\newblock In: CVPR (2017)

\bibitem{giuliari2021transformer}
Giuliari, F., Hasan, I., Cristani, M., Galasso, F.: Transformer networks for
  trajectory forecasting.
\newblock In: 2020 25th International Conference on Pattern Recognition (ICPR),
  pp. 10335--10342. IEEE (2021)

\bibitem{Goodale1992}
Goodale, M.A., Milner, A.D.: Separate visual pathways for perception and
  action.
\newblock Trends in Neurosciences \textbf{15}(1), 20--25 (1992)

\bibitem{GorelickPAMI2007}
Gorelick, L., Blank, M., Shechtman, E., Irani, M., Basri, R.: Actions as
  space-time shapes.
\newblock Transactions on Pattern Analysis and Machine Intelligence
  \textbf{29}(12), 2247--2253 (2007)

\bibitem{goyal2017something}
Goyal, R., Kahou, S.E., Michalski, V., Materzynska, J., Westphal, S., Kim, H.,
  Haenel, V., Fruend, I., Yianilos, P., Mueller-Freitag, M., et~al.: The''
  something something'' video database for learning and evaluating visual
  common sense.
\newblock In: Proc. ICCV (2017)

\bibitem{AVA2018}
Gu, C., Sun, C., Ross, D.A., Vondrick, C., Pantofaru, C., Li, Y.,
  Vijayanarasimhan, S., Toderici, G., Ricco, S., Sukthankar, R., et~al.: {AVA}:
  A video dataset of spatio-temporally localized atomic visual actions.
\newblock In: CVPR (2018)

\bibitem{gu2017ava}
Gu, C., Sun, C., Vijayanarasimhan, S., Pantofaru, C., Ross, D.A., Toderici, G.,
  Li, Y., Ricco, S., Sukthankar, R., Schmid, C., et~al.: Ava: A video dataset
  of spatio-temporally localized atomic visual actions.
\newblock arXiv preprint arXiv:1705.08421  (2017)

\bibitem{NGMN_few_shot}
Guo, M., Chou, E., Huang, D.A., Song, S., Yeung, S., Fei-Fei, L.: Neural graph
  matching networks for fewshot 3d action recognition.
\newblock In: ECCV (2018)

\bibitem{GuptaCVPR2018}
Gupta, A., Johnson, J., Fei-Fei, L., Savarese, S., Alahi, A.: Social gan:
  Socially acceptable trajectories with generative adversarial networks.
\newblock In: CVPR (2018)

\bibitem{HadfieldCVPR2013}
Hadfield, S., Bowden, R.: Hollywood 3d: Recognizing actions in 3d natural
  scenes.
\newblock In: CVPR. Portland, Oregon (2013)

\bibitem{Harris1988}
Harris, C., Stephens., M.: A combined corner and edge detector.
\newblock In: Alvey Vision Conference (1988)

\bibitem{HasanECCV2014}
Hasan, M., Roy-Chowdhury, A.K.: Continuous learning of human activity models
  using deep nets.
\newblock In: ECCV (2014)

\bibitem{ActivityNet2015}
{Heilbron}, F.C., {Escorcia}, V., {Ghanem}, B., {Niebles}, J.C.: {ActivityNet}:
  A large-scale video benchmark for human activity understanding.
\newblock In: CVPR (2015)

\bibitem{HerathIVC2017}
Herath, S., Harandi, M., Porikli, F.: Going deeper into action recognition: A
  survey.
\newblock Image and Vision Computing  (2017)

\bibitem{HoaiCVPR2012}
Hoai, M., la~Torre, F.D.: Max-margin early event detectors.
\newblock In: CVPR (2012)

\bibitem{Horn1981}
Horn, B., Schunck, B.: Determining optical flow.
\newblock Artificial Intelligence \textbf{17}, 185--203 (1981)

\bibitem{HuCVPR2015}
Hu, J.F., Zheng, W.S., Lai, J., Zhang, J.: Jointly learning heterogeneous
  features for rgb-d activity recognition.
\newblock In: CVPR (2015)

\bibitem{HuTIP2007}
Hu, W., Xie, D., Fu, Z., Zeng, W., Maybank, S.: Semantic-based surveillance
  video retrieval.
\newblock Image Processing, IEEE Transactions on \textbf{16}(4), 1168--1181
  (2007)

\bibitem{huang2016connectionist}
Huang, D.A., Fei-Fei, L., Niebles, J.C.: Connectionist temporal modeling for
  weakly supervised action labeling.
\newblock In: European Conference on Computer Vision, pp. 137--153. Springer
  (2016)

\bibitem{HuangECCV2008}
Huang, D.A., Kitani, K.M.: Action-reaction: Forecasting the dynamics of human
  interaction.
\newblock In: ECCV (2008)

\bibitem{IkizlerCVPR2007}
Ikizler, N., Forsyth, D.: Searching video for complex activities with finite
  state models.
\newblock In: CVPR (2007)

\bibitem{JainCVPR2014}
Jain, M., van Gemert, J., Jegou, H., Bouthemy, P., Snoek, C.G.: Action
  localization with tubelets from motion.
\newblock In: CVPR (2014)

\bibitem{JainCVPR2013}
Jain, M., J\'{e}gou, H., Bouthemy, P.: Better exploiting motion for better
  action recognition.
\newblock In: CVPR (2013)

\bibitem{JiICML2010}
Ji, S., Xu, W., Yang, M., Yu, K.: 3d convolutional neural networks for human
  action recognition.
\newblock In: ICML (2010)

\bibitem{JiTPAMI2013}
Ji, S., Xu, W., Yang, M., Yu, K.: 3d convolutional neural networks for human
  action recognition.
\newblock IEEE Trans. Pattern Analysis and Machine Intelligence  (2013)

\bibitem{JiaMM2014}
Jia, C., Kong, Y., Ding, Z., Fu, Y.: Latent tensor transfer learning for rgb-d
  action recognition.
\newblock In: ACM Multimedia (2014)

\bibitem{JiaCVPR2008}
Jia, K., Yeung, D.Y.: Human action recognition using local spatio-temporal
  discriminant embedding.
\newblock In: CVPR (2008)

\bibitem{THUMOS14}
Jiang, Y.G., Liu, J., Roshan~Zamir, A., Toderici, G., Laptev, I., Shah, M.,
  Sukthankar, R.: {THUMOS} challenge: Action recognition with a large number of
  classes.
\newblock \url{http://crcv.ucf.edu/THUMOS14/} (2014)

\bibitem{FCVID}
Jiang, Y.G., Wu, Z., Wang, J., Xue, X., Chang, S.F.: Exploiting feature and
  class relationships in video categorization with regularized deep neural
  networks.
\newblock {IEEE} Transactions on Pattern Analysis and Machine Intelligence
  \textbf{40}(2), 352--364 (2018).
\newblock \doi{10.1109/TPAMI.2017.2670560}.
\newblock \urlprefix\url{https://doi.org/10.1109/TPAMI.2017.2670560}

\bibitem{LiuCVPR2009a}
Jingen~Liu, J.L., Shah, M.: Recognizing realistic actions from videos "in the
  wild".
\newblock In: CVPR (2009)

\bibitem{red}
Jiyang~Gao Zhenheng~Yang, R.N.: Red: Reinforced encoder-decoder networks for
  action anticipation.
\newblock In: BMVC (2017)

\bibitem{KarCVPR2017}
Kar, A., Rai, N., Sikka, K., Sharma, G.: Adascan: Adaptive scan pooling in deep
  convolutional neural networks for human action recognition in videos.
\newblock In: CVPR (2017)

\bibitem{KaramanTHUMOS2014}
Karaman, S., Seidenari, L., Bimbo, A.D.: Fast saliency based pooling of fisher
  encoded dense trajectories.
\newblock In: ECCV THUMOS Workshop (2014)

\bibitem{KarpathyCVPR2014}
Karpathy, A., Toderici, G., Shetty, S., Leung, T., Sukthankar, R., Fei-Fei, L.:
  Large-scale video classification with convolutional neural networks.
\newblock In: CVPR (2014)

\bibitem{kay2017kinetics}
Kay, W., Carreira, J., Simonyan, K., Zhang, B., Hillier, C., Vijayanarasimhan,
  S., Viola, F., Green, T., Back, T., Natsev, P., et~al.: The kinetics human
  action video dataset.
\newblock arXiv preprint arXiv:1705.06950  (2017)

\bibitem{ke2017new}
Ke, Q., Bennamoun, M., An, S., Sohel, F., Boussaid, F.: A new representation of
  skeleton sequences for 3d action recognition.
\newblock In: Proceedings of the IEEE conference on computer vision and pattern
  recognition, pp. 3288--3297 (2017)

\bibitem{Qiuhong}
Ke, Q., Fritz, M., Schiele, B.: Time-conditioned action anticipation in one
  shot.
\newblock In: CVPR (2019)

\bibitem{ke2021future}
Ke, Q., Fritz, M., Schiele, B.: Future moment assessment for action query.
\newblock In: Proceedings of the IEEE/CVF Winter Conference on Applications of
  Computer Vision (2021)

\bibitem{Keestra2015}
Keestra, M.: Understanding human action. integraiting meanings, mechanisms,
  causes, and contexts.
\newblock TRANSDISCIPLINARITY IN PHILOSOPHY AND SCIENCE: APPROACHES, PROBLEMS,
  PROSPECTS pp. 201--235 (2015)

\bibitem{Soomro2012}
Khurram~Soomro, A.R.Z., Shah, M.: Ucf101: A dataset of 101 human action classes
  from videos in the wild (2012).
\newblock CRCV-TR-12-01

\bibitem{KimICCV2011}
Kim, K., Lee, D., Essa, I.: Gaussian process regression flow for analysis of
  motion trajectories.
\newblock In: ICCV (2011)

\bibitem{KitaniECCV2012}
Kitani, K.M., Ziebart, B.D., Bagnell, J.A., Hebert, M.: Activity forecasting.
\newblock In: ECCV (2012)

\bibitem{KlaserBMVC2008}
Klaser, A., Marszalek, M., Schmid, C.: A spatio-temporal descriptor based on
  3d-gradients.
\newblock In: BMVC (2008)

\bibitem{GrossPAMI2012}
Kliper-Gross, O., Hassner, T., Wolf, L.: The action similarity labeling
  challenge.
\newblock IEEE Transactions on Pattern Analysis and Machine Intelligence
  \textbf{34}(3) (2012)

\bibitem{KongECCVW2014}
Kong, Y., Fu, Y.: Modeling supporting regions for close human interaction
  recognition.
\newblock In: ECCV workshop (2014)

\bibitem{KongCVPR2015}
Kong, Y., Fu, Y.: Bilinear heterogeneous information machine for rgb-d action
  recognition.
\newblock In: CVPR (2015)

\bibitem{KongTPAMI2016}
Kong, Y., Fu, Y.: Max-margin action prediction machine.
\newblock TPAMI \textbf{38}(9), 1844 -- 1858 (2016)

\bibitem{KongIJCV2017}
Kong, Y., Fu, Y.: Max-margin heterogeneous information machine for rgb-d action
  recognition.
\newblock International Journal of Computer Vision (IJCV) \textbf{123}(3),
  350--371 (2017)

\bibitem{KongAAAI2018}
Kong, Y., Gao, S., Sun, B., Fu, Y.: Action prediction from videos via
  memorizing hard-to-predict samples.
\newblock In: AAAI (2018)

\bibitem{KongECCV2012}
Kong, Y., Jia, Y., Fu, Y.: Learning human interaction by interactive phrases.
\newblock In: Proc. European Conf. on Computer Vision (2012)

\bibitem{KongPAMI2014}
Kong, Y., Jia, Y., Fu, Y.: Interactive phrases: Semantic descriptions for human
  interaction recognition.
\newblock In: PAMI (2014)

\bibitem{KongECCV2014}
Kong, Y., Kit, D., Fu, Y.: A discriminative model with multiple temporal scales
  for action prediction.
\newblock In: ECCV (2014)

\bibitem{KongCVPR2017}
Kong, Y., Tao, Z., Fu, Y.: Deep sequential context networks for action
  prediction.
\newblock In: CVPR (2017)

\bibitem{KongTPAMI2018}
Kong, Y., Tao, Z., Fu, Y.: Adversarial action prediction networks.
\newblock IEEE TPAMI  (2018)

\bibitem{KooijECCV2014}
Kooij, J.F.P., Schneider, N., Flohr, F., Gavrila, D.M.: Context-based
  pedestrian path prediction.
\newblock In: European Conference on Computer Vision, pp. 618--633. Springer
  (2014)

\bibitem{Koppula2013}
Koppula, H.S., Gupta, R., Saxena, A.: Learning human activities and object
  affordances from rgb-d videos.
\newblock International Journal of Robotics Research  (2013)

\bibitem{robotic_antic}
Koppula, H.S., Saxena, A.: Anticipating human activities for reactive robotic
  response.
\newblock In: IROS (2013)

\bibitem{KoppulaICML2013}
Koppula, H.S., Saxena, A.: Learning spatio-temporal structure from rgb-d videos
  for human activity detection and anticipation.
\newblock In: ICML (2013)

\bibitem{KoppulaTPAMI2016}
Koppula, H.S., Saxena, A.: Anticipating human activities using object
  affordances for reactive robotic response.
\newblock IEEE transactions on pattern analysis and machine intelligence
  \textbf{38}(1), 14--29 (2016)

\bibitem{kosaraju2019social}
Kosaraju, V., Sadeghian, A., Mart{\'\i}n-Mart{\'\i}n, R., Reid, I.,
  Rezatofighi, S.H., Savarese, S.: Social-bigat: Multimodal trajectory
  forecasting using bicycle-gan and graph attention networks.
\newblock arXiv preprint arXiv:1907.03395  (2019)

\bibitem{KretzschmarICRA2014}
Kretzschmar, H., Kuderer, M., Burgard, W.: Learning to predict trajecteories of
  cooperatively navigation agents.
\newblock In: International Conference on Robotics and Automation (2014)

\bibitem{KuehneICCV2011}
Kuehne, H., Jhuang, H., Garrote, E., Poggio, T., Serre, T.: Hmdb: A large video
  database for human motion recognition.
\newblock In: ICCV (2011)

\bibitem{Kurakin2012}
Kurakin, A., Zhang, Z., Liu, Z.: A real-time system for dynamic hand gesture
  recognition with a depth sensor.
\newblock In: EUSIPCO (2012)

\bibitem{LaiTIP2018}
Lai, S., Zhang, W.S., Hu, J.F., Zhang, J.: Global-local temporal saliency
  action prediction.
\newblock IEEE Transactions on Image Processing \textbf{27}(5), 2272--2285
  (2018)

\bibitem{LanECCV2014}
Lan, T., Chen, T.C., Savarese, S.: A hierarchical representation for future
  action prediction.
\newblock In: European Conference on Computer Vision, pp. 689--704. Springer
  (2014)

\bibitem{LanCVPR2012}
Lan, T., Sigal, L., Mori, G.: Social roles in hierarchical models for human
  activity.
\newblock In: CVPR (2012)

\bibitem{LanPAMI2012}
Lan, T., Wang, Y., Yang, W., Robinovitch, S.N., Mori, G.: Discriminative latent
  models for recognizing contextual group activities.
\newblock TPAMI \textbf{34}(8), 1549--1562 (2012)

\bibitem{LaptevIJCV2005}
Laptev, I.: On space-time interest points.
\newblock IJCV \textbf{64}(2), 107--123 (2005)

\bibitem{LaptevICCV2003}
Laptev, I., Lindeberg, T.: Space-time interest points.
\newblock In: ICCV, pp. 432--439 (2003)

\bibitem{LaptevCVPR2008}
Laptev, I., Marszalek, M., Schmid, C., Rozenfeld, B.: Learning realistic human
  actions from movies.
\newblock In: CVPR (2008)

\bibitem{laptev2008learning}
Laptev, I., Marsza{\l}ek, M., Schmid, C., Rozenfeld, B.: Learning realistic
  human actions from movies (2008)

\bibitem{LaptevICCV2007}
Laptev, I., Perez, P.: Retrieving actions in movies.
\newblock In: ICCV (2007)

\bibitem{LeCVPR2011}
Le, Q.V., Zou, W.Y., Yeung, S.Y., Ng, A.Y.: Learning hierarchical invariant
  spatio-temporal features for action recognition with independent subspace
  analysis.
\newblock In: CVPR (2011)

\bibitem{lee2017unsupervised}
Lee, H.Y., Huang, J.B., Singh, M., Yang, M.H.: Unsupervised representation
  learning by sorting sequences.
\newblock In: Proceedings of the IEEE International Conference on Computer
  Vision, pp. 667--676 (2017)

\bibitem{LeeCVPR2017}
Lee, N., Choi, W., Vernaza, P., Choy, C.B., Torr, P.H., Chandraker, M.: Desire:
  Distant future prediction in dynamic scenes with interacting agents.
\newblock In: CVPR (2017)

\bibitem{LeeWACV2016}
Lee, N., Kitani, K.M.: Predicting wide receiver trajectories in american
  football.
\newblock In: WACV2016

\bibitem{li2019conditional}
Li, J., Ma, H., Tomizuka, M.: Conditional generative neural system for
  probabilistic trajectory prediction.
\newblock In: 2019 IEEE/RSJ International Conference on Intelligent Robots and
  Systems (IROS), pp. 6150--6156. IEEE (2019)

\bibitem{LiTPAMI2014}
Li, K., Fu, Y.: Prediction of human activity by discovering temporal sequence
  patterns.
\newblock IEEE Transactions on Pattern Analysis and Machine Intelligence
  \textbf{36}(8), 1644--1657 (2014)

\bibitem{LiECCV2012}
Li, K., Hu, J., Fu, Y.: Modeling complex temporal composition of actionlets for
  activity prediction.
\newblock In: ECCV (2012)

\bibitem{LiCVPRW2010}
Li, W., Zhang, Z., Liu, Z.: Action recognition based on a bag of 3d points.
\newblock In: CVPR workshop (2010)

\bibitem{multi-sports}
Li, Y., Chen, L., He, R., Wang, Z., Wu, G., Wang, L.: Multisports: {A}
  multi-person video dataset of spatio-temporally localized sports actions.
\newblock In: ICCV (2021)

\bibitem{three_birds}
Li, Z., Yao, L.: Three birds with one stone: Multi-task temporal action
  detection via recycling temporal annotations.
\newblock In: Proceedings of the IEEE/CVF Conference on Computer Vision and
  Pattern Recognition (CVPR), pp. 4751--4760 (2021)

\bibitem{liang2019peeking}
Liang, J., Jiang, L., Niebles, J.C., Hauptmann, A.G., Fei-Fei, L.: Peeking into
  the future: Predicting future person activities and locations in videos.
\newblock In: Proceedings of the IEEE/CVF Conference on Computer Vision and
  Pattern Recognition, pp. 5725--5734 (2019)

\bibitem{lin2019bmn}
Lin, T., Liu, X., Li, X., Ding, E., Wen, S.: Bmn: Boundary-matching network for
  temporal action proposal generation.
\newblock In: Proceedings of the IEEE/CVF International Conference on Computer
  Vision, pp. 3889--3898 (2019)

\bibitem{lin2018bsn}
Lin, T., Zhao, X., Su, H., Wang, C., Yang, M.: Bsn: Boundary sensitive network
  for temporal action proposal generation.
\newblock In: Proceedings of the European Conference on Computer Vision (ECCV),
  pp. 3--19 (2018)

\bibitem{LinCVPR2014}
Lin, Y.Y., Hua, J.H., Tang, N.C., Chen, M.H., Liao, H.Y.M.: Depth and skeleton
  associated action recognition without online accessible rgb-d cameras.
\newblock In: CVPR (2014)

\bibitem{LiuCVPR2011}
Liu, J., Kuipers, B., Savarese, S.: Recognizing human actions by attributes.
\newblock In: CVPR (2011)

\bibitem{LiuCVPR2009}
Liu, J., Luo, J., Shah, M.: Recognizing realistic actions from videos ``in the
  wild''.
\newblock In: Proc. IEEE Conf. on Computer Vision and Pattern Recognition
  (2009)

\bibitem{nturgb120}
Liu, J., Shahroudy, A., Perez, M., Wang, G., Duan, L.Y., Kot, A.C.: Ntu rgb+d
  120: A large-scale benchmark for 3d human activity understanding.
\newblock IEEE Transactions on Pattern Analysis and Machine Intelligence
  \textbf{42}(10), 2684--2701 (2020)

\bibitem{liu2016spatio}
Liu, J., Shahroudy, A., Xu, D., Wang, G.: Spatio-temporal lstm with trust gates
  for 3d human action recognition.
\newblock In: European Conference on Computer Vision, pp. 816--833. Springer
  (2016)

\bibitem{LiuIJCAI2013}
Liu, L., Shao, L.: Learning discriminative representations from rgb-d video
  data.
\newblock In: IJCAI (2013)

\bibitem{no_frames}
Liu, X., Pintea, S.L., Nejadasl, F.K., Booij, O., van Gemert, J.C.: No frame
  left behind: Full video action recognition.
\newblock In: CVPR (2021)

\bibitem{liu2019multi}
Liu, Y., Ma, L., Zhang, Y., Liu, W., Chang, S.F.: Multi-granularity generator
  for temporal action proposal.
\newblock In: Proceedings of the IEEE/CVF Conference on Computer Vision and
  Pattern Recognition, pp. 3604--3613 (2019)

\bibitem{liu2020social}
Liu, Y., Yan, Q., Alahi, A.: Social nce: Contrastive learning of socially-aware
  motion representations.
\newblock arXiv preprint arXiv:2012.11717  (2020)

\bibitem{LuCVPR2014}
Lu, C., Jia, J., Tang, C.K.: Range-sample depth feature for action recognition.
\newblock In: CVPR (2014)

\bibitem{Lucas1981}
Lucas, B.D., Kanade, T.: An iterative image registration technique with an
  application to stereo vision.
\newblock In: Proceedings of Imaging Understanding Workshop (1981)

\bibitem{luo2014learning}
Luo, G., Yang, S., Tian, G., Yuan, C., Hu, W., Maybank, S.J.: Learning human
  actions by combining global dynamics and local appearance.
\newblock IEEE Transactions on Pattern Analysis and Machine Intelligence
  \textbf{36}(12), 2466--2482 (2014)

\bibitem{LuoICCV2013}
Luo, J., Wang, W., Qi, H.: Group sparsity and geometry constrained dictionary
  learning for action recognition from depth maps.
\newblock In: ICCV (2013)

\bibitem{LuoECCV2018}
Luo, Z., Hsieh, J.T., Jiang, L., Carlos~Niebles, J., Fei-Fei, L.: Graph
  distillation for action detection with privileged modalities.
\newblock In: ECCV (2018)

\bibitem{MaCVPR2016}
Ma, S., Sigal, L., Sclaroff, S.: Learning activity progression in lstms for
  activity detection and early detection.
\newblock In: CVPR (2016)

\bibitem{Mainprice2016}
Mainprice, J., Hayne, R., Berenson, D.: Goal set inverse optimal control and
  iterative re-planning for predicting human reaching motions in shared
  workspace.
\newblock In: arXiv preprint arXiv:1606.02111 (2016)

\bibitem{mangalam2020goals}
Mangalam, K., An, Y., Girase, H., Malik, J.: From goals, waypoints \& paths to
  long term human trajectory forecasting.
\newblock arXiv preprint arXiv:2012.01526  (2020)

\bibitem{mangalam2020not}
Mangalam, K., Girase, H., Agarwal, S., Lee, K.H., Adeli, E., Malik, J., Gaidon,
  A.: It is not the journey but the destination: Endpoint conditioned
  trajectory prediction.
\newblock In: European Conference on Computer Vision, pp. 759--776. Springer
  (2020)

\bibitem{marchetti2020mantra}
Marchetti, F., Becattini, F., Seidenari, L., Bimbo, A.D.: Mantra: Memory
  augmented networks for multiple trajectory prediction.
\newblock In: Proceedings of the IEEE/CVF Conference on Computer Vision and
  Pattern Recognition, pp. 7143--7152 (2020)

\bibitem{MarszalekCVPR2009}
Marsza{\l}ek, M., Laptev, I., Schmid, C.: Actions in context.
\newblock In: IEEE Conference on Computer Vision \& Pattern Recognition (2009)

\bibitem{Mass1971}
Mass, J., Johansson, G., Jason, G., Runeson, S.: Motion perception I and II
  [film].
\newblock Boston: Houghton Mifflin (1971)

\bibitem{vae_anticip}
Mehrasa, N., Jyothi, A.A., Durand, T., He, J., Sigal, L., Mori, G.: A
  variational auto-encoder model for stochastic point processes.
\newblock In: CVPR (2019)

\bibitem{MessingICCV2009}
Messing, R., Pal, C., Kautz, H.: Activity recognition using the velocity
  histories of tracked keypoints.
\newblock In: ICCV (2009)

\bibitem{gao2021woad}
Mingfei~Gao Yingbo~Zhou, R.X.R.S.C.X.: Woad: Weakly supervised online action
  detection in untrimmed videos.
\newblock In: CVPR (2021)

\bibitem{ashish_few_shot}
Mishra, A., Verma, V., Reddy, M.K.K., Subramaniam, A., Rai, P., Mittal, A.: A
  generative approach to zero-shot and few-shot action recognition (2018)

\bibitem{misra2016shuffle}
Misra, I., Zitnick, C.L., Hebert, M.: Shuffle and learn: unsupervised learning
  using temporal order verification.
\newblock In: European Conference on Computer Vision, pp. 527--544. Springer
  (2016)

\bibitem{mohamed2020social}
Mohamed, A., Qian, K., Elhoseiny, M., Claudel, C.: Social-stgcnn: A social
  spatio-temporal graph convolutional neural network for human trajectory
  prediction.
\newblock In: Proceedings of the IEEE/CVF Conference on Computer Vision and
  Pattern Recognition, pp. 14424--14432 (2020)

\bibitem{monfortmoments}
Monfort, M., Zhou, B., Bargal, S.A., Yan, T., Andonian, A., Ramakrishnan, K.,
  Brown, L., Fan, Q., Gutfruend, D., Vondrick, C., et~al.: Moments in time
  dataset: one million videos for event understanding

\bibitem{MorencyCVPR2007}
Morency, L.P., Quattoni, A., Darrell, T.: Latent-dynamic discriminative models
  for continuous gesture recognition.
\newblock In: CVPR (2007)

\bibitem{MorrisandTPAMI2011}
Morrisand, B., Trivedi, M.: Trajectory learning for activity understanding:
  Unsupervised, multilevel, and long-term adaptive approach.
\newblock Pattern Analysis and Machine Intelligence, IEEE Transactions on
  \textbf{33}(11), 2287--2301 (2011)

\bibitem{NarayanICCV2019}
Narayan, S., Cholakkal, H., Khan, F.S., Shao, L.: {3C-Net}: Category count and
  center loss for weakly-supervised action localization.
\newblock In: ICCV (2019)

\bibitem{narayanan2021divide}
Narayanan, S., Moslemi, R., Pittaluga, F., Liu, B., Chandraker, M.:
  Divide-and-conquer for lane-aware diverse trajectory prediction.
\newblock In: CVPR (2021)

\bibitem{NgCVPR2015}
Ng, J.Y.H., Hausknecht, M., Vijayanarasimhan, S., Vinyals, O., Monga, R.,
  Toderici, G.: Beyond short snippets: Deep networks for video classification.
\newblock In: CVPR (2015)

\bibitem{Ni2011}
Ni, B., Wang, G., Moulin, P.: {RGBD-HuDaAct}: A color-depth video database for
  human daily activity recognition.
\newblock In: ICCV Workshop on CDC3CV (2011)

\bibitem{NieblesECCV2010}
Niebles, J.C., Chen, C.W., Fei-Fei, L.: Modeling temporal structure of
  decomposable motion segments for activity classification.
\newblock In: ECCV (2010)

\bibitem{NieblesCVPR2007}
Niebles, J.C., Fei-Fei, L.: A hierarchical model of shape and appearance for
  human action classification.
\newblock In: CVPR (2007)

\bibitem{NieblesIJCV2008}
Niebles, J.C., Wang, H., Fei-Fei, L.: Unsupervised learning of human action
  categories using spatial-temporal words.
\newblock International Journal of Computer Vision \textbf{79}(3), 299--318
  (2008)

\bibitem{Ofli2013}
Ofli, F., Chaudhry, R., Kurillo, G., Vidal, R., Bajcsy, R.: Berkeley mhad: A
  comprehensive multimodal human action database.
\newblock In: Proceedings of the IEEE Workshop on Applications on Computer
  Vision (2013)

\bibitem{OliverPAMI2000}
Oliver, N.M., Rosario, B., Pentland, A.P.: A bayesian computer vision system
  for modeling human interactions.
\newblock PAMI \textbf{22}(8), 831--843 (2000)

\bibitem{OreifejCVPR2013}
Oreifej, O., Liu, Z.: Hon4d: Histogram of oriented 4d normals for activity
  recognition from depth sequences.
\newblock In: CVPR (2013)

\bibitem{OzaCVPR2019}
Oza, P., Patel, V.M.: {C2AE}: Class conditioned auto-encoder for open-set
  recognition.
\newblock In: CVPR (2019)

\bibitem{PerezPAMI2012}
Patron-Perez, A., Marszalek, M., Reid, I., Zissermann, A.: Structured learning
  of human interaction in tv shows.
\newblock PAMI \textbf{34}(12), 2441--2453 (2012)

\bibitem{PerezBMVC2010}
Patron-Perez, A., Marszalek, M., Zisserman, A., Reid, I.: High five:
  Recognising human interactions in tv shows.
\newblock In: Proc. British Conference on Machine Vision (2010)

\bibitem{PeiICCV2011}
Pei, M., Jia, Y., Zhu, S.C.: Parsing video events with goal inference and
  intent prediction.
\newblock In: ICCV, pp. 487--494. IEEE (2011)

\bibitem{PereraCVPR2020}
Perera, P., Morariu, V.I., Jain, R., Manjunatha, V., Wigington, C., Ordonez,
  V., Patel, V.M.: Generative-discriminative feature representations for
  open-set recognition.
\newblock In: CVPR (2020)

\bibitem{TRX}
Perrett, T., Masullo, A., Burghardt, T., Mirmehdi, M., Damen, D.:
  Temporal-relational crosstransformers for few-shot action recognition.
\newblock In: CVPR (2021)

\bibitem{PerronninCVPR2006}
Perronnin, F., Dance, C.: Fisher kernels on visual vocabularies for image
  categorization.
\newblock In: CVPR (2006)

\bibitem{PlotzIJCAI2011}
Plotz, T., Hammerla, N.Y., Olivier, P.: Feature learning for activity
  recognition in ubiquitous computing.
\newblock In: IJCAI (2011)

\bibitem{PoppeIVC2010}
Poppe, R.: A survey on vision-based human action recognition.
\newblock Image and Vision Computing \textbf{28}, 976--990 (2010)

\bibitem{purushwalkam2016pose}
Purushwalkam, S., Gupta, A.: Pose from action: Unsupervised learning of pose
  features based on motion.
\newblock arXiv preprint arXiv:1609.05420  (2016)

\bibitem{QiuICCV2017}
Qiu, Z., Yao, T., Mei, T.: Learning spatio-temporal representation with
  pseudo-3d residual network.
\newblock In: ICCV (2017)

\bibitem{qiu2019learning}
Qiu, Z., Yao, T., Ngo, C.W., Tian, X., Mei, T.: Learning spatio-temporal
  representation with local and global diffusion.
\newblock In: Proceedings of the IEEE Conference on Computer Vision and Pattern
  Recognition, pp. 12056--12065 (2019)

\bibitem{RajkoCVPR2007}
Rajko, S., Qian, G., Ingalls, T., James, J.: Real-time gesture recognition with
  minimal training requirements and on-line learning.
\newblock In: CVPR (2007)

\bibitem{RamanathanCVPR2013}
Ramanathan, V., Yao, B., Fei-Fei, L.: Social role discovery in human events.
\newblock In: CVPR (2013)

\bibitem{Ramezani2016}
Ramezani, M., Yaghmaee, F.: A review on human action analysis in videos for
  retrieval applications.
\newblock Artificial Intelligence Review \textbf{46}(4), 485--514 (2016)

\bibitem{RaptisCVPR2013}
Raptis, M., Sigal, L.: Poselet key-framing: A model for human activity
  recognition.
\newblock In: CVPR (2013)

\bibitem{RaptisECCV2010}
Raptis, M., Soatto, S.: Tracklet descriptors for action modeling and video
  analysis.
\newblock In: ECCV (2010)

\bibitem{rasouli2021bifold}
Rasouli, A., Rohani, M., Luo, J.: Bifold and semantic reasoning for pedestrian
  behavior prediction.
\newblock In: CVPR (2021)

\bibitem{ReddyMVA2012}
Reddy, K.K., Shah, M.: Recognizing 50 human action categories of web videos.
\newblock Machine Vision and Applications Journal  (2012)

\bibitem{Ricoeur1992}
Ricoeur, P.: Oneself as another (K. Blamey, Trans.).
\newblock Chicago: University of Chicago Press (1992)

\bibitem{Rizzolatti2004}
Rizzolatti, G., Craighero, L.: The mirror-neuron system.
\newblock Annu. Rev. Neurosci. \textbf{27}, 169--192 (2004)

\bibitem{Rizzolatti2010}
Rizzolatti, G., Sinigaglia, C.: The functional role of the parieto-frontal
  mirror circuit: interpretations and misinterpretations.
\newblock Nat. Rev. Neurosci. \textbf{11}, 264--274 (2010)

\bibitem{RodriguezCVPR2008}
Rodriguez, M.D., Ahmed, J., Shah, M.: Action mach: A spatio-temporal maximum
  average correlation height filter for action recognition.
\newblock In: CVPR (2008)

\bibitem{Rohit21}
Rohit, G., Kristen, G.: Anticipative video transformer.
\newblock In: ICCV (2021)

\bibitem{RoitbergIV2020}
Roitberg, A., Ma, C., Haurilet, M., Stiefelhagen, R.: Open set driver activity
  recognition.
\newblock In: IVS (2020)

\bibitem{RyooCVPR2006}
Ryoo, M., Aggarwal, J.: Recognition of composite human activities through
  context-free grammar based representation.
\newblock In: CVPR, vol.~2, pp. 1709--1718 (2006)

\bibitem{RyooICCV2009}
Ryoo, M., Aggarwal, J.: Spatio-temporal relationship match: Video structure
  comparison for recognition of complex human activities.
\newblock In: ICCV, pp. 1593--1600 (2009)

\bibitem{RyooIJCV2011}
Ryoo, M., Aggarwal, J.: Stochastic representation and recognition of high-level
  group activities.
\newblock IJCV \textbf{93}, 183--200 (2011)

\bibitem{RyooHRI2015}
Ryoo, M., Fuchs, T.J., Xia, L., Aggarwal, J.K., Matthies, L.: Robot-centric
  activity prediction from first-person videos: What will they do to me?
\newblock In: Proceedings of the Tenth Annual ACM/IEEE International Conference
  on Human-Robot Interaction, pp. 295--302. ACM (2015)

\bibitem{RyooICCV2011}
Ryoo, M.S.: Human activity prediction: Early recognition of ongoing activities
  from streaming videos.
\newblock In: ICCV (2011)

\bibitem{UT-Interaction-Data}
Ryoo, M.S., Aggarwal, J.K.: {UT}-{I}nteraction {D}ataset, {ICPR} contest on
  {S}emantic {D}escription of {H}uman {A}ctivities ({SDHA}).
\newblock http://cvrc.ece.utexas.edu/SDHA2010/Human\_Interaction.html (2010)

\bibitem{MuHAVi2010}
S~Singh, S.V., Ragheb, H.: Muhavi: A multicamera human action video dataset for
  the evaluation of action recognition methods.
\newblock In: 2nd Workshop on Activity monitoring by multi-camera surveillance
  systems (AMMCSS), pp. 48--55 (2010)

\bibitem{sadeghian2019sophie}
Sadeghian, A., Kosaraju, V., Sadeghian, A., Hirose, N., Rezatofighi, H.,
  Savarese, S.: Sophie: An attentive gan for predicting paths compliant to
  social and physical constraints.
\newblock In: Proceedings of the IEEE/CVF Conference on Computer Vision and
  Pattern Recognition, pp. 1349--1358 (2019)

\bibitem{SatkinECCV2010}
Satkin, S., Hebert, M.: Modeling the temporal extent of actions.
\newblock In: ECCV (2010)

\bibitem{ScheirerTPAMI2014}
Scheirer, W.J., Jain, L.P., Boult, T.E.: Probability models for open set
  recognition.
\newblock IEEE transactions on pattern analysis and machine intelligence
  \textbf{36}(11), 2317--2324 (2014)

\bibitem{ScheirerTPAMI2012}
Scheirer, W.J., de~Rezende~Rocha, A., Sapkota, A., Boult, T.E.: Toward open set
  recognition.
\newblock IEEE transactions on pattern analysis and machine intelligence
  \textbf{35}(7), 1757--1772 (2012)

\bibitem{SchuldtICPR2004}
Sch\"{u}ldt, C., Laptev, I., Caputo, B.: Recognizing human actions: A local svm
  approach.
\newblock In: IEEE ICPR (2004)

\bibitem{Scovanner2007}
Scovanner, P., Ali, S., Shah, M.: A 3-dimensional sift descriptor and its
  application to action recognition.
\newblock In: Proc. ACM Multimedia (2007)

\bibitem{Shahroudy_2016_NTURGBD}
Shahroudy, A., Liu, J., Ng, T.T., Wang, G.: Ntu rgb+d: A large scale dataset
  for 3d human activity analysis.
\newblock In: IEEE Conference on Computer Vision and Pattern Recognition (2016)

\bibitem{nturgb60}
Shahroudy, A., Liu, J., Ng, T.T., Wang, G.: Ntu rgb+d: A large scale dataset
  for 3d human activity analysis.
\newblock In: CVPR (2016)

\bibitem{ShiIJCV2011}
Shi, Q., Cheng, L., Wang, L., Smola, A.: Human action segmentation and
  recognition using discriminative semi-markov models.
\newblock IJCV \textbf{93}, 22--32 (2011)

\bibitem{ShottonPAMI2013}
Shotton, J., Girshick, R., Fitzgibbon, A., Sharp, T., Cook, M., Finocchio, M.,
  Moore, R., Kohli, P., Criminisi, A., Kipman, A., Blake, A.: Efficient human
  pose estimation from single depth images.
\newblock PAMI  (2013)

\bibitem{ShouCVPR2017}
Shou, Z., Chan, J., Zareian, A., Miyazawa, K., Chang, S.F.: {CDC}:
  Convolutional-de-convolutional networks for precise temporal action
  localization in untrimmed videos.
\newblock In: CVPR (2017)

\bibitem{ShouCVPR2016}
Shou, Z., Wang, D., Chang, S.F.: Temporal action localization in untrimmed
  videos via multi-stage {CNNs}.
\newblock In: CVPR (2016)

\bibitem{ShuICME2018}
Shu, Y., Shi, Y., Wang, Y., Zou, Y., Yuan, Q., Tian, Y.: {ODN}: Opening the
  deep network for open-set action recognition.
\newblock In: ICME (2018)

\bibitem{si2019attention}
Si, C., Chen, W., Wang, W., Wang, L., Tan, T.: An attention enhanced graph
  convolutional lstm network for skeleton-based action recognition.
\newblock In: Proceedings of the IEEE Conference on Computer Vision and Pattern
  Recognition, pp. 1227--1236 (2019)

\bibitem{SimonyanNIPS2014}
Simonyan, K., Zisserman, A.: Two-stream convolutional networks for action
  recognition in videos.
\newblock In: NIPS (2014)

\bibitem{Singh2010}
Singh, S., Velastin, S.A., Ragheb, H.: Muhavi: A multicamera human action video
  dataset for the evaluation of action recognition methods.
\newblock In: Advanced Video and Signal Based Surveillance (AVSS), 2010 Seventh
  IEEE International Conference on, pp. 48--55. IEEE (2010)

\bibitem{SminchisescuICCV2005}
Sminchisescu, C., Kanaujia, A., Li, Z., Metaxas, D.: Conditional models for
  contextual human motion recognition.
\newblock In: International Conference on Computer Vision (2005)

\bibitem{SongTMM2019}
Song, H., Wu, X., Zhu, B., Wu, Y., Chen, M., Jia, Y.: Temporal action
  localization in untrimmed videos using action pattern trees.
\newblock IEEE Transactions on Multimedia (TMM) \textbf{21}(3), 717--730 (2019)

\bibitem{SongCVPR2019}
Song, L., Zhang, S., Yu, G., Sun, H.: {TACNet}: Transition-aware context
  network for spatio-temporal action detection.
\newblock In: CVPR (2019)

\bibitem{SongTIP2018}
Song, S., Lan, C., Xing, J., Zeng, W., Liu, J.: Spatio-temporal attention-based
  {LSTM} networks for 3d action recognition and detection.
\newblock IEEE Transactions on Image Processing (TIP) \textbf{27}(7),
  3459--3471 (2018)

\bibitem{SuIJCAI2017}
Su, H., Zhu, J., Dong, Y., Zhang, B.: Forecast the plausible paths in crowd
  scenes.
\newblock In: IJCAI (2017)

\bibitem{Sumi2000}
Sumi, S.: Perception of point-light walker produced by eight lights attached to
  the back of the walker.
\newblock Swiss J. Psychol. \textbf{59}, 126--32 (2000)

\bibitem{SunCVPR2010}
Sun, D., Roth, S., Black, M.J.: Secrets of optical flow estimation and their
  principles.
\newblock In: CVPR (2010)

\bibitem{SunCVPR2009}
Sun, J., Wu, X., Yan, S., Cheong, L., Chua, T., Li, J.: Hierarchical
  spatio-temporal context modeling for action recognition.
\newblock In: CVPR (2009)

\bibitem{SunCVPR2014}
Sun, L., Jia, K., Chan, T.H., Fang, Y., Wang, G., Yan, S.: Dl-sfa:
  Deeply-learned slow feature analysis for action recognition.
\newblock In: CVPR (2014)

\bibitem{SungAAAIW2011}
Sung, J., Ponce, C., Selman, B., Saxena, A.: Human activity detection from rgbd
  images.
\newblock In: AAAI workshop on Pattern, Activity and Intent Recognition (2011)

\bibitem{SungICRA2012}
Sung, J., Ponce, C., Selman, B., Saxena, A.: Unstructured human activity
  detection from rgbd images.
\newblock In: ICRA (2012)

\bibitem{suris2021learning}
Sur{\'\i}s, D., Liu, R., Vondrick, C.: Learning the predictability of the
  future.
\newblock In: CVPR (2021)

\bibitem{TangCVPR2012}
Tang, K., Fei-Fei, L., Koller, D.: Learning latent temporal structure for
  complex event detection.
\newblock In: CVPR (2012)

\bibitem{TangNIPS2012}
Tang, K., Ramanathan, V., Fei-Fei, L., Koller, D.: Shifting weights: Adapting
  object detectors from image to video.
\newblock In: Advances in Neural Information Processing Systems (2012)

\bibitem{COIN2019}
Tang, Y., Ding, D., Rao, Y., Zheng, Y., Zhang, D., Zhao, L., Lu, J., Zhou, J.:
  {COIN}: A large-scale dataset for comprehensive instructional video analysis.
\newblock In: CVPR (2019)

\bibitem{TaylorECCV2010}
Taylor, G.W., Fergus, R., LeCun, Y., Bregler, C.: Convolutional learning of
  spatio-temporal features.
\newblock In: ECCV (2010)

\bibitem{TranICCV2015}
Tran, D., Bourdev, L., Fergus, R., Torresani, L., Paluri, M.: Learning
  spatiotemporal features with 3d convolutional networks.
\newblock In: ICCV (2015)

\bibitem{TranECCV2008}
Tran, D., Sorokin, A.: Human activity recognition with metric learning.
\newblock In: ECCV (2008)

\bibitem{Troje2002}
Troje, N.: Decomposing biological motion: a framework for analysis and
  synthesis of human gait patterns.
\newblock J. Vis. \textbf{2}, 371--87 (2002)

\bibitem{Troje2005}
Troje, N., Westhoff, C., Lavrov, M.: Person identification from biological
  motion: effects of structural and kinematic cues.
\newblock Percept. Psychophys \textbf{67}, 667--75 (2005)

\bibitem{TurekECCV2010}
Turek, M., Hoogs, A., Collins, R.: Unsupervised learning of functional
  categories in video scenes.
\newblock In: ECCV (2010)

\bibitem{Vahdat2011}
Vahdat, A., Gao, B., Ranjbar, M., Mori, G.: A discriminative key pose sequence
  model for recognizing human interactions.
\newblock In: ICCV Workshops, pp. 1729 --1736 (2011)

\bibitem{VarolTPAMI2017}
Varol, G., Laptev, I., Schmid, C.: Long-term temporal convolutions for action
  recognition.
\newblock IEEE Transactions on Pattern Analysis and Machine Intelligence
  (2017)

\bibitem{Vondrick}
Vondrick, C., Pirsiavash, H., Torralba, A.: Anticipating visual representations
  from unlabeled video.
\newblock In: CVPR (2016)

\bibitem{WalkerCVPR2014}
Walker, J., Gupta, A., Hebert, M.: Patch to the future: Unsupervised visual
  prediction.
\newblock In: Proceedings of the IEEE Conference on Computer Vision and Pattern
  Recognition, pp. 3302--3309 (2014)

\bibitem{wang2021stepwise}
Wang, C., Wang, Y., Xu, M., Crandall, D.J.: Stepwise goal-driven networks for
  trajectory prediction.
\newblock arXiv preprint arXiv:2103.14107  (2021)

\bibitem{WangIJCV2013}
Wang, H., Kla$\:{a}$ser, A., Schmid, C., Liu, C.L.: Dense trajectories and
  motion boundary descriptors for action recognition.
\newblock IJCV \textbf{103}(60-79) (2013)

\bibitem{WangCVPR2011}
Wang, H., Kl{\"a}ser, A., Schmid, C., Liu, C.L.: {Action Recognition by Dense
  Trajectories}.
\newblock In: IEEE Conference on Computer Vision \& Pattern Recognition, pp.
  3169--3176. Colorado Springs, United States (2011).
\newblock \urlprefix\url{http://hal.inria.fr/inria-00583818/en}

\bibitem{WangIJCV2015}
Wang, H., Oneata, D., Verbeek, J., Schmid, C.: A robust and efficient video
  representation for action recognition.
\newblock IJCV  (2015)

\bibitem{WangICCV2013}
Wang, H., Schmid, C.: Action recognition with improved trajectories.
\newblock In: IEEE International Conference on Computer Vision. Sydney,
  Australia (2013).
\newblock \urlprefix\url{http://hal.inria.fr/hal-00873267}

\bibitem{WangBMVC2009}
Wang, H., Ullah, M.M., Kl$\:{a}$ser, A., Laptev, I., Schmid, C.: Evaluation of
  local spatio-temporal features for action recognition.
\newblock In: BMVC (2009)

\bibitem{WangECCV2012}
Wang, J., Liu, Z., Chorowski, J., Chen, Z., Wu, Y.: Robust 3d action
  recognition with random occupancy patterns.
\newblock In: ECCV (2012)

\bibitem{WangJCVPR2012}
Wang, J., Liu, Z., Wu, Y., Yuan, J.: Mining actionlet ensemble for action
  recognition with depth cameras.
\newblock In: CVPR (2012)

\bibitem{WangMM2014}
Wang, K., Wang, X., Lin, L., Wang, M., Zuo, W.: 3d human activity recognition
  with reconfigurable convolutional neural networks.
\newblock In: ACM Multimedia (2014)

\bibitem{WangTHUMOS2014}
Wang, L., Qiao, Y., Tang, X.: Action recognition and detection by combining
  motion and appearance features.
\newblock In: ECCV THUMOS Workshop (2014)

\bibitem{WangCVPR2015}
Wang, L., Qiao, Y., Tang, X.: Action recognition with trajectory-pooled
  deep-convolutional descriptors.
\newblock In: CVPR (2015)

\bibitem{WangCVPR2007}
Wang, L., Suter, D.: Recognizing human activities from silhouettes: Motion
  subspace and factorial discriminative graphical model.
\newblock In: CVPR (2007)

\bibitem{tdn}
Wang, L., Tong, Z., Ji, B., Wu, G.: Tdn: Temporal difference networks for
  efficient action recognition.
\newblock In: CVPR, pp. 1895--1904 (2021)

\bibitem{WangCVPR2017}
Wang, L., Xiong, Y., Lin, D., Van~Gool, L.: {UntrimmedNets} for weakly
  supervised action recognition and detection.
\newblock In: CVPR (2017)

\bibitem{WangECCV2016}
Wang, L., Xiong, Y., Wang, Z., Qiao, Y., Lin, D., Tang, X., Gool, L.V.: Temoral
  segment networks: Toward good practices for deep action recognition.
\newblock In: ECCV (2016)

\bibitem{WangCVPR2006}
Wang, S.B., Quattoni, A., Morency, L.P., Demirdjian, D., Darrell, T.: Hidden
  conditional random fields for gesture recognition.
\newblock In: CVPR (2006)

\bibitem{wang2015unsupervised}
Wang, X., Gupta, A.: Unsupervised learning of visual representations using
  videos.
\newblock In: Proceedings of the IEEE International Conference on Computer
  Vision, pp. 2794--2802 (2015)

\bibitem{wang2017transitive}
Wang, X., He, K., Gupta, A.: Transitive invariance for self-supervised visual
  representation learning.
\newblock In: Proceedings of the IEEE international conference on computer
  vision, pp. 1329--1338 (2017)

\bibitem{WangNIPS2008}
Wang, Y., Mori, G.: Learning a discriminative hidden part model for human
  action recognition.
\newblock In: NIPS (2008)

\bibitem{WangPAMI2010}
Wang, Y., Mori, G.: Hidden part models for human action recognition:
  Probabilistic vs. max-margin.
\newblock PAMI  (2010)

\bibitem{WangCVPR2012}
Wang, Z., Wang, J., Xiao, J., Lin, K.H., Huang, T.S.: Substructural and
  boundary modeling for continuous action recognition.
\newblock In: CVPR (2012)

\bibitem{WeinlandCVIU2006}
Weinland, D., Ronfard, R., Boyer, E.: Free viewpoint action recognition using
  motion history volumes.
\newblock Computer Vision and Image Understanding \textbf{104}(2-3), 249--257
  (2006)

\bibitem{WillemsECCV2008}
Willems, G., Tuytelaars, T., Gool, L.: An efficient dense and scale-invariant
  spatio-temporal interest poing detector.
\newblock In: ECCV (2008)

\bibitem{wolf2014evaluation}
Wolf, C., Lombardi, E., Mille, J., Celiktutan, O., Jiu, M., Dogan, E., Eren,
  G., Baccouche, M., Dellandr{\'e}a, E., Bichot, C.E., et~al.: Evaluation of
  video activity localizations integrating quality and quantity measurements.
\newblock Computer Vision and Image Understanding \textbf{127}, 14--30 (2014)

\bibitem{WongCVPR2007}
Wong, S.F., Kim, T.K., Cipolla, R.: Learning motion categories using both
  semantic and structural information.
\newblock In: CVPR (2007)

\bibitem{wu2014human}
Wu, B., Yuan, C., Hu, W.: Human action recognition based on context-dependent
  graph kernels.
\newblock In: Proceedings of the IEEE Conference on Computer Vision and Pattern
  Recognition, pp. 2609--2616 (2014)

\bibitem{WuNIPS2015}
Wu, J., Yildirim, I., Lim, J.J., Freeman, W.T., Tenenbaum, J.B.: Galileo:
  Perceiving physical object properties by integrating a physics engine with
  deep learning.
\newblock In: Advances in Neural Information Processing Systems, pp. 127--135
  (2015)

\bibitem{WuCVPR2011}
Wu, X., Xu, D., Duan, L., Luo, J.: Action recognition using context and
  appearance distribution features.
\newblock In: CVPR (2011)

\bibitem{WuMM2015}
Wu, Z., Wang, X., Jiang, Y.G., Ye, H., Xue, X.: Modeling spatial-temporal clues
  in a hybrid deep learning framework for video classification.
\newblock In: ACM Multimedia (2015)

\bibitem{Wulfmeier2016}
Wulfmeier, M., Wang, D., Posner, I.: Watch this: scalable cost function
  learning for path planning in urban environment.
\newblock In: arXiv preprint arXiv:1607:02329 (2016)

\bibitem{XiaCVPR2013}
Xia, L., Aggarwal, J.: Spatio-temporal depth cuboid similarity feature for
  activity recognition using depth camera.
\newblock In: CVPR (2013)

\bibitem{XiaCVPRW2012}
Xia, L., Chen, C., Aggarwal, J.: View invariant human action recognition using
  histograms of 3d joints.
\newblock In: Computer Vision and Pattern Recognition Workshops (CVPRW), 2012
  IEEE Computer Society Conference on, pp. 20--27. IEEE (2012)

\bibitem{utkinect-action3d}
Xia, L., Chen, C.C., Aggarwal, J.K.: View invariant human action recognition
  using histograms of 3d joints.
\newblock In: CVPRW (2012)

\bibitem{xu2019self}
Xu, D., Xiao, J., Zhao, Z., Shao, J., Xie, D., Zhuang, Y.: Self-supervised
  spatiotemporal learning via video clip order prediction.
\newblock In: Proceedings of the IEEE Conference on Computer Vision and Pattern
  Recognition, pp. 10334--10343 (2019)

\bibitem{xu2017r}
Xu, H., Das, A., Saenko, K.: R-c3d: Region convolutional 3d network for
  temporal activity detection.
\newblock In: Proceedings of the IEEE international conference on computer
  vision, pp. 5783--5792 (2017)

\bibitem{xu2019two}
Xu, H., Das, A., Saenko, K.: Two-stream region convolutional 3d network for
  temporal activity detection.
\newblock IEEE transactions on pattern analysis and machine intelligence
  \textbf{41}(10), 2319--2332 (2019)

\bibitem{XuICCV2019}
Xu, M., Gao, M., Chen, Y.T., Davis, L.S., Crandall, D.J.: Temporal recurrent
  networks for online action detection.
\newblock In: ICCV (2019)

\bibitem{yan2018spatial}
Yan, S., Xiong, Y., Lin, D.: Spatial temporal graph convolutional networks for
  skeleton-based action recognition.
\newblock In: Thirty-Second AAAI Conference on Artificial Intelligence (2018)

\bibitem{YangCVPR2018}
Yang, H., He, X., Porikli, F.: One-shot action localization by learning
  sequence matching network.
\newblock In: CVPR (2018)

\bibitem{yang2015multi-feature}
Yang, S., Yuan, C., Wu, B., Hu, W., Wang, F.: Multi-feature max-margin
  hierarchical bayesian model for action recognition.
\newblock In: Proceedings of the IEEE Conference on Computer Vision and Pattern
  Recognition, pp. 1610--1618 (2015)

\bibitem{uncertainty_action_detection}
Yang, W., Zhang, T., Yu, X., Qi, T., Zhang, Y., Wu, F.: Uncertainty guided
  collaborative training for weakly supervised temporal action detection.
\newblock In: CVPR (2021)

\bibitem{YangCVPR2014}
Yang, X., Tian, Y.: Super normal vector for activity recognition using depth
  sequences.
\newblock In: CVPR (2014)

\bibitem{YangCVPR2019}
Yang, X., Yang, X., Liu, M.Y., Xiao, F., Davis, L.S., Kautz, J.: {STEP}:
  Spatio-temporal progressive learning for video action detection.
\newblock In: CVPR (2019)

\bibitem{YangPR2019}
Yang, Y., Hou, C., Lang, Y., Guan, D., Huang, D., Xu, J.: Open-set human
  activity recognition based on micro-doppler signatures.
\newblock Pattern Recognition \textbf{85}, 60--69 (2019)

\bibitem{YangECCV2012}
Yang, Y., Shah, M.: Complex events detection using data-driven concepts.
\newblock In: ECCV (2012)

\bibitem{YaoECCV2012}
Yao, B., Fei-Fei, L.: Action recognition with exemplar based 2.5d graph
  matching.
\newblock In: ECCV (2012)

\bibitem{YaoPAMI2012}
Yao, B., Fei-Fei, L.: Recognizing human-object interactions in still images by
  modeling the mutual context of objects and human poses.
\newblock TPAMI \textbf{34}(9), 1691--1703 (2012)

\bibitem{YeffetCVPR2009}
Yeffet, L., Wolf, L.: Local trinary patterns for human action recognition.
\newblock In: CVPR (2009)

\bibitem{yeung2016end}
Yeung, S., Russakovsky, O., Mori, G., Fei-Fei, L.: End-to-end learning of
  action detection from frame glimpses in videos.
\newblock In: Proceedings of the IEEE conference on computer vision and pattern
  recognition, pp. 2678--2687 (2016)

\bibitem{YilmazCVPR2005}
Yilmaz, A., Shah, M.: Actions sketch: A novel action representation.
\newblock In: CVPR (2005)

\bibitem{YuACCV2014}
Yu, G., Liu, Z., Yuan, J.: Discriminative orderlet mining for real-time
  recognition of human-object interaction.
\newblock In: ACCV (2014)

\bibitem{YuICCV2019}
Yu, T., Ren, Z., Li, Y., Yan, E., Xu, N., Yuan, J.: Temporal structure mining
  for weakly supervised action detection.
\newblock In: ICCV (2019)

\bibitem{YuBMVC2010}
Yu, T.H., Kim, T.K., Cipolla, R.: Real-time action recognition by
  spatiotemporal semantic and structural forests.
\newblock In: BMVC (2010)

\bibitem{yuan2013multi-task}
Yuan, C., Hu, W., Tian, G., Yang, S., Wang, H.: Multi-task sparse learning with
  beta process prior for action recognition.
\newblock In: Proceedings of the IEEE Conference on Computer Vision and Pattern
  Recognition, pp. 423--429 (2013)

\bibitem{yuan20133d}
Yuan, C., Li, X., Hu, W., Ling, H., Maybank, S.J.: 3d r transform on
  spatio-temporal interest points for action recognition.
\newblock In: Proceedings of the IEEE Conference on Computer Vision and Pattern
  Recognition, pp. 724--730 (2013)

\bibitem{yuan2014modeling}
Yuan, C., Li, X., Hu, W., Ling, H., Maybank, S.J.: Modeling geometric-temporal
  context with directional pyramid co-occurrence for action recognition.
\newblock IEEE Transactions on Image Processing \textbf{23}(2), 658--672 (2014)

\bibitem{yuan2016fusing}
Yuan, C., Wu, B., Li, X., Hu, W., Maybank, S.J., Wang, F.: Fusing r features
  and local features with context-aware kernels for action recognition.
\newblock International Journal of Computer Vision \textbf{118}(2), 151--171
  (2016)

\bibitem{YuanCVPR2009}
Yuan, J., Liu, Z., Wu, Y.: Discriminative subvolume search for efficient action
  detection.
\newblock In: IEEE Conference on Computer Vision and Pattern Recognition (2009)

\bibitem{YuanPAMI2010}
Yuan, J., Liu, Z., Wu, Y.: Discriminative video pattern search for efficient
  action detection.
\newblock IEEE Transactions on Pattern Analysis and Machine Intelligence
  (2010)

\bibitem{yuan2021agentformer}
Yuan, Y., Weng, X., Ou, Y., Kitani, K.: Agentformer: Agent-aware transformers
  for socio-temporal multi-agent forecasting.
\newblock arXiv preprint arXiv:2103.14023  (2021)

\bibitem{ZengICCV2019}
Zeng, R., Huang, W., Tan, M., Rong, Y., Zhao, P., Huang, J., Gan, C.: Graph
  convolutional networks for temporal action localization.
\newblock In: ICCV (2019)

\bibitem{Zhai2013}
Zhai, X., Peng, Y., Xiao, J.: Cross-media retrieval by intra-media and
  inter-media correlation mining.
\newblock Multimedia Systems \textbf{19}(5), 395--406 (2013)

\bibitem{ZhangTPAMI2016}
Zhang, H., Patel, V.M.: Sparse representation-based open set recognition.
\newblock IEEE transactions on pattern analysis and machine intelligence
  \textbf{39}(8), 1690--1696 (2016)

\bibitem{ARN_few_shot}
Zhang, H., Zhang, L., Qi, X., Li, H., Torr, P.H.S., Koniusz, P.: Few-shot
  action recognition with permutation-invariant attention.
\newblock In: ECCV (2020)

\bibitem{HACS2019}
Zhao, H., Torralba, A., Torresani, L., Yan, Z.: {HACS}: Human action clips and
  segments dataset for recognition and temporal localization.
\newblock In: ICCV (2019)

\bibitem{zhao2021you}
Zhao, H., Wildes, R.P.: Where are you heading? dynamic trajectory prediction
  with expert goal examples.
\newblock In: ICCV (2021)

\bibitem{zhao2017slac}
Zhao, H., Yan, Z., Wang, H., Torresani, L., Torralba, A.: Slac: A sparsely
  labeled dataset for action classification and localization.
\newblock arXiv preprint arXiv:1712.09374  (2017)

\bibitem{ZhaoICCV2017}
Zhao, Y., Xiong, Y., Wang, L., Wu, Z., Tang, X., Lin, D.: Temporal action
  detection with structured segment networks.
\newblock In: ICCV (2017)

\bibitem{zhou2018temporal}
Zhou, B., Andonian, A., Oliva, A., Torralba, A.: Temporal relational reasoning
  in videos.
\newblock In: Proceedings of the European Conference on Computer Vision (ECCV),
  pp. 803--818 (2018)

\bibitem{ZhouCVPR2011}
Zhou, B., Wang, X., Tang, X.: Random field topic model for semantic region
  analysis in crowded scenes from tracklets.
\newblock In: CVPR (2011)

\bibitem{CMN_few_shot}
Zhu, L., Yang, Y.: Compound memory networks for few-shot video classification.
\newblock In: ECCV (2018)

\bibitem{zhu2016co}
Zhu, W., Lan, C., Xing, J., Zeng, W., Li, Y., Shen, L., Xie, X.: Co-occurrence
  feature learning for skeleton based action recognition using regularized deep
  lstm networks.
\newblock In: Thirtieth AAAI Conference on Artificial Intelligence (2016)

\bibitem{ZiebartAAAI2008}
Ziebart, B., Maas, A., Bagnell, J., Dey, A.: Maximum entropy inverse
  reinforcement learning.
\newblock In: AAAI (2008)

\bibitem{ZiebartIROS2009}
Ziebart, B., Ratliff, N., Gallagher, G., Mertz, C., Peterson, K., Bagnell, J.,
  Hebert, M., Dey, A., Srinivasa, S.: Planning-based prediction for
  pedestrians.
\newblock In: IROS (2009)

\end{thebibliography}
\balance
\end{document}